\documentclass[10pt,journal,compsoc]{IEEEtran}
\usepackage{times}
\usepackage[nocompress]{cite}
\usepackage{amsmath}
\usepackage{amsfonts,amssymb,times}
\usepackage{acronym}
\usepackage{algorithm}
\usepackage{algpseudocode}
\usepackage{algcompatible}
\usepackage{array}
\usepackage{tabularx}
\usepackage{multirow}
\usepackage{setspace}

\usepackage{mdwmath}
\usepackage{mdwtab}
\usepackage{eqparbox}

\usepackage{graphicx}
\usepackage{subfigure}
\usepackage{psfrag}

\usepackage{url}
\usepackage{flushend}
\usepackage{here}

\usepackage{bm}
\selectfont
\newcommand{\argmax}{\operatornamewithlimits{\mathrm{arg\,max}}}
\newcommand{\argmin}{\operatornamewithlimits{\mathrm{arg\,min}}}

\begin{document}
\title{Inverted-File k-Means Clustering:\\
	Performance Analysis}
\author{Kazuo~Aoyama,~\IEEEmembership{Member,~IEEE,}
        Kazumi~Saito,
        and~Tetsuo~Ikeda% <-this % stops a space
\IEEEcompsocitemizethanks{\IEEEcompsocthanksitem K. Aoyama is with 
NTT Communication Science Laboratories, 
2-4, Hikaridai, Seika-cho, Soraku-gun, Kyoto 619-0237 Japan.\protect\\
E-mail: kazuo.aoyama.rd@hco.ntt.co.jp,~issei@ieee.org
\IEEEcompsocthanksitem K. Saito is with Kanagawa University,
2946, Tsuchiya, Hiratuka-shi, Kanagawa 259-1293 Japan.
\IEEEcompsocthanksitem 
T. Ikeda is with University of Shizuoka,
52-1, Yada, Suruga-ku, Shizuoka 422-8526 Japan.}% <-this % stops a space
}
%-- \thanks{Manuscript received xxx xxx, 2019; revised xxx xxx, 2019.}}

% The paper headers
%-- \markboth{IEEE TRANSACTIONS ON PATTERN ANALYSIS AND MACHINE INTELLIGENCE,~VOL.~xx,~NO.~xx,~XXX~2020}%
%{Shell \MakeLowercase{\textit{et al.}}: Bare Advanced Demo of IEEEtran.cls for IEEE Computer Society Journals}
% The only time the second header will appear is for the odd numbered pages
% after the title page when using the twoside option.
% 
% *** Note that you probably will NOT want to include the author's ***
% *** name in the headers of peer review papers.                   ***
% You can use \ifCLASSOPTIONpeerreview for conditional compilation here if
% you desire.

\IEEEtitleabstractindextext{%
\begin{abstract}
This paper presents an inverted-file $k$-means clustering algorithm 
(IVF) suitable for a large-scale sparse data set with potentially numerous classes.
Given such a data set, 
IVF efficiently works at high-speed and with low memory consumption, which 
keeps the same solution as a standard Lloyd's algorithm.
The high performance arises from two distinct data representations.
One is a sparse expression for both the object and mean feature vectors.
The other is an inverted-file data structure for a set of the mean feature vectors.
To confirm the effect of these representations, 
we design three algorithms using distinct data structures and expressions for comparison.
We experimentally demonstrate that IVF achieves better performance 
than the designed algorithms 
when they are applied to large-scale real document data sets 
in a modern computer system equipped with 
superscalar out-of-order processors and a deep hierarchical memory system.
We also introduce a simple yet practical clock-cycle per instruction (CPI) model 
for speed-performance analysis.
Analytical results reveal that 
IVF suppresses three performance degradation factors: 
the numbers of cache misses, branch mispredictions, and the completed instructions.
\end{abstract}

}
\maketitle

\IEEEpeerreviewmaketitle

\ifCLASSOPTIONcompsoc
\IEEEraisesectionheading{\section{Introduction}\label{sec:intro}}
\else
\section{Introduction}\label{sec:intro}
\fi
Based on the rapid growth in the ability of various systems 
to collect vast amounts of data, 
machine learning is utilizing large-scale data sets 
for many applications \cite{jordan}.
In this situation, machine learning algorithms are required 
to efficiently process such large-scale data sets 
to withstand practical use.
A leading trend for managing data sets is to employ 
large-scale parallel and distributed computing platforms
\cite{jordan}.
To execute algorithms in the platform, 
modifying and adapting them to the platform is necessary.
By contrast, we must develop a novel algorithm that efficiently operates 
even in a single thread by a single process in a modern computer system,
which maintains adaptability to parallel and distributed platforms.

We deal with a Lloyd-type $k$-means clustering algorithm \cite{wu} 
for operating in a modern computer system.
A standard Lloyd's algorithm \cite{lloyd,macqueen}, 
which is an iterative heuristic algorithm, 
partitions a given object data set into $k$ subsets (clusters)
with given positive integer $k$. 
Repeating two steps of an assignment and an update step 
until a convergence is achieved from a given initial state,
it locally minimizes an objective function 
defined by 
the sum of the squared Euclidean distances between all pairs of 
an object feature vector and a mean feature vector of the cluster 
to which the object is assigned.
The acceleration algorithms, e.g., those in previous work \cite{elkan,hamerly},
have also been reported and 
maintain the same solution as the Lloyd's algorithm 
if they start with an identical initial state.
These general algorithms are independent of a type of object data sets.

A large-scale data set like document collection 
often consists of high-dimensional sparse object feature vectors, 
each of which has a small number of non-zero elements.
A spherical $k$-means algorithm \cite{dhillon} is 
a Lloyd-type algorithm for such a document data set consisting of texts.
Unlike general ones, 
the spherical $k$-means uses feature vectors normalized by their $L_2$ norms, 
i.e., points on a unit hypersphere, as an input data set  
and adopts a cosine similarity for a similarity measure between a pair of points.
A mean vector of each cluster is also normalized by its $L_2$ norm.
An objective function is defined by 
the sum of the cosine similarities between all the pairs of 
an object feature vector and a mean feature vector of the cluster 
to which the object is assigned.
By this procedure, 
a solution by the spherical $k$-means coincides with that 
by the standard Lloyd's algorithm.

It is not trivial what data structures the spherical $k$-means 
should employ for a large-scale sparse data set 
to achieve high performance, i.e., 
to operate at high-speed and with low memory consumption.
Our challenge is to develop 
a high-performance Lloyd-type $k$-means clustering algorithm
for a large-scale data set 
with the low sparsity of a few non-zero elements and potentially numerous classes 
in the same settings as the spherical $k$-means.
We also identify the main factors that determine the performance 
of our newly developed algorithm
by analyzing its operation in a modern computer system.

A modern computer system contains two main components: 
processors and a hierarchical memory system.
A processor has several operating units each of which has 
deep pipelines with superscalar out-of-order execution 
and multilevel cache hierarchy \cite{hennepat}.
The memory system consists of registers and caches 
in a processor and external memories, such as a main memory
and flash storages.
To efficiently operate an algorithm at high throughput in such a system, 
we must prevent pipeline hazards, which cause the pipeline stalls, 
as well as reduce the number of instructions.
One serious hazard is a control hazard induced by 
branch mispredictions \cite{evers,eyerman}.
Another type is data hazards that can occur
when data dependence exists between instructions 
and degrades the pipeline performance \cite{hennepat}.
In the case of cache misses that result in access to external memories,
the degradation becomes conspicuous.
For designing an efficient algorithm, 
the numbers of both branch mispredictions and cache misses must be reduced.

We propose an inverted-file $k$-means clustering algorithm: {\em IVF}.
{\em IVF} utilizes sparse expressions for both the sets of 
given object feature vectors and the mean feature vectors 
for low memory consumption.
In particular, it exploits an inverted-file data structure 
for the mean feature vectors.
An inverted-file data structure is often adopted in search algorithms
for a document data set \cite{samet,harman,knuth,zobel,buttcher}.
In search algorithms, a set of object feature vectors corresponding to 
an {\em invariant} database is structured with an inverted-file format.
Given a query, a search algorithm can find preferable documents quickly
from an inverted-file database.
Our {\em IVF} applies the inverted-file data structure to 
{\em variable} mean feature vectors by 
varying every iteration instead of {\em invariant} object feature vectors.

\vspace*{2mm}
Our contributions are threefold:
\begin{enumerate}
\vspace*{-1mm}
\item 
We present a novel $k$-means clustering algorithm, 
an inverted-file $k$-means clustering algorithm referred to as 
{\em IVF}, for a large-scale and high-dimensional sparse data set 
with potentially numerous classes in Section \ref{sec:prop}.
Our proposed {\em IVF} exploits an inverted-file data structure 
for a set of mean feature vectors, 
while the search algorithms employ the data structure 
for an {\em invariant} sparse data set
like document collection \cite{zobel,buttcher,knuth}.
\item
We propose a simple yet practical clock-cycle per instruction (CPI) model 
for analyzing the factors of computational cost.
To identify them based on the CPI model, 
we prepare different data structures for a set of mean feature vectors
and compare {\em IVF} to the algorithms using those data structures.
\item
We experimentally demonstrate that {\em IVF}
achieves superior high-speed and low memory consumption performance
when it is applied to large-scale and high-dimensional 
real document data sets with large $k$ values.
The low memory consumption is caused by the data structure with sparse expressions
of both data object and mean feature vectors.
By analyzing the results obtained with the {\em perf tool} \cite{perf} 
based on the CPI model,
{\em IVF}'s high speed is clearly attributed to three factors: 
fewer cache misses, fewer branch mispredictions, and 
fewer instructions.
They are detailed in Sections~\ref{sec:exp} and \ref{sec:analy}.
\end{enumerate}

The remainder of this paper consists of the following seven sections.
Section \ref{sec:relate} briefly reviews related work 
from viewpoints that clarify the distinct aspects of our work.
Section \ref{sec:prop} explains our proposed {\em IVF}.
Section \ref{sec:algos} describes the designed algorithms for comparison.
Section \ref{sec:exp} shows our experimental settings and demonstrates the results.
Section \ref{sec:analy} determines why {\em IVF} achieves high performance
with a simple yet practical CPI model.
Section \ref{sec:disc} discusses {\em IVF}'s performance and 
compares it to other similar algorithms.
The final section provides our conclusion and future work.

\section{Related work}\label{sec:relate}
This section reviews four distinct topics: 
Lloyd-type $k$-means clustering algorithms, 
a spherical $k$-means for document data sets, 
which is a variant of Lloyd-type algorithms,
an inverted-file data structure for sparse data sets, and 
design guidelines for efficient algorithms suitable 
for modern computer systems.

\subsection{Lloyd-Type k-Means Clustering Algorithm}\label{subsec:lloyd}
We begin by defining a $k$-means clustering problem.
Given a set of object feature vectors that are points
in a $D$-dimensional Euclidean space, 
${\cal X}\!=\!\{ \bm{x}_1, \bm{x}_2,\cdots,\bm{x}_N \}$, 
$|{\cal X}|\! =\! N$, $\bm{x}_i\! \in\! \mathbb{R}^D$, 
and a positive integer of $k$,
a $k$-means clustering problem 
finds 
a set of $k$ clusters, 
${\cal C}^*\!=\!\{ C^*_1,C^*_2,\cdots,C^*_k \}$: 
\begin{equation}
{\cal C}^* = \argmin_{{\cal C}=\{ C_1,\cdots, C_k\}}
\left(\: \sum_{C_j\in{\cal C}}\:\sum_{\bm{x}_i\in C_j}
\|\bm{x}_i-\bm{\mu}_j \|_2^2 \:\right)\:, 
\label{eq:obj_funct}
\end{equation}
where $\|\!\star\!\|_2$ denotes the $L_2$ norm of a vector,
${\cal C}$ is the set of $k$ clusters, and  
$\bm{\mu}_j\! \in\! \mathbb{R}^D$ is 
the mean feature vector of cluster $C_j$.
Solving the $k$-means clustering problem 
expressed by Eq.~(\ref{eq:obj_funct}) 
is difficult 
in practical use due to a high computational cost \cite{aloise}.

Instead of a precise solution to the problem,
a standard Lloyd's algorithm \cite{lloyd,macqueen} finds a local minimum 
in an iterative heuristic manner.
The algorithm repeats two steps of an assignment and an update step
until the convergence or a predetermined termination condition is satisfied.

{\bf Algorithm} \ref{algo:lloyd} shows an overview 
of a standard Lloyd's algorithm at the $r$-th iteration.
The assignment step at lines~5--13 
assigns a point represented by object feature vector $\bm{x}_i$ 
to cluster $C_j$ whose centroid 
(mean at the previous iteration $\bm{\mu}_j^{[r-1]}$) 
is closest to $\bm{x}_i$.
At line~9, $d_{min}$ denotes the tentative minimum distance 
from $\bm{x}_i$ to the centroids and 
$a(\bm{x}_i)$ is a function of $\bm{x}_i$ that returns closest
centroid ID $j$.
The update step at line~15 calculates mean feature vector 
$\bm{\mu}_j^{[r]}\!\in\!{\cal M}^{[r]}$ at the $r$-th iteration 
using object feature vectors $\bm{x}_{i}\!\in\! C_j^{[r]}$.

\begin{algorithm}[!t]
  \caption{~~Standard Lloyd's algorithm at the $r$-th iteration} 
  \label{algo:lloyd}
    \begin{algorithmic}[1]
		\STATE{\textbf{Input:} ${\cal X}$,~~
		${\cal M}^{[r-1]}\!=\!\{\bm{\mu}_1^{[r-1]},\cdots,\bm{\mu}_k^{[r-1]}\}$,
		~~($k$)}
		\STATE{\textbf{Output:} ${\cal C}^{[r]}\!=\!\{ C_1^{[r]},C_2^{[r]},\cdots,C_k^{[r]}\}$,~
				${\cal M}^{[r]}$}
		\STATE{$C_{j}^{[r]}\leftarrow\emptyset$~,~~$j=1,2,\cdots,k$~}
		\\\COMMENT{~$/\!/$--~{\bf Assignment step}~--$/\!/$~}
		\FORALL{$\bm{x}_i \in {\cal X}$~}\label{algo:lloyd-start}
		\STATE{$d_{min}\leftarrow 
			d(\bm{x}_i,\bm{\mu}_{a(\bm{x}_i)}^{[r-1]})\!=\!
			\| \bm{x}_i -\bm{\mu}_{a(\bm{x}_i)}^{[r-1]} \|_2$}
		\FORALL{$\bm{\mu}_j^{[r-1]} \in {\cal M}^{[r-1]}$}
			\IF{$d(\bm{x}_i,\bm{\mu}_j^{[r-1]})< d_{min}$~}
				\STATE{$d_{min}\leftarrow d(\bm{x}_i,\bm{\mu}_j^{[r-1]})$~~and~~
						$a(\bm{x}_i)\leftarrow j^{[r-1]}$}
			\ENDIF
		\ENDFOR
		\STATE{$C_{a(\bm{x}_i)}^{[r]}\leftarrow 
				C_{a(\bm{x}_i)}^{[r]}\cup\{ \bm{x}_i\}$}
		\ENDFOR \label{algo:lloyd-end}
		\\\COMMENT{~$/\!/$--~{\bf Update step}~--$/\!/$~}
		\STATE{$\bm{\mu}_j^{[r]}\leftarrow 
			\left( \sum_{\bm{x}_i\in C_j^{[r]}} \bm{x}_i \right)\,/\,|\,C_j^{[r]}\,|$,
			~~$j=1,2,\cdots,k$}
			\label{algo:lloyd-update}
       	\STATE{\textbf{return}~~${\cal C}^{[r]}\!=\!\{ C_1^{[r]},C_2^{[r]},\cdots,C_k^{[r]}\}$,~
				${\cal M}^{[r]}$}
\end{algorithmic}
\end{algorithm}

Acceleration algorithms have also been reported
\cite{elkan,hamerly,drake,ding,newling,hattori,aoyama}, 
which find the same local minimum 
as the standard Lloyd's algorithm if they start at the identical initial state. 
To eliminate the costly distance calculations at line 8, 
they exploit the inexpensive lower bound on the exact distance.
Since the lower bound is calculated based on the triangle inequality in a metric space,
the acceleration strategy is a general one 
independent of the type of given object feature vectors.

\subsection{Spherical k-Means Clustering Algorithm}\label{subsec:sph-kmeans}
A spherical $k$-means algorithm \cite{dhillon} is a special type 
for document data sets where each object is a text that consists of terms, 
such as words and phrases.
The object is represented by a sparse feature vector, 
where the dimensionality of a feature space containing all the feature vectors
is the number of distinct terms and 
an element of a feature vector is a feature value given to a term
such as {\em tf-idf} (term-frequency inverse-document-frequency) \cite{buttcher}.
Define {\em sparsity} $\eta(\bm{x}_i)$ of feature vector 
$\bm{x}_i\!\in\!\mathbb{R}^D$
and {\em average sparsity} $\bar{\eta}({\cal X})$ of set 
${\cal X}\!=\!\{\bm{x}_1,\bm{x}_2,\cdots,\bm{x}_N\}$: 
\begin{eqnarray}
\eta(\bm{x}_i) &=& \|\bm{x}_i\|_0 / D\: ,\\
\label{eq:sparsity}
\bar{\eta}({\cal X}) &=& \textstyle{\sum_{i=1}^N \eta(\bm{x}_i) / N}\: ,
\label{eq:avg_sparsity}
\end{eqnarray}
where $\|\bm{x}_i\|_0$ denotes the $L_0$ norm of $\bm{x}_i$.

\begin{algorithm}[!t]
  \caption{~~Spherical $k$-means algorithm at the $r$-th iteration} 
  \label{algo:sph}
    \begin{algorithmic}[1]
		\STATE{\textbf{Input:} ${\cal X}$,~~
		${\cal M}^{[r-1]}\!=\!\{\bm{\mu}_1^{[r-1]},\cdots,\bm{\mu}_k^{[r-1]}\}$,
		~~($k$)}
		\STATE{\textbf{Output:} ${\cal C}^{[r]}\!=\!\{ C_1^{[r]},C_2^{[r]},\cdots,C_k^{[r]}\}$,~
				${\cal M}^{[r]}$}
		\STATE{$C_{j}^{[r]}\leftarrow\emptyset$~,~~$j=1,2,\cdots,k$~}
		\\\COMMENT{~$/\!/$--~{\bf Assignment step}~--$/\!/$~}
		\FORALL{$\bm{x}_i \in {\cal X}$~}\label{algo:sph-start}
		\STATE{$\rho_{max}\leftarrow 
			\bm{x}_i\cdot \bm{\mu}_{a(\bm{x}_i)}^{[r-1]}$ }
		\FORALL{$\bm{\mu}_j^{[r-1]} \in {\cal M}^{[r-1]}$}
			\IF{$\bm{x}_i\cdot \bm{\mu}_j^{[r-1]} > \rho_{max}$~}
				\STATE{$\rho_{max}\leftarrow \bm{x}_i\cdot \bm{\mu}_j^{[r-1]}$~~and~~
						$a(\bm{x}_i)\leftarrow j^{[r-1]}$}
			\ENDIF
		\ENDFOR
		\STATE{$C_{a(\bm{x}_i)}^{[r]}\leftarrow 
				C_{a(\bm{x}_i)}^{[r]}\cup\{ \bm{x}_i\}$}
		\ENDFOR \label{algo:sph-end}
		\\\COMMENT{~$/\!/$--~{\bf Update step}~--$/\!/$~}
		\STATE{$\bm{\mu}_j^{[r]}\leftarrow 
			\left( \sum_{\bm{x}_i\in C_j^{[r]}} \bm{x}_i \right)\,/\,|\,C_j^{[r]}\,|$,
			~~$j=1,2,\cdots,k$}
		\STATE{$\bm{\mu}_j^{[r]}\leftarrow \bm{\mu}_j^{[r]}\,/\,\|\bm{\mu}_j^{[r]}\|$}
			\label{algo:sph-update}
       	\STATE{\textbf{return}~~${\cal C}^{[r]}\!=\!\{ C_1^{[r]},C_2^{[r]},\cdots,C_k^{[r]}\}$,~
				${\cal M}^{[r]}$}
\end{algorithmic}
\end{algorithm}

The spherical $k$-means assumes that 
object feature vector $\bm{x}_i\!\in\!\mathbb{R}^D$
is normalized by its $L_2$ norm as $\|\bm{x}_i\|_2\!=\!1$, i.e.,  
a point on a unit hypersphere.
Instead of a Euclidean distance used by the standard  $k$-means algorithm, 
the spherical $k$-means algorithm employs
a cosine similarity between $\bm{x}_i$ and $\bm{\mu}_j$,  
expressed by 
\begin{equation}
\rho(\bm{x}_i,\bm{\mu}_j)\!=\! \bm{x}_i\cdot \bm{\mu}_j~ ,
\end{equation}
where 
$\bm{x}_i\!\cdot\! \bm{\mu}_j$ denotes the inner product of $\bm{x}_i$ and
$\bm{\mu}_j$ and $\|\bm{\mu}_j\|_2\!=\!1$, 
i.e., $\bm{\mu}_j$ is a point on the unit hypersphere.
Then the spherical $k$-means clustering problem is formulated as 
\begin{equation}
{\cal C}^* = \argmax_{{\cal C}=\{ C_1,\cdots, C_k\}}
\left(\: \sum_{C_j\in{\cal C}}\:\sum_{\bm{x}_i\in C_j}
\bm{x}_i\cdot\bm{\mu}_j \:\right)\:.
\label{eq:sph_obj_funct}
\end{equation}
Under the condition of $\|\bm{x}_i\|_2\!=\!\|\bm{\mu}_j\|_2\!=\!1$,
Eqs.~(\ref{eq:obj_funct}) and (\ref{eq:sph_obj_funct}) are equivalent 
\footnote{
If mean feature vectors are not normalized by their $L_2$ norms, i.e.,
they are not points on the unit hypersphere, 
a solution by the spherical $k$-means algorithm does not always coincide with
that by the standard $k$-means algorithm.
}
because $\|\bm{x}_i-\bm{\mu}_j\|_2^2\!= 2(1 -\!\bm{x}_i\cdot \bm{\mu}_j)$. 
The spherical $k$-means algorithm based on the same iterative heuristics
as the standard one is shown in {\bf Algorithm} \ref{algo:sph}.
Thus the spherical $k$-means algorithm \cite{dhillon} 
corresponds to the standard $k$-means algorithm 
for a general data set in the Euclidean space.
In the previous work \cite{dhillon}, neither its acceleration algorithms nor 
how to leverage sparseness of a data (object) set is disclosed. 
Our work is based on the same settings as the spherical $k$-means 
and provides an efficient algorithm that exploits the sparseness of 
a data set.

\subsection{Inverted-File Data Structure}\label{subsec:ivfstr}
\begin{figure}[t]
\begin{center}
	\subfigure[Standard structure]{
		\psfrag{W}[c][c][0.82]{${\bm x}_i$}
		\psfrag{M}[c][c][0.82]{$\hat{\bm x}_i$}
		\includegraphics[width=61.55mm]{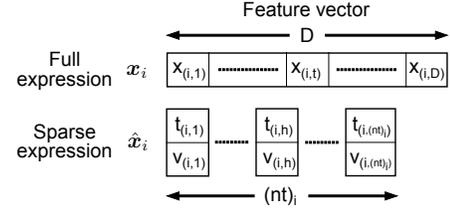}}\\
	\subfigure[Inverted-file data structure]{
		\psfrag{W}[c][c][0.78]{${\bm x}_i$}
		\psfrag{Y}[c][c][0.82]{${\bm y}_t$}
		\psfrag{P}[c][c][0.82]{$\breve{\bm y}_t$}
		\includegraphics[width=88mm]{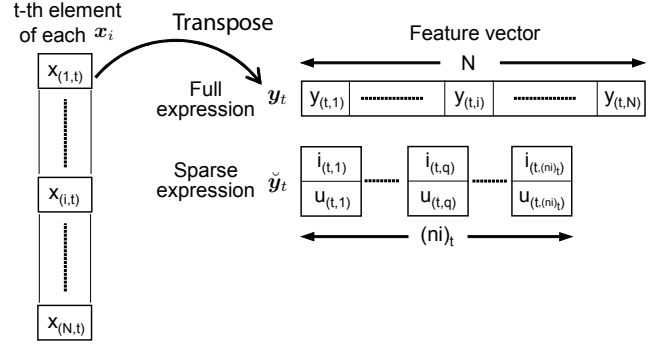} }
\vspace*{-5mm}
\end{center}
\caption{Data structures and expressions of feature vectors: 
(a) In standard structure, full expression of feature vector 
$\bm{x}_i\!\in\! \mathbb{R}^D$ and sparse one of 
$\hat{\bm x}_i\!=\! (t_{(i,h)},v_{(i,h)})$, $h\!=\! 1,2,\cdots,(nt)_i$ are
illustrated at top and bottom figures.
(b) In inverted-file data structure,
right top figure shows vector $\bm{y}_t$ with a full expression,
which contains the $t$-th elements of $\bm{x}_i$, $i\!=\!1,\cdots,N$.
Right bottom figure shows sparse $\breve{\bm y}_t$ that consists of tuples of
object ID $i_{(t,q)}$ and feature $u_{(t,q)}$, $q\!=\! 1,2,\cdots,(ni)_t$.
}
\label{fig:str}
\end{figure}

An inverted file is a type of data structures that is often employed 
for a data set of texts consisting of sparse feature vectors \cite{samet}.
Instead of listing the feature elements of a given object, 
we list the objects with a given feature element for the inverted file \cite{knuth}.

Figure~\ref{fig:str} shows a full and a sparse expression of the object feature vectors
in (a) a standard structure and (b) an inverted-file data structure.
In the standard structure, a feature vector with a full expression is represented by 
$\bm{x}_i\!\in\!\mathbb{R}^D$, $i\!=\!1,\cdots,N$, where element $x_{(i,t)}$ whose  
term does not appear in the $i$-th object is padded by zero.
Define a set of non-zero elements in $\bm{x}_i$ as 
$\hat{\cal X}_i\!=\!\{ (t,x_{(i,t)})| x_{(i,t)}\!\neq\!0 \}$.
Then a feature vector with sparse expression $\hat{\bm x}_i$ is represented by
$\hat{\bm x}_i\!=\!(t_{(i,h)},v_{(i,h)})$, $h\!=\!1,2,\cdots,|\hat{\cal X}_i|\!=\!(nt)_i$, 
where $v_{(i,h)}\!=\!x_{(i,t_{(i,h)})}$.
Assume that the $t$-th elements in each $\bm{x}_i$ are picked up as a vector like
Fig.~\ref{fig:str}(b), left.
The transpose of the vector is a feature vector with a full expression 
in the inverted-file data structure, which is represented by 
$\bm{y}_t\!=\!(y_{(t,1)},\cdots,y_{(t,N)})$, $y_{(t,q)}\!=\!x_{(q,t)}$.
We define a set of $\bm{y}_t$ as 
$\bar{\cal X}\!=\!\{\bm{y}_1,\bm{y}_2,\cdots,\bm{y}_D\}$.
Similar to the standard structure, 
a feature vector with sparse expression $\hat{\bm y}_t$ is represented by 
$\breve{\bm y}_t\!=\!(i_{(t,q)},u_{(t,q)})$, $q\!=\!1,2,\cdots,(ni)_t$, 
where $u_{(t,q)}\!=\!y_{(t,i_{(t,q)})}\!=\!x_{(i_{(t,q)},t)}$.
Besides, $\breve{\cal X}\!=\!\{\breve{\bm y}_1,\breve{\bm y}_2,\cdots,\breve{\bm y}_D\}$.
Note that we adopt a simple array among several sparse expressions.

By applying the sparse expression to a given object set with low sparsity,
we can conserve the memory size 
although the sparse expression needs extra memory capacitance for storing
term IDs as $t_{(i,h)}$ or object IDs as $i_{(t,q)}$.
Most text-search algorithms utilize  
an inverted file (or an inverted index) prebuilt from a text data set 
as a database.
Given a query that is often a set of terms such as words or phrases, 
the search algorithms find relevant texts to the query 
from the text data set using terms in the query as search keys
\cite{harman,knuth,buttcher,zobel}.

As well as text search, 
an inverted-file structure has also been employed for image search \cite{sivic}.
A search algorithm for object retrieval in videos employed 
visual words for a feature of an image (a video frame) \cite{sivic}.
The visual words are generated by the vector quantization of local descriptors 
extracted from images.
Consequently, each image is represented as a sparse feature vector, 
each element of which is a tuple of a visual word ID and a feature value ({\em tf-idf}).
Based on this representation, similar to a text search, 
the inverted-file data structure for an image database is utilized to perform a fast search.

So far, the data structure is based on the relationship between an object and the terms 
contained by it.
By using not the foregoing relationship but
the relationship between an object and the clusters to which the object belongs, 
the concept of the inverted-file structure is extended in an image search \cite{jegou,babenko}.
In this case \cite{jegou,babenko}, 
each object is assigned to a disjoint cluster by vector quantization 
based on $k$-means clustering.
In the inverted-file structure, 
objects are listed for each cluster that contains the objects as its members.
Given an image as a query,
product quantization \cite{jegou} narrows down a search space 
to a subset (cluster) to which the query belongs.
This extended inverted-file structure resembles a hash table employed 
in a local-sensitive hashing (LSH) approach \cite{indyk,andoni,charikar}.

There is a $k$-means clustering algorithm that directly exploits 
a search algorithm using an inverted-file structure at its assignment step 
\cite{broder}.
A Lloyd-type algorithm uses 
a linear scan (brute-force) search at the assignment step 
to find the most similar centroid to each object.
Similar to text-search algorithms, 
the reported algorithm called wand-k-means \cite{broder} 
applies an inverted-file structure to a set of {\em invariant} data objects.
The wand-k-means regards a set of centroids as queries and 
finds similar objects to each of the queries by a heuristic search algorithm 
called WAND at the assignment step.
Except for the search algorithm,
an important difference between wand-k-means and our {\em IVF} 
is in feature vectors represented with an inverted-file structure: 
{\em invariant} data object feature vectors and 
{\em variable} mean feature vectors at every iteration.
This difference prompts the question: which can better achieve high performance?
We discuss this issue in Section~\ref{subsec:ivfd}.

\subsection{Design Guidelines of Efficient Algorithms}\label{subsec:design}
For efficiently processing 
a large-scale high-dimensional data set in a modern computer system, 
parallel processing is effective.
There are several levels in parallel processing: 
instruction-level parallelism (ILP),
data-level parallelism (DLP), and thread-level parallelism (TLP) \cite{hennepat}. 
We focus on ILP 
and design an efficient algorithm for a single thread by a single process.
A Lloyd-type $k$-means clustering algorithm 
operating at high throughput in ILP 
achieves high performance in other parallelisms.
This is because its procedure is suitable 
for explicit parallelisms at their costly assignment step, where 
a linear scan search for each object independently identifies the most similar centroid 
(mean) to the object in all the $k$ centroids \cite{jian,bhimani}.

To completely exploit ILP in a modern computer system,
which has deep pipelines with superscalar out-of-order execution in a CPU core
and a deep memory hierarchy from registers to external storages,
pipeline hazards that cause stalls must be reduced.
Among them, 
control hazards caused by branch mispredictions and 
data hazards arising from the dependency of instructions 
on the results of previously executed instructions
are critical to increase the performance of the algorithms and their implementations.

The impact of branch mispredictions on algorithm performance has 
been analyzed, and 
algorithms that reduce the branch mispredictions have been developed 
\cite{kaligosi,edelkamp,green}.
For a classical quicksort, which is a well-known sort algorithm,
a counterintuitive observation of 
selecting as a pivot not a median of a partitioned array but a skewed pivot 
(an entry distant from the median) leads to high performance
is analyzed and explained based on the balance of the number of comparison operations
and branch mispredictions \cite{kaligosi}.
BlockQuicksort \cite{edelkamp}, 
which is a kind of the dual-pivot quicksort, 
suppresses the branch mispredictions incurred by conditional branches.
Besides sort algorithms, 
the classic Shiloach-Vishkin algorithm that finds connected components, 
which is one graph algorithm, was improved in terms of speed performance 
by avoiding branch mispredictions \cite{green}.

Data hazards accompanied by access to external memories like DRAMs 
seriously affect the speed performance because of high memory latency.
To prevent this performance degradation,
algorithms and their implementations have been studied,
which efficiently exploit caches in a CPU core 
for reducing expensive access to external memories 
\cite{kowarschik,frigo}.
Cache-aware (-conscious) algorithms \cite{kowarschik} are optimized 
based on such actual parameters as capacity, block size, and associativity 
for increasing the cache hit rate, 
while cache-oblivious algorithms \cite{frigo} 
are designed and tuned with cache consideration and  
without variables that are dependent on the actual parameters.
Frequent pattern mining algorithms \cite{ghoting} reduce cache misses by 
improving spatial and temporal locality in data access with cache-conscious methods,
resulting in high performance.
A similarity join algorithm \cite{perdacher} achieves high-speed performance
by transforming a conventional loop iteration into a cache-oblivious one. 

Thus, preventing pipeline hazards is important 
for designing a high-performance algorithm suitable for a modern computer system.
Although our proposed {\em IVF} is not a cache-aware algorithm, 
its structure suppresses the pipeline hazards 
shown in Sections~\ref{sec:prop} and \ref{sec:exp}.
The algorithm is analyzed from the viewpoint of the foregoing 
performance degradation factors that cause pipeline hazards 
in Section~\ref{sec:analy}.

\section{Proposed algorithm: IVF}\label{sec:prop}
\begin{algorithm}[!t]
  \caption{~~Proposed {\em IVF} at the $r$-th iteration} 
  \label{algo:ours}
    \begin{algorithmic}[1]
		\STATE{\textbf{Input:} $\hat{\cal X}$,~~
			\framebox[19mm][c]{$\breve{\cal M}^{[r-1]}$~~{\small (I)}}~,~~($k$)}
		\STATE{\textbf{Output:} ${\cal C}^{[r]}\!=\!\{ C_1^{[r]},C_2^{[r]},\cdots,C_k^{[r]}\}$,~
				\framebox[16mm][c]{$\breve{\cal M}^{[r]}$~~{\small (I')}}}
		\STATE{$C_{j}^{[r]}\leftarrow\emptyset$~,~~$j=1,2,\cdots,k$~}
		%
		%-- Assignment step --%
		\\\COMMENT{~$/\!/$--~{\bf Assignment step}~--$/\!/$~}
		\FORALL{$\hat{\bm{x}}_i\!=\!(t_{(i,h)},v_{(i,h)})_{h=1}^{(nt)_i}\in \hat{\cal X}$~~}
			\label{algo:prop-start}
		\STATE{$\rho_{max}\!\leftarrow\! 0$,~~
				$\bm{\rho}\!=\!(\rho_1,\rho_2,\cdots,\rho_j,\cdots,\rho_k)\!\leftarrow\!\bm{0}$}
		\STATE{$S_i\!=\!\{t_{(i,1)},~t_{(i,2)},~\cdots,~t_{(i,h)},~\cdots,~t_{(i,(nt)_i)}\}$}
		\FORALL{~$s\!\leftarrow\!t_{(i,h)}\in S_i$~}
		\FORALL{~\framebox[48mm][c]{
			$(c_{(s,q)},u_{(s,q)})^{[r-1]} \in \breve{\bm{\xi}}_s^{[r-1]}$~~{\small (I\hspace{-.1em}I)}}~~}
		\STATE{\framebox[71mm][l]{
			$\breve{\bm{\xi}}_s^{[r-1]}\!=\![(c_{(s,q)},u_{(s,q)})_{q=1}^{(nc)_s}]^{[r-1]} 
				\in\!\breve{\cal M}^{[r-1]}$~{\small (I\hspace{-.1em}I')} }
		}
		\STATE{\framebox[71mm][c]{\small (I\hspace{-.1em}I\hspace{-.1em}I)}}
		\STATE{\framebox[71mm][l]{
			~$\rho_{c_{(s,q)}}\leftarrow \rho_{c_{(s,q)}}+v_{(i,h)}\!\times\! u_{(s,q)}$~~~~
			{\small (I\hspace{-.1em}V)}}}
		\ENDFOR\ENDFOR
		\FOR{~~$j\!=\! 1$ to~~$k$~~}
			\STATE{{\bf if}~~~$\rho_j\!>\! \rho_{max}$~~~{\bf then}~
					$\rho_{max}\!\leftarrow\!\rho_j$~and~$a(\hat{\bm{x}}_i)\!\leftarrow\! j$}
		\ENDFOR
		\STATE{$C_{a(\hat{\bm{x}}_i)}^{[r]}\leftarrow 
			C_{a(\hat{\bm{x}}_i)}^{[r]}\cup\{ \hat{\bm{x}}_i\}$}
		\ENDFOR
		%
		%-- Update step --%
		\\\COMMENT{~$/\!/$--~{\bf Update step}~--$/\!/$~}
		\STATE{$q_p\!\leftarrow\!0$,~~~$p=1,2,\cdots,D$}
		\FORALL{$C_j^{[r]} \in {\cal C}^{[r]}$}
		\STATE{$\bm{w}\!=\!(w_1,w_2,\cdots,w_D)\!\leftarrow\! \bm{0}$}
		\FORALL{$\hat{\bm x}_i \in C_j^{[r]}$}
		\STATE{{\bf for~all}~$s\!\leftarrow\!t_{(i,h)}\in S_i$~{\bf do}~
			$w_{s}\!\leftarrow\! w_{s}\! +\!v_{(i,h)}$~{\bf end~for}}
		\ENDFOR
		\STATE{{\bf for}~~$p\!=\!1$~~to~~$D$~~{\bf do}~
			$w_p\!\leftarrow\! w_p/|C_j^{[r]}|$~~{\bf end~for}}
		\FOR{~$p\!=\!1$~~to~~$D$~} 
		\IF{~$w_p\!\neq\!0$~~}
		\STATE{$c_{(p,q_p)}\!\leftarrow\! j$,~~$u_{(p,q_p)}\!\leftarrow\! w_p/\|\bm{w}\|_2$,~~
			$q_p\!\leftarrow\! q_p\!+\!1$
		}
		\ENDIF\ENDFOR
		\ENDFOR
       	\STATE{\textbf{return}~~${\cal C}^{[r]}\!=\!\{ C_1^{[r]},C_2^{[r]},\cdots,C_k^{[r]}\}$,~
				\framebox[16mm][c]{$\breve{\cal M}^{[r]}$~~{\small (I')}} }
\end{algorithmic}
\end{algorithm}

We propose an inverted-file $k$-means clustering algorithm ({\em IVF}) 
for a large-scale and high-dimensional sparse data set with potentially
numerous classes.
{\em IVF} is a Lloyd-type algorithm, i.e., an iterative heuristic algorithm, 
which keeps the same solution as a standard Lloyd algorithm \cite{lloyd,macqueen} 
under an identical initial state.
Due to this property, we do not discuss accuracy
(or an objective function value) 
as performance.
We evaluate both the maximum memory capacitance and the CPU time (or the clock cycles)
required by the algorithm through iterations.

{\bf Algorithm}~\ref{algo:ours} shows the {\em IVF} pseudocode 
at the $r$-th iteration.
{\em IVF} receives 
a centroid set, which is the mean set at the previous iteration, 
with inverted-file sparse expression $\breve{\cal M}^{[r-1]}$ 
and uses an invariant object set 
with standard sparse expression $\hat{\cal X}$ 
and returns cluster set ${\cal C}^{[r]}$ consisting of $k$ clusters and 
$\breve{\cal M}^{[r]}$.

{\em IVF} has two steps; assignment and update.
The assignment step at lines~5--19 executes a linear-scan search in 
the triple loop, where an object feature vector is regarded as a query.
The $i$-th object feature vector ($\hat{\bm{x}}_i$) 
consists of 
$(nt)_i$ tuples $(t_{(i,h)},v_{(i,h)})$, $h\!=\!1,2,\cdots,(nt)_i$, where 
$(nt)_i$ denotes the $L_0$ norm of $\bm{x}_i$ ($(nt)_i\!=\!\| \bm{x}_i\|_0$),
$h$ is the local counter, $t_{(i,h)}$ is the global (serial) term ID from 1 to $D$,
and $v_{(i,h)}$ is a corresponding value
such as {\em tf-idf}.
For each term with term ID $t_{(i,h)}$ ($s$ for simplicity),
inverted-file centroid array $\breve{\bm{\xi}}_s^{[r-1]}$ is selected.
This array consists of $(nc)_s$ tuples $(c_{(s,q)},u_{(s,q)})^{[r-1]}$, 
$q\!=\!1,2,\cdots,(nc)_s$, 
where $c_{(s,q)}$ denotes the global centroid ID from 1 to $k$, 
$u_{(s,q)}$ is the corresponding value, and 
$(nc)_s$ denotes the centroid (mean) frequency of term ID $s$. 
Then the partial similarity (inner product) between the $i$-th object and 
the $c_{(s,q)}$-th centroid is calculated and stored at $\rho_{c_{(s,q)}}$.
Just after the inner double loop has been completed,
the $i$-th object is assigned to the $a(\hat{\bm{x}}_i)$-th cluster 
whose centroid most closely resembles.

The update step at lines~21--33 calculates each mean of $k$ clusters 
based on the object assignment.
For cluster $C_j^{[r]}$ whose members $\hat{\bm x}_i\!\in\! C_j^{[r]}$ 
are determined at the assignment step,
each feature value $v_{(i,h)}$ is added to $w_s$, where $s$ denotes global term ID 
$t_{(i,h)}$ from 1 to $D$.
After the addition for all the members, 
each value $w_p$ ($1\!\leq\! p\!\leq\! D$) is divided by cluster size $|C_j^{[r]}|$
and $\|\bm{w}\|_2$ is calculated.
To represent the mean of $C_j^{[r]}$ with the inverted-file sparse expression,
we perform the procedure at lines~28--32, 
where $p$ denotes the global term ID and $q_p$ is the local counter of $p$.
Then the mean of $C_j^{[r]}$ is expressed by 
a set of tuples $(c_{(p,q_p)},u_{(p,q_p)})$ where 
$c_{(p,q_p)}$ denotes cluster ID $j$ and 
$u_{(p,q_p)}$ is the corresponding feature value.
Thus the tuple $(c_{(p,q_p)},u_{(p,q_p)})$, which is the $q_p$-th element of 
$\breve{\bm \xi}_p^{[r]}$, is obtained.

{\em IVF} simultaneously 
satisfies the two requirements of low memory consumption and high speed.
The sparse expressions for both object set $\hat{\cal X}$ 
and mean set $\breve{\cal M}$ suppress memory consumption.
The inverted-file data structure for the mean (centroid) set 
achieves high-speed performance.
To qualitatively evaluate the {\em IVF} performance,
we design three algorithms in Section~\ref{sec:algos} and
compare {\em IVF} with them in two distinct real document data sets 
in Section~\ref{sec:exp}.
Furthermore, we analyze the speed performance to identify factors that 
determine the performance in Section~\ref{sec:analy}.

\section{Compared algorithms}\label{sec:algos}
\begin{table}[t]
\centering
\caption{Classification of compared algorithms}
\vspace*{-2mm}
\begin{spacing}{1.08}
\begin{tabular}{|c|c|c|c|}\hline
Data & \multicolumn{3}{c|}{Mean expression}\tabularnewline \cline{2-4}
structure & Sparse & \multicolumn{2}{c|}{Full} \tabularnewline\hline
\multirow{2}{*}{Standard} & Two-way merge & Non-branch & Branch 
\tabularnewline
 & {\em TWM} \cite{knuth} & {\bf\em MFN} & {\em MFB} \tabularnewline \hline
\multirow{2}{*}{Inverted-file} & Proposed & Non-branch & Branch 
\tabularnewline
 & {\bf\em IVF} & {\bf\em IFN} & {\bf\em IFB} \tabularnewline\hline
\end{tabular}\label{table:class}
\end{spacing}
\vspace*{-2mm}
\end{table}

To shed light on the characteristics of {\em IVF},
we designed three algorithms, which may be not suitable for practical use
due to their required memory capacitances.
One is called a mean full-expression algorithm with a non-branch ({\em MFN}).
The others are 
an inverted-file full-expression algorithm with branch ({\em IFB}) and 
non-branch ({\em IFN}).
Similar to {\em IVF}, 
all three algorithms represent a given object set with 
a standard-structure sparse expression in Fig.~\ref{fig:str}(a) bottom.
The difference is in their data structures and expressions for a mean set.
Table~\ref{table:class} shows the classification of the three algorithms and 
{\em IVF}.

{\em MFN} employs a standard data structure with a full expression 
for a mean (centroid) set shown in Fig.~\ref{fig:str}(a) top, 
where subscript $i$ is replaced with $j$ for the means, $j\!=\! 1,2,\cdots,k$.
Mean sets $\breve{\cal M}^{[r-1]}$ and $\breve{\cal M}^{[r]}$ at 
lines 1 (I), 2 (I'), and 34 (I') in {\bf Algorithm}~\ref{algo:ours} are replaced with 
${\cal M}^{[r-1]}$ and ${\cal M}^{[r]}$.
When mean feature vector $\bm{\mu}_j$ is represented with a full expression,
values of entries for some global term IDs may be undefined.
Then each of the entries is padded with zero.
The similarity between object feature vector 
$\hat{\bm x}_i\!=\!((t_{(i,1)},v_{(i,1)}),\cdots,(t_{(i,(nt)_i)},v_{(i,(nt)_i)}))$ 
and centroid (mean) feature vector $\bm{\mu}_j$ is calculated by 
\begin{equation}
\rho_j = \sum_{h=1}^{(nt)_i} v_{(i,h)}\!\times\! \mu_{(j,t_{(i,h)})}\: ,
\label{eq:mfn_sim}
\end{equation}
where $\mu_{(j,t_{(i,h)})}$ denotes the element with the global term ID of $t_{(i,h)}$
in $\bm{\mu}_j$.
When $\mu_{(j,t_{(i,h)})}\!=\!0$ in Eq.~(\ref{eq:mfn_sim}), 
there are two approaches:
the execution of zero multiplication and 
the insertion of the conditional branch for skipping the zero multiplication.
{\em MFN} employs the former approach.
We call the former approach {\em non-branch} and the latter {\em branch}.
From the algorithmic point of view,  
lines 9 (I\hspace{-.08em}I) and 10 (I\hspace{-.08em}I') in {\bf Algorithm}~\ref{algo:ours} 
are replaced as follows.\\

\vspace*{-1.1mm}
\noindent\hspace*{8mm}{\bf for~all~~}
\framebox[34mm][l]{~
$\mu_{(j,s)}^{[r-1]} \in {\bm \mu}_j^{[r-1]}$~~~{\small (I\hspace{-.1em}I)}
}~~{\bf do}

\vspace*{0.5mm}
\noindent\hspace*{19mm}
\framebox[29mm][l]{~${\bm \mu}_j^{[r-1]}\in {\cal M}$~~~{\small (I\hspace{-.1em}I')}
}

\vspace*{0.5mm}
\noindent\hspace*{19mm}
\framebox[50mm][l]{~
$\rho_j \leftarrow \rho_j +v_{(i,h)}\!\times\! \mu_{(j,s)}$~~~~
{\small (I\hspace{-.1em}V)}
}\: \\

\noindent
The update step is modified from that of the spherical $k$-means algorithm shown in 
{\bf Algorithm}~\ref{algo:sph} for the use of object feature vectors with sparse expression.
We can evaluate the effect of the inverted-file data structure {\em itself} 
on the speed performance by comparing {\em MFN} with the following {\em IFN}.

Both {\em IFN} and {\em IFB} utilize an inverted-file data structure 
with full expressions for the means, which resembles that 
in Fig.~\ref{fig:str}(b) right top. 
The inverted file has all the $k$ entries for each term 
while {\em IVF} has $(nc)_s\!\leq\!k$ entries 
for a term whose global term ID is $s$.
Then
lines~9~(I\hspace{-.08em}I), 10~(I\hspace{-.08em}I'), and 12~(I\hspace{-.08em}V) 
in {\bf Algorithm~\ref{algo:ours}} are replaced with\\

\vspace*{-0.8mm}
\noindent\hspace*{8mm}{\bf for~all~~}
\framebox[42mm][l]{~
$(u_{(s,j)})^{[r-1]} \in \bar{\bm{\xi}}_s^{[r-1]}$~~~{\small (I\hspace{-.1em}I)}
}~~{\bf do}

\vspace*{0.5mm}
\noindent\hspace*{19mm}
\framebox[55mm][l]{~
$\bar{\bm{\xi}}_s^{[r\!-\!1]}\!=\![(u_{(s,j)})_{j=1}^k]^{[r-1]}\in \bar{\cal M}$~
{\small (I\hspace{-.1em}I')}
}

\vspace*{0.5mm}
\noindent\hspace*{19mm}
\framebox[52mm][l]{~
$\rho_j \leftarrow \rho_j +v_{(i,h)}\!\times\! u_{(s,j)}$~~~~
{\small (I\hspace{-.1em}V)}
}\: ,\\

\noindent
where $\bar{\cal M}$ indicates the set of the mean feature vectors 
represented by the inverted-file data structure 
with all the $k$ entries for each of the $D$ terms and 
$\bar{\bm{\xi}}_s$ denotes the value array of the $s$-th term.
Mean sets $\breve{\cal M}^{[r-1]}$ and $\breve{\cal M}^{[r]}$ 
in {\bf Algorithm}~\ref{algo:ours} are replaced with 
$\bar{\cal M}^{[r-1]}$ and $\bar{\cal M}^{[r]}$.
The undefined values in $\bar{\bm{\xi}}_s$ are padded with zeros.
Then the similarity between $\hat{\bm x}_i$ and the $j$-th centroid (mean) 
is expressed by 
\begin{equation}
\rho_j = \sum_{h=1}^{(nt)_i} v_{(i,h)}\!\times\! u_{(s,j)},
\hspace{3mm} s = t_{(i,h)}\: .
\end{equation}
The difference between {\em IFB} and {\em IFN} is concerned with 
whether the zero multiplications in the partial similarity calculations 
are skipped, based on the conditional branch statement of\\

\noindent\hspace*{8mm}
\framebox[56mm][l]{~
{\bf if}~~$u_{(s,j)}\!=\!0$~~{\bf then}~~go~to~line~9~~
{\small (I\hspace{-.1em}I\hspace{-.1em}I)}
}\: ,\\

\noindent which is inserted at line~11 (I\hspace{-.08em}I\hspace{-.08em}I) 
in {\bf Algorithm~\ref{algo:ours}}.
The algorithm with the conditional branch is {\em IFB} and 
the other is {\em IFN}. 

Using the conditional branch at 
(I\hspace{-.1em}I\hspace{-.1em}I) in {\bf Algorithm~\ref{algo:ours}} 
has an advantage and a disadvantage.
The advantage is the decrease of the number of costly operations related to 
floating-point multiplications and additions at line~12
in {\bf Algorithm~\ref{algo:ours}}.
The disadvantage is the increase of the numbers of both instructions 
and branch mispredictions.
Comparing {\em IFB} and {\em IFN} in Section~\ref{sec:exp} 
explains the impact of branch mispredictions on speed performance.

Let us briefly review the relationship among the four algorithms.
Consider {\em IFN} as a baseline algorithm.
The difference of {\em MFN} and {\em IFN} is only in their 
standard and inverted-file data structures. 
The difference of {\em IFB} and {\em IFN} is only in how to process 
the zero multiplications, whether they are skipped by
the inserted conditional branch or calculated 
without the conditional-branch insertion.
The difference of {\em IVF} and {\em IFN} is only in their mean expressions:
sparse or full.

Note that {\em MFB} and {\em TWM} in Table~\ref{table:class} were not compared.
The {\em MFB} performance can be estimated by the comparison results 
of {\em IFB} and {\em IFN}.
{\em TWM} was prepared as an algorithm for both the object and the mean feature vectors
represented by the standard data structure with a sparse expression.
To calculate the similarity of $\hat{\bm x}_i$ and centroid feature vector $\hat{\bm \mu}_j$,
the feature values with identical global term IDs have to be detected in both the vectors,
i.e., the set-intersection operation in terms of global term ID has to be executed.
{\em TWM} uses a {\em two-way merge} algorithm for the set-intersection operation 
\cite{knuth}.
{\em TWM}, which has many conditional branches that induce cache misses, 
operated very slowly in our preliminary experiments 
based on identical settings as the others.

\section{Experiments}\label{sec:exp}
We describe data sets, a platform for executing the algorithms,
and the performance of the four algorithms, 
our proposed {\em IVF} and three others in Section~\ref{sec:algos}.

\begin{figure}[t]
\begin{center}\hspace*{1mm}
	\begin{tabular}{cc}	
	\subfigure{
		\psfrag{K}[c][c][0.9]{Number of clusters: $k$ ($\times 10^3$)}
		\psfrag{W}[c][c][0.9]{
			\begin{picture}(0,0)
				\put(0,0){\makebox(0,20)[c]{Max. memory size (GB)}}
			\end{picture}
		}
		\psfrag{O}[c][c][0.82]{$0$}
		\psfrag{P}[c][c][0.82]{$5$}
		\psfrag{Q}[c][c][0.82]{$10$}
		\psfrag{R}[c][c][0.82]{$15$}
		\psfrag{S}[c][c][0.82]{$20$}
		\psfrag{X}[r][r][0.82]{$5$}
		\psfrag{Y}[r][r][0.82]{$10$}
		\psfrag{Z}[r][r][0.82]{$15$}
		\psfrag{U}[r][r][0.82]{$20$}
		\psfrag{T}[r][r][0.82]{$25$}
		\psfrag{A}[r][r][0.65]{\em IVF}
		\psfrag{B}[r][r][0.65]{\em IFN}
		\psfrag{C}[r][r][0.65]{\em IFB}
		\psfrag{D}[r][r][0.65]{\em MFN}
		\includegraphics[width=40mm]{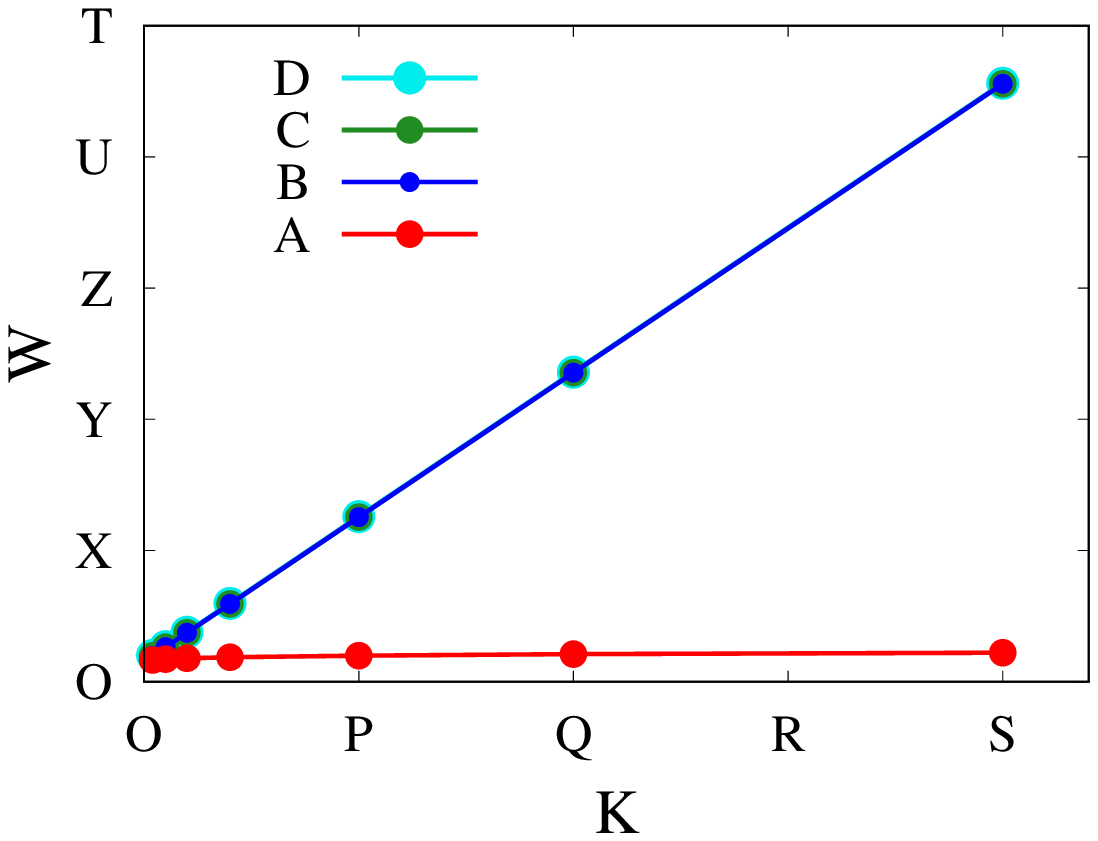}
	} &
	\subfigure{
		\psfrag{K}[c][c][0.9]{Number of clusters: $k$ ($\times 10^3$)}
		\psfrag{W}[c][c][0.9]{
			\begin{picture}(0,0)
				\put(0,0){\makebox(0,20)[c]{Max. memory size (GB)}}
			\end{picture}
		}
		\psfrag{O}[c][c][0.82]{$0$}
		\psfrag{P}[c][c][0.82]{$5$}
		\psfrag{Q}[c][c][0.82]{$10$}
		\psfrag{R}[c][c][0.82]{$15$}
		\psfrag{S}[c][c][0.82]{$20$}
		\psfrag{X}[r][r][0.82]{$20$}
		\psfrag{Y}[r][r][0.82]{$40$}
		\psfrag{Z}[r][r][0.82]{$60$}
		\psfrag{U}[r][r][0.82]{$80$}
		\psfrag{A}[r][r][0.65]{\em IVF}
		\psfrag{B}[r][r][0.65]{\em IFN}
		\psfrag{C}[r][r][0.65]{\em IFB}
		\psfrag{D}[r][r][0.65]{\em MFN}
		\includegraphics[width=40mm]{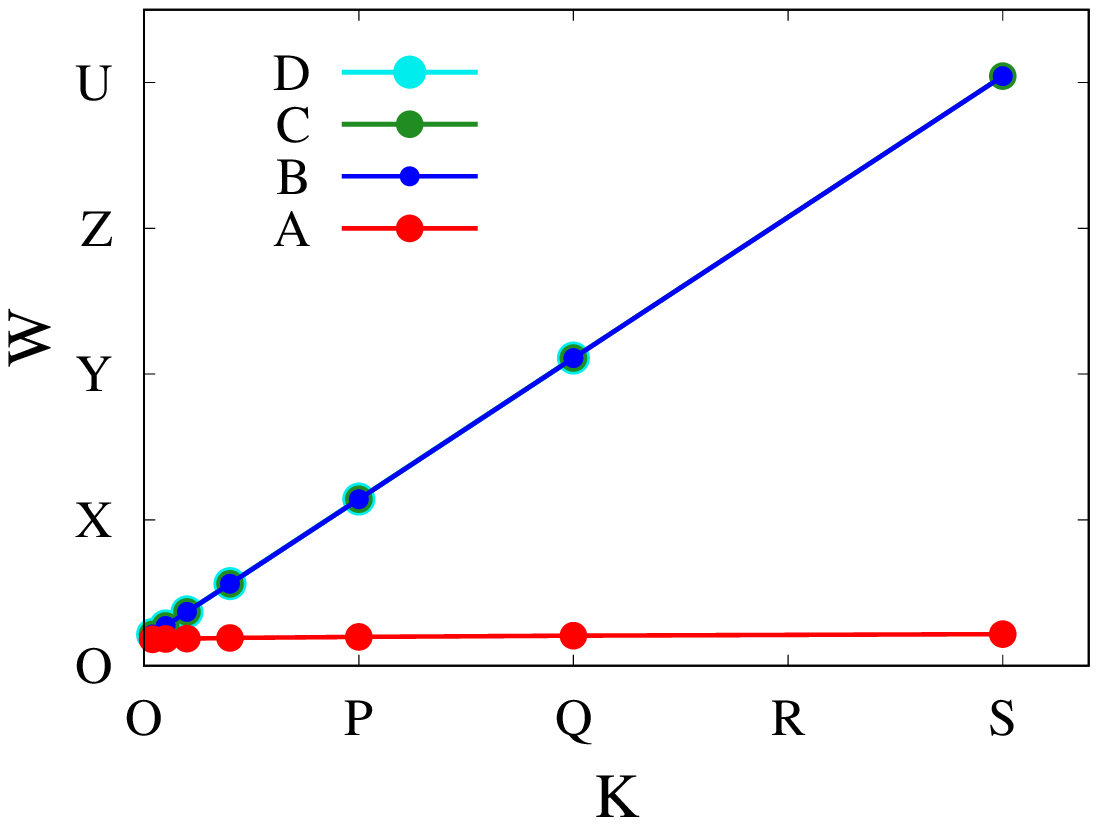}
	}\tabularnewline
	(a)~PubMed & (b)~NYT \tabularnewline
	\end{tabular}
\end{center}
\vspace*{-3mm}
\caption{Maximum memory size through iterations where all four algorithms 
were required. Occupied memory size was plotted along $k$ values 
when each algorithm was applied to (a) PubMed and (b) NYT.
}
\label{fig:mem}
\vspace*{-2mm}
\end{figure}

\subsection{Data Sets}\label{subsec:data}
We employed two different types of large-scale and high-dimensional 
sparse real document data sets: 
{\em PubMed Abstracts} (PubMed for short) \cite{pubmed}
and {\em The New York Times Articles} (NYT).

The PubMed data set contains 8,200,000 documents (texts) 
represented by the term (distinct word) counts in each.
We made a feature vector normalized by its $L_2$ norm from each document,
each of which consisted of the {\em tf-idf} values of the corresponding terms.
Each feature vector was regarded as a point on a unit hypersphere.
We chose 1,000,000 feature vectors at random without duplication 
from all of the vectors as our 1M-sized experimental data sets.
The number of distinct terms in the data set 
(dimensionality) was 140,914. 
Their average frequency in the documents, i.e., 
the average number of non-zero elements in the feature vectors, was 58.95, 
and the average sparsity in Eq.~(\ref{eq:avg_sparsity}) was $3.93\!\times\!10^{-4}$.

We extracted 1,285,944 articles from {\em The New York Times Articles} 
from 1994 to 2006 
and counted the frequency of the term occurrences 
after stemming and stop word removal.
In the same manner as PubMed, we made a set of feature vectors
with 495,714 dimensionality.
The average number of non-zero elements in the feature vectors was
225.76, corresponding to an average sparsity of $4.56\!\times\!10^{-4}$.

\subsection{Platform and Measures}\label{subsec:platform}
All the algorithms were executed on a computer system, 
which was equipped with two Xeon E5-2697v3 2.6-GHz CPUs 
with three-level caches from levels 1 to level 3 \cite{hammarlund} 
and a 256-GB main memory, 
by a single thread on a single process within the memory capacity.
When the algorithms were executed, 
two hardware prefetchers related to the level-2 caches in the CPU 
were disabled by BIOS control \cite{intel} 
to measure the effect of the cache misses themselves.
The algorithms were implemented in C and compiled with 
the GNU compiler collection (gcc) version 8.2.0 on the optimization level of 
{\sf -O0}. 
The performances of the algorithms were evaluated with 
CPU time (or clock cycles) until convergence and 
the maximum physical memory size occupied through the iterations.

\begin{figure}[t]
\begin{center}\hspace*{1mm}
	\begin{tabular}{cc}	
	\subfigure{
		\psfrag{X}[c][c][0.93]{
			\begin{picture}(0,0)
				\put(0,0){\makebox(0,-6)[c]{Iteration} }
			\end{picture}
		}
		\psfrag{Y}[c][c][0.88]{
			\begin{picture}(0,0)
				\put(0,0){\makebox(0,16)[c]{Number of terms ($\times 10^4$)} }
			\end{picture}
		}
		\psfrag{A}[c][c][0.82]{$0$}
		\psfrag{B}[c][c][0.82]{$50$}
		\psfrag{C}[c][c][0.82]{$100$}
		\psfrag{D}[c][c][0.82]{$150$}
		\psfrag{E}[c][c][0.82]{$200$}
		\psfrag{F}[r][r][0.82]{$1$}
		\psfrag{G}[r][r][0.82]{$2$}
		\psfrag{H}[r][r][0.82]{$3$}
		\psfrag{a}[l][l][0.62]{$20000$}
		\psfrag{b}[l][l][0.62]{$10000$}
		\psfrag{c}[l][l][0.62]{$5000$}
		\psfrag{d}[l][l][0.62]{$2000$}
		\psfrag{e}[l][l][0.62]{$1000$}
		\psfrag{f}[l][l][0.62]{$500$}
		\psfrag{g}[c][c][0.62]{$200$}
		\psfrag{h}[c][c][0.7]{$k$}
		\includegraphics[width=40mm]{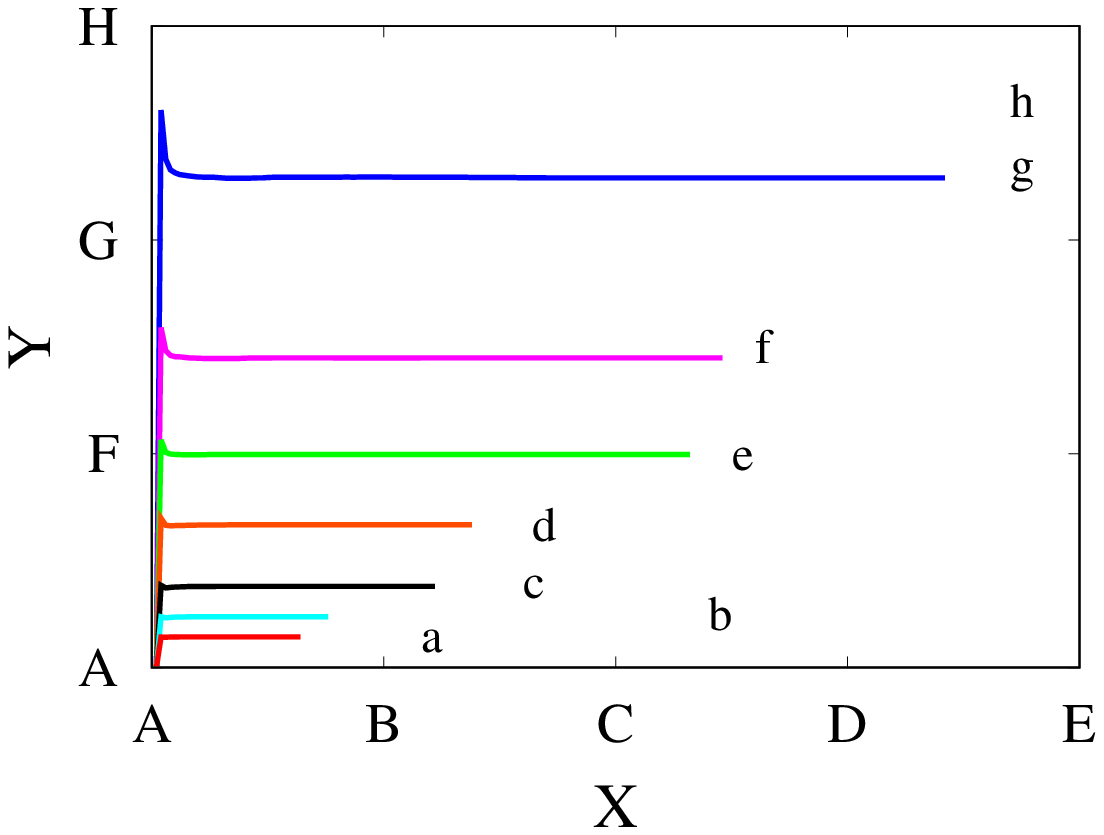}
	} &
	\subfigure{
		\psfrag{X}[c][c][0.9]{Number of clusters: $k$}
		\psfrag{Y}[c][c][0.9]{
			\begin{picture}(0,0)
				\put(0,0){\makebox(0,22)[c]{Number of terms}}
			\end{picture}
		}
		\psfrag{S}[r][r][0.68]{At convergence}
		\psfrag{T}[r][r][0.68]{Maximum}
		\psfrag{A}[c][c][0.82]{$10^2$}
		\psfrag{B}[c][c][0.82]{$10^3$}
		\psfrag{C}[c][c][0.82]{$10^4$}
		\psfrag{P}[r][r][0.82]{$10^3$}
		\psfrag{Q}[r][r][0.82]{$10^4$}
		\psfrag{R}[r][r][0.82]{$10^5$}
		\includegraphics[width=40mm]{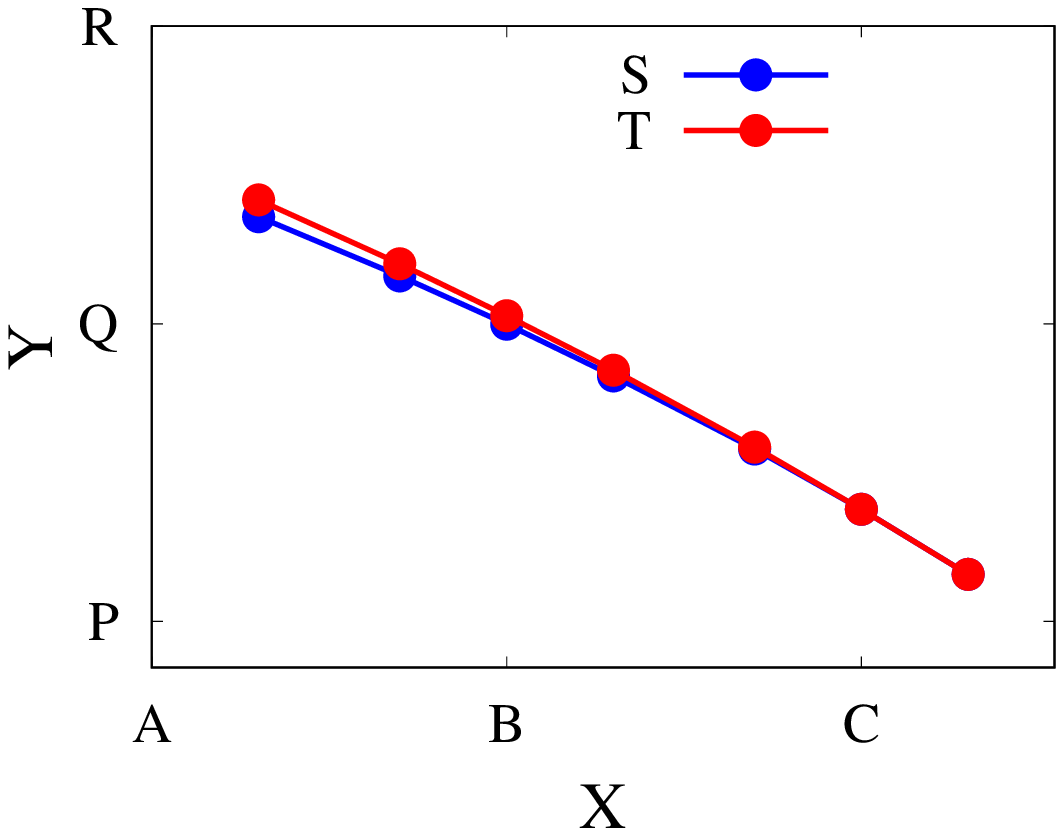}
	}\tabularnewline
	(a)~Avg. \#terms in means  & (b)~\#terms in means \tabularnewline
	\end{tabular}
\end{center}
\vspace*{-3mm}
\caption{Number of distinct terms in means 
when all algorithms were applied to PubMed: 
(a) Averages of terms in all means for each iteration 
when various $k$ values are illustrated in linear scale.
(b) Maximum of avg. \#terms in means per iteration and 
avg. \#terms at convergence are done along $k$ in log-log scale.
}
\label{fig:mterms}
\vspace*{-2mm}
\end{figure}

\begin{figure}[t]
\begin{center}\hspace*{1mm}
	\psfrag{X}[c][c][0.9]{Number of clusters: $k$}
	\psfrag{Y}[c][c][0.9]{
		\begin{picture}(0,0)
			\put(0,0){\makebox(0,22)[c]{Max. memory size (MB)}}
		\end{picture}
	}
	\psfrag{A}[c][c][0.82]{$10^2$}
	\psfrag{B}[c][c][0.82]{$10^3$}
	\psfrag{C}[c][c][0.82]{$10^4$}
	\psfrag{P}[r][r][0.82]{$10^1$}
	\psfrag{Q}[r][r][0.82]{$10^2$}
	\psfrag{R}[r][r][0.82]{$10^3$}
	\includegraphics[width=50mm]{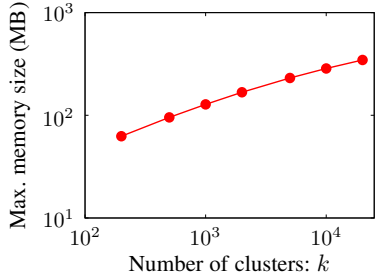}
\end{center}
\vspace*{-4mm}
\caption{Maximum memory capacitance for mean tuples 
required by {\em IVF} through iteration in PubMed.
}
\label{fig:mmem}
\vspace*{-2mm}
\end{figure}

\subsection{Performance}\label{subsec:perform}
\subsubsection{Required Maximum Memory Size}
We measured the maximum memory size required by the algorithms through the iterations
until the convergence (Fig.~\ref{fig:mem}).
The four algorithms represented the object data set with the sparse expression
of the tuple $(t_{(i,h)},v_{(i,h)})$ shown in Sections~\ref{sec:prop} and \ref{sec:algos}.
As types of elements $t_{(i,h)}$ and $v_{(i,h)}$, 
an integer (int) and a 64-bit floating point (double) were used\footnote{
The tuple was not implemented with a {\em structure type} consisting of 
an int-type and a double-type member 
to avoid unnecessary memory usage caused by an 8-byte memory alignment 
adopted by 64-bit CPUs. 
}.
The memory capacitance occupied by the object set is expressed by 
$(\sum_{i=1}^N (nt)_i)\!\times\!(\mbox{sizeof(int + double)})$.
Those of PubMed and NYT were 706.8 MB and 3,484 MB.

\begin{figure}[t]
\begin{center}\hspace*{1mm}
	\begin{minipage}[h]{60mm}
	\centering
		\psfrag{K}[c][c][0.9]{Number of clusters: $k$}
		\psfrag{W}[c][c][0.9]{
			\begin{picture}(0,0)
				\put(0,0){\makebox(0,22)[c]{Avg. CPU time (sec)}}
			\end{picture}
		}
		\psfrag{P}[c][c][0.82]{$10^2$}
		\psfrag{Q}[c][c][0.82]{$10^3$}
		\psfrag{R}[c][c][0.82]{$10^4$}
		\psfrag{S}[c][c][0.82]{}
		\psfrag{X}[r][r][0.82]{$10^1$}
		\psfrag{Y}[r][r][0.82]{$10^2$}
		\psfrag{Z}[r][r][0.82]{$10^3$}
		\psfrag{U}[r][r][0.82]{$10^4$}
		\psfrag{T}[r][r][0.82]{$10^5$}
		\psfrag{A}[r][r][0.7]{\em IVF}
		\psfrag{B}[r][r][0.7]{\em IFN}
		\psfrag{C}[r][r][0.7]{\em IFB}
		\psfrag{D}[r][r][0.7]{\em MFN}
		\includegraphics[width=53mm]{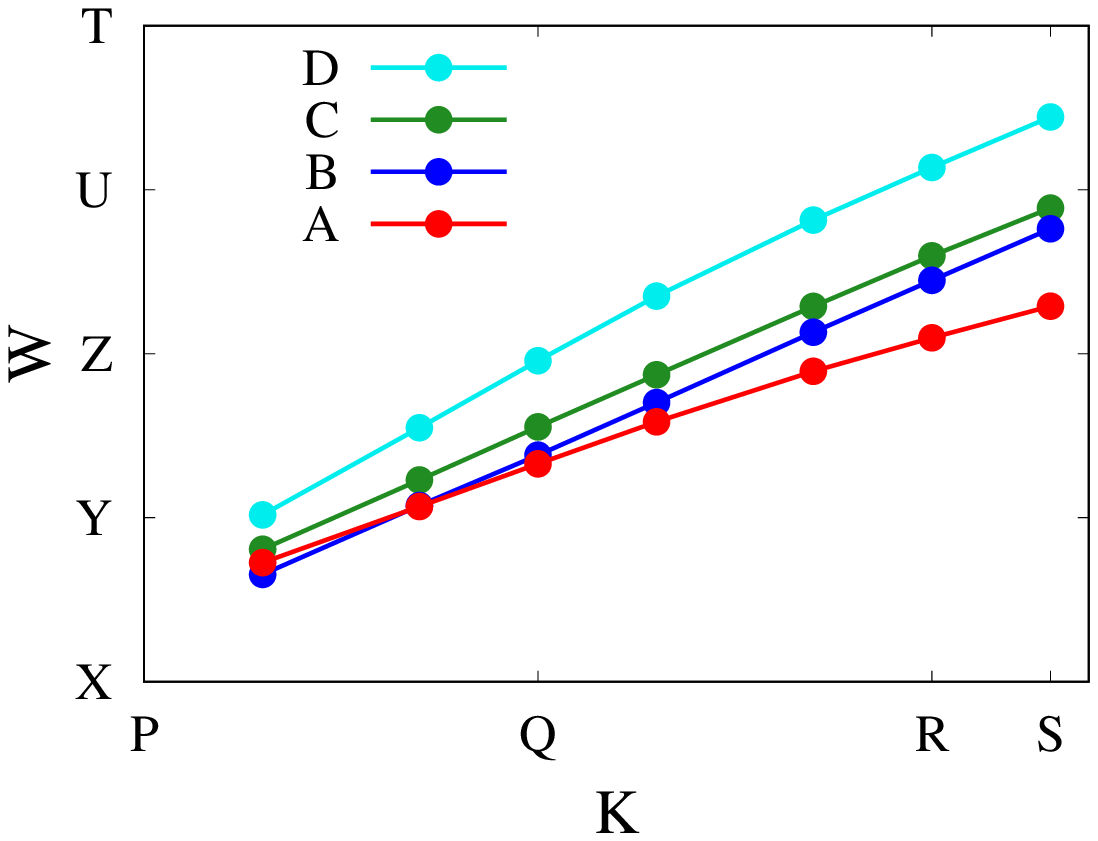}\\
		(a)~PubMed
	\end{minipage}\\ \vspace*{1mm}
	\begin{minipage}[h]{60mm}
	\centering
		\psfrag{K}[c][c][0.9]{Number of clusters: $k$}
		\psfrag{W}[c][c][0.9]{
			\begin{picture}(0,0)
				\put(0,0){\makebox(0,22)[c]{Avg. CPU time (sec)}}
			\end{picture}
		}
		\psfrag{P}[c][c][0.82]{$10^2$}
		\psfrag{Q}[c][c][0.82]{$10^3$}
		\psfrag{R}[c][c][0.82]{$10^4$}
		\psfrag{S}[c][c][0.82]{}
		\psfrag{X}[r][r][0.82]{$10$}
		\psfrag{Y}[r][r][0.82]{$10^2$}
		\psfrag{Z}[r][r][0.82]{$10^3$}
		\psfrag{U}[r][r][0.82]{$10^4$}
		\psfrag{T}[r][r][0.82]{$10^5$}
		\psfrag{A}[r][r][0.7]{\em IVF}
		\psfrag{B}[r][r][0.7]{\em IFN}
		\psfrag{C}[r][r][0.7]{\em IFB}
		\psfrag{D}[r][r][0.7]{\em MFN}
		\includegraphics[width=53mm]{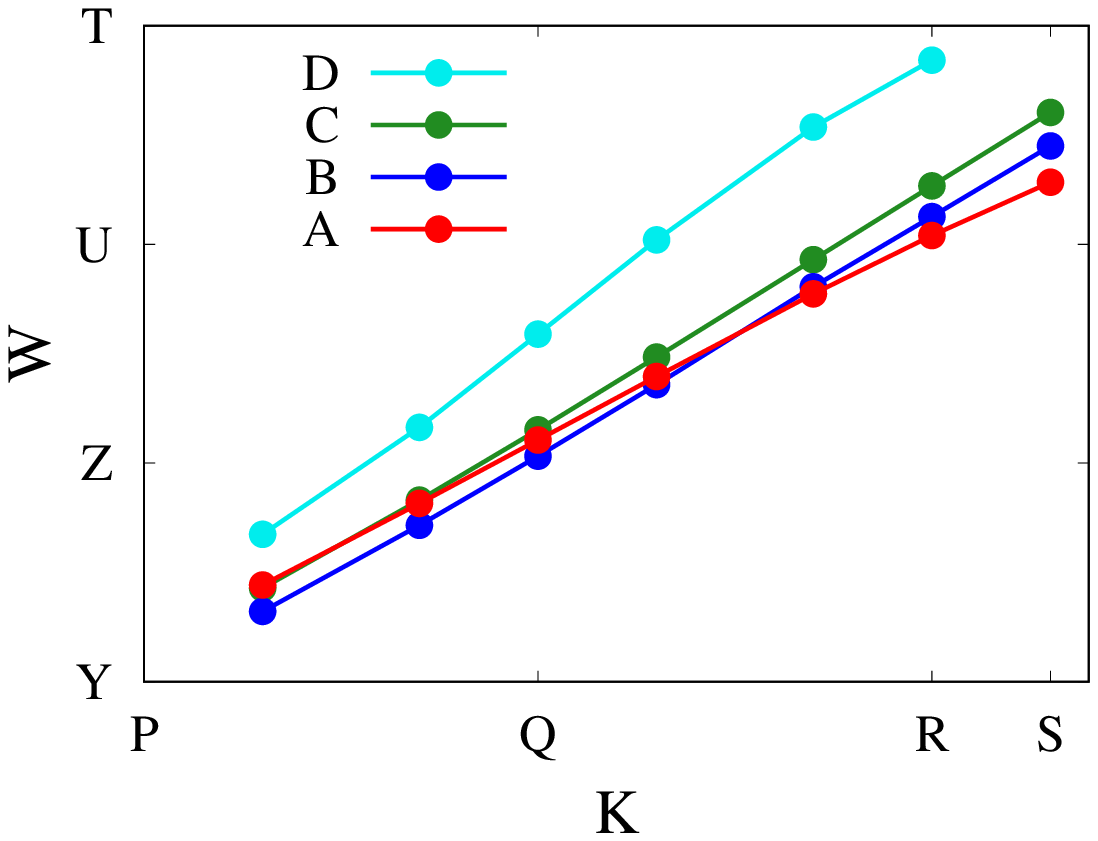}\\
		(b)~NYT
	\end{minipage}
\end{center}
\vspace*{-3mm}
\caption{Average CPU time per iteration required by 
each of four algorithms.
CPU time was plotted along $k$ with log-log scale 
when each algorithm was applied to (a) PubMed and (b) NYT.
}
\label{fig:time}
\vspace*{-2mm}
\end{figure}

By contrast, the memory capacitance for the mean set depends on
the algorithms and the number of means $k$.
The three algorithms with full expressions ({\em MFN}, {\em IFB}, and {\em IFN}) 
need an identical memory capacitance 
expressed by
$k\!\times\! D\!\times\! (\mbox{sizeof(double)})$, 
where $D$ denotes not the number of distinct terms in the mean feature vectors
but the dimensionality, including zero padding.
The memory capacitances for PubMed and NYT were 
$1.20\!\times\!k$ MB and $3.96\!\times\!k$ MB 
and reached 24.0 GB and 79.2 GB at $k\!=\!20,000$.

The memory capacitance required by {\em IVF} 
for the mean feature vectors is expressed by 
$(\sum_{p=1}^{D} (nc)_p)\!\times\! (\mbox{sizeof(int + double)})$, which 
is equivalent to 
$(\sum_{j=1}^k (ntm)_j)\!\times\! (\mbox{sizeof(int + double)})$, 
where $(ntm)_j$ denotes the number of distinct terms 
in the $j$-th mean feature vector.
Figures~\ref{fig:mterms}(a) and (b) show $(\sum_{j=1}^k (ntm)_j)/k$ for each iteration
when {\em IVF} started at the initial state chosen randomly in PubMed 
and both the maximum $(\sum_{j=1}^k (ntm)_j)/k$ through iterations and 
that at the convergence.
As shown in Fig.~\ref{fig:mterms}(a),
the average number of mean terms became stable after several iterations 
for each $k$ value.
Figure~\ref{fig:mterms}(b) indicates that
the maximum average number almost coincided with the average number at the convergence,
and both numbers decreased as 
a power-law function of $k$. 
Using the maximum average number of mean terms in Fig.~\ref{fig:mterms},
the maximum memory capacitance that {\em IVF} needed was calculated 
with various $k$ for PubMed.
Figure~\ref{fig:mmem} shows that the memory size increased as a sublinear function of $k$,
and even when $k\!=\!20,000$, the memory size was only 345.7 MB.

Thus by applying the sparse expressions to a sparse data set 
we significantly reduced the memory capacitance occupied by 
the object and mean feature vectors.

\subsubsection{CPU Time}
Figures~\ref{fig:time}(a) and (b) show the average CPU times per iteration 
with $k$ in the log-log scale required by 
the four algorithms until the convergence, 
when they were applied to PubMed and NYT. 
Regarding the speed performance in the two distinct data sets,
the relationships among the algorithms were almost the same. 
{\em IVF} achieved the best performance in the range of large $k$ values.
When $k\!=\!20,000$ in PubMed shown in Fig.~\ref{fig:time}(a),
the CPU time of {\em IVF} was only $33.7\%$ of {\em IFN} (the second best).
% IVF: 1.945447e+03, IFN: 5.776748e+03 --> rate=0.336772
By contrast, both algorithms were competitive in the small $k$ range. 
These performances are analyzed in Section~\ref{sec:analy} and 
scrutinized in Section~\ref{subsec:ifn}.

{\em MFN} needed much more CPU time than the others that employed 
the inverted-file data structure.
The CPU time for PubMed reached $4.89$ times more than 
that of {\em IFN}, 
which only differs from {\em MFN} in the mean data structure, 
whether it is the inverted-file or the standard,
as described in Section~\ref{sec:algos}.
This actually indicates that the inverted-file data structure is 
useful for a large-scale sparse data set.

Our comparison of {\em IFB} and {\em IFN} was interesting.
It intuitively seems that {\em IFB}, which skips 
costly unnecessary floating-point multiplications using the conditional branch,
operates faster than {\em IFN} that directly executes the multiplications.
Surprisingly, 
{\em IFN} was faster than {\em IFB} in every range of $k$ 
in both data sets.
{\em IFB} required $1.28$ to $1.49$ times more CPU time than {\em IFN}.
Executing the conditional branch many times,
e.g., in the innermost loop of the triple loop, risks 
degrading the speed performance.

\section{Performance Analysis}\label{sec:analy}
\begin{figure}[t]
\begin{center}\hspace*{1mm}
	\begin{tabular}{cc}	
	\subfigure{
		\psfrag{K}[c][c][0.9]{Number of clusters: $k$}
		\psfrag{W}[c][c][0.9]{
			\begin{picture}(0,0)
				\put(0,0){\makebox(0,28)[c]{Avg. \# instructions}}
			\end{picture}
		}
		\psfrag{P}[c][c][0.82]{$10^2$}
		\psfrag{Q}[c][c][0.82]{$10^3$}
		\psfrag{R}[c][c][0.82]{$10^4$}
		\psfrag{S}[c][c][0.82]{}
		\psfrag{T}[r][r][0.82]{$10^{11}$}
		\psfrag{X}[r][r][0.82]{$10^{12}$}
		\psfrag{Y}[r][r][0.82]{$10^{13}$}
		\psfrag{Z}[r][r][0.82]{$10^{14}$}
		\psfrag{A}[r][r][0.65]{\em IVF}
		\psfrag{B}[r][r][0.65]{\em IFN}
		\psfrag{C}[r][r][0.65]{\em IFB}
		\psfrag{D}[r][r][0.65]{\em MFN}
		\includegraphics[width=40mm]{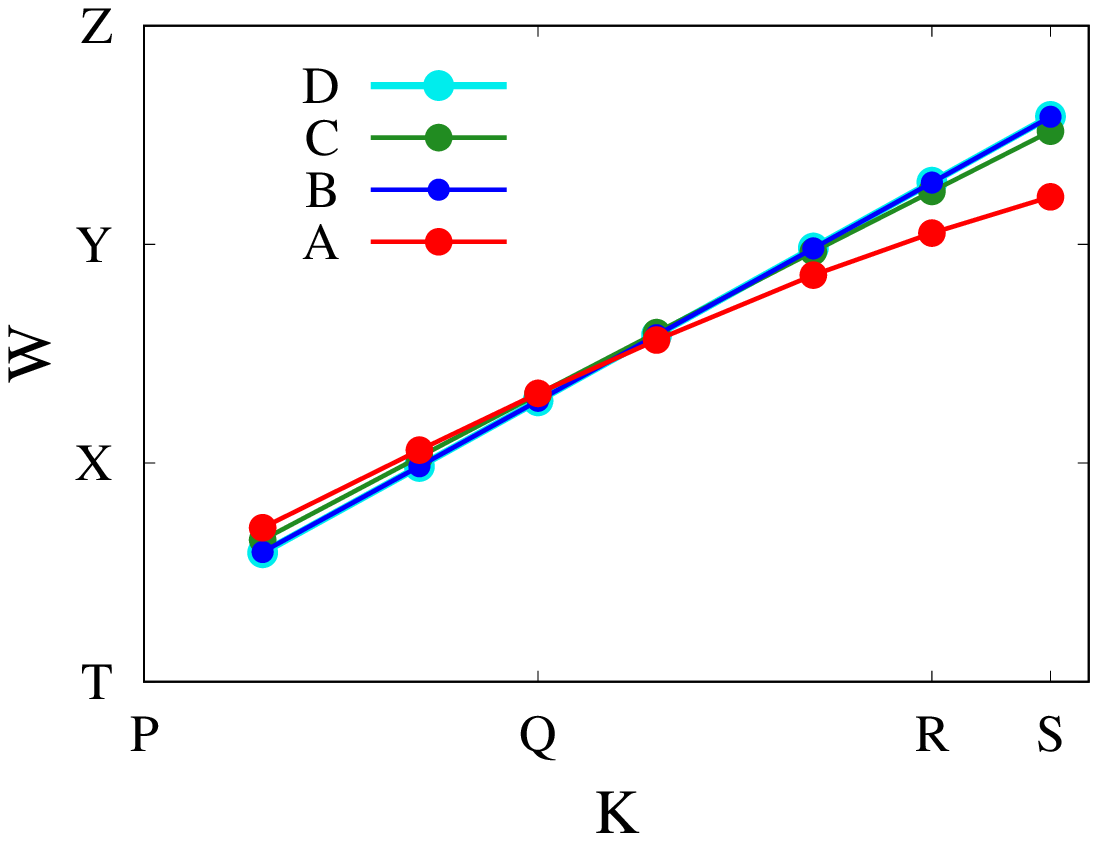}
	} &
	\subfigure{
		\psfrag{K}[c][c][0.9]{Number of clusters: $k$}
		\psfrag{W}[c][c][0.9]{
			\begin{picture}(0,0)
				\put(0,0){\makebox(0,28)[c]{Avg. \# instructions}}
			\end{picture}
		}
		\psfrag{P}[c][c][0.82]{$10^2$}
		\psfrag{Q}[c][c][0.82]{$10^3$}
		\psfrag{R}[c][c][0.82]{$10^4$}
		\psfrag{S}[c][c][0.82]{}
		\psfrag{X}[r][r][0.82]{$10^{12}$}
		\psfrag{Y}[r][r][0.82]{$10^{13}$}
		\psfrag{Z}[r][r][0.82]{$10^{14}$}
		\psfrag{U}[r][r][0.82]{$10^{15}$}
		\psfrag{A}[r][r][0.65]{\em IVF}
		\psfrag{B}[r][r][0.65]{\em IFN}
		\psfrag{C}[r][r][0.65]{\em IFB}
		\psfrag{D}[r][r][0.65]{\em MFN}
		\includegraphics[width=40mm]{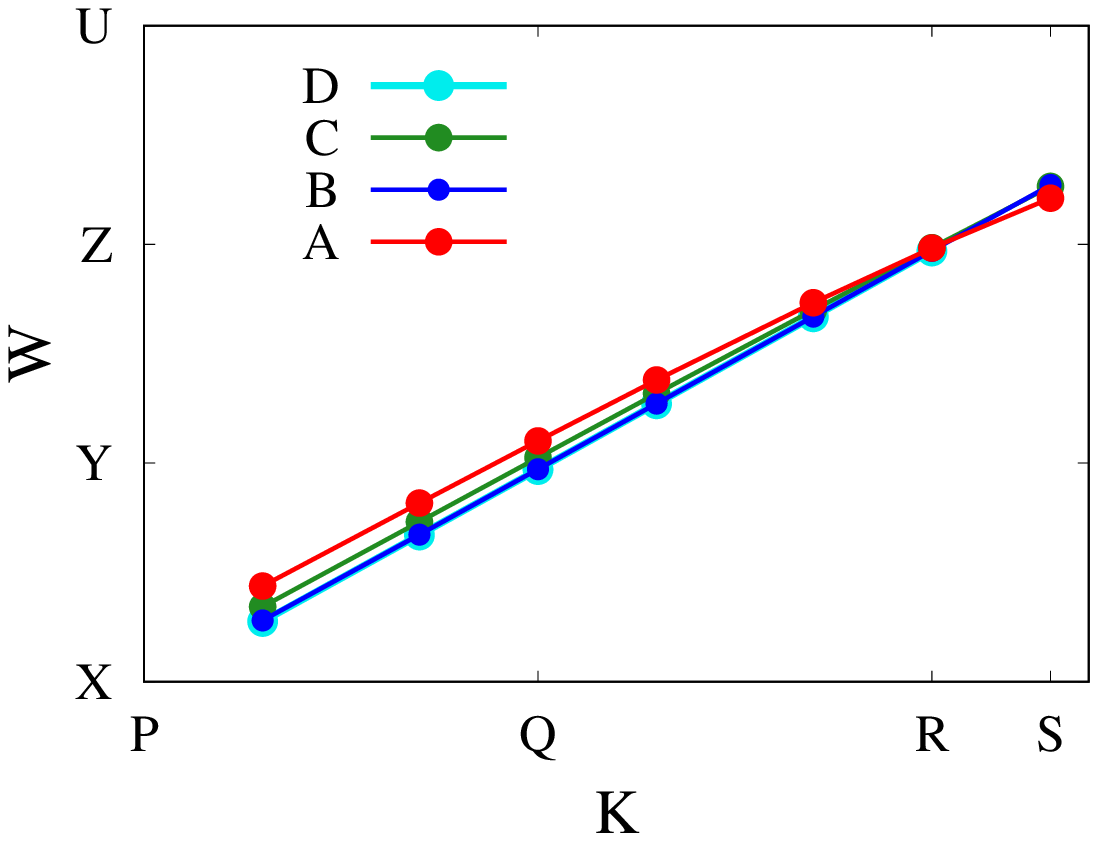}
	}\tabularnewline
	(a)~PubMed & (b)~NYT \tabularnewline
	\end{tabular}
\end{center}
\vspace*{-3mm}
\caption{Average number of instructions per iteration
that each of four algorithms needed with various $k$. 
Algorithms were applied to (a) PubMed and (b) NYT.
}
\label{fig:inst}
\vspace*{-2mm}
\end{figure}

When executing the four algorithms in PubMed and NYT,
we measured the number of completed (retired) instructions, 
cache misses, and 
branch mispredictions with the {\em perf tool} 
(Linux profiling with performance counters) \cite{perf}.
Hereinafter, 
we label the four numbers as follows:
the instructions, the level-1 (L1) data cache misses,
the last-level (LL) cache misses, and the branch mispredictions 
as Inst, L1CM, LLCM, and BM.
These four numbers are collectively called 
{\em performance degradation factors} (DFs).
As they increase, the speed performance worsens.
To estimate the effects of each DF on the total clock cycles (or the CPU time),
we introduced a simple yet practical clock-cycle per instruction (CPI) model
and analyzed the four algorithms based on it.

\subsection{Performance Degradation Factor Characteristics}\label{subsec:dfs}
Figures~\ref{fig:inst}(a) and (b) show the average number of 
completed instructions through iterations until convergence 
when the four algorithms were executed in PubMed and NYT.
The algorithms had almost the same characteristics, 
and their relationships were similar in the distinct data sets 
and shared three characteristic points:
\begin{enumerate}
\item 
The number of instructions of {\em MFN} coincided with that of {\em IFN}.
The rate expressed by 
$| (\mbox{Inst}_{[\mbox{\tiny\em IFN}]}/\mbox{Inst}_{[\mbox{\tiny\em MFN}]})\!-\! 1|$ 
was within 1.1\%.
This fact is adopted as the assumption of the parameter optimization 
in Section~\ref{subsec:cpimodel}.
\item 
$\mbox{Inst}_{[\mbox{\tiny\em IFB}]}$ started at a larger value than 
$\mbox{Inst}_{[\mbox{\tiny\em IFN}]}$ and 
ended at a smaller value at $k\!=\!20,000$.
This is related to the sparsity of the mean feature vectors.
Comparing {\em IFB} with {\em IFN},
$\mbox{Inst}_{[\mbox{\tiny\em IFB}]}$ increased by the insertion of
a conditional branch to avoid unnecessary operations of 
both zero-multiplications and additions at line~12
in {\bf Algorithm~\ref{algo:ours}}.
As the sparsity is lowered, 
i.e., fewer terms appeared in the mean feature vectors,
more instructions related to the multiplications and additions are skipped.
The sparsity became lower with $k$, as shown in Fig.~\ref{fig:mterms}(b).
Thus 
$\mbox{Inst}_{[\mbox{\tiny\em IFB}]}$ and 
$\mbox{Inst}_{[\mbox{\tiny\em IFN}]}$ intersected at a large $k$ value.
\item 
$\mbox{Inst}_{[\mbox{\tiny\em IVF}]}$ had remarkable characteristics 
to $\mbox{Inst}_{[\mbox{\tiny\em IFN}]}$, similar to 
$\mbox{Inst}_{[\mbox{\tiny\em IFB}]}$.
This is discussed in connection with the CPU time in Section~\ref{subsec:ifn}.
\end{enumerate}

\begin{figure}[t]
\begin{center}\hspace*{1mm}
	\begin{tabular}{cc}	
	\subfigure{
		\psfrag{K}[c][c][0.9]{Number of clusters: $k$}
		\psfrag{W}[c][c][0.9]{
			\begin{picture}(0,0)
				\put(0,0){\makebox(0,25)[c]{(Clock cycles)~/~Inst}}
			\end{picture}
		}
		\psfrag{P}[c][r][0.82]{$10^2$}
		\psfrag{Q}[c][c][0.82]{$10^3$}
		\psfrag{R}[c][c][0.82]{$10^4$}
		\psfrag{X}[r][r][0.82]{$0$}
		\psfrag{Y}[r][r][0.82]{$0.5$}
		\psfrag{Z}[r][r][0.82]{$1.0$}
		\psfrag{S}[r][r][0.82]{$1.5$}
		\psfrag{T}[r][r][0.82]{$2.0$}
		\psfrag{A}[r][r][0.65]{\em IVF}
		\psfrag{B}[r][r][0.65]{\em IFN}
		\psfrag{C}[r][r][0.65]{\em IFB}
		\psfrag{D}[r][r][0.65]{\em MFN}
		\includegraphics[width=40mm]{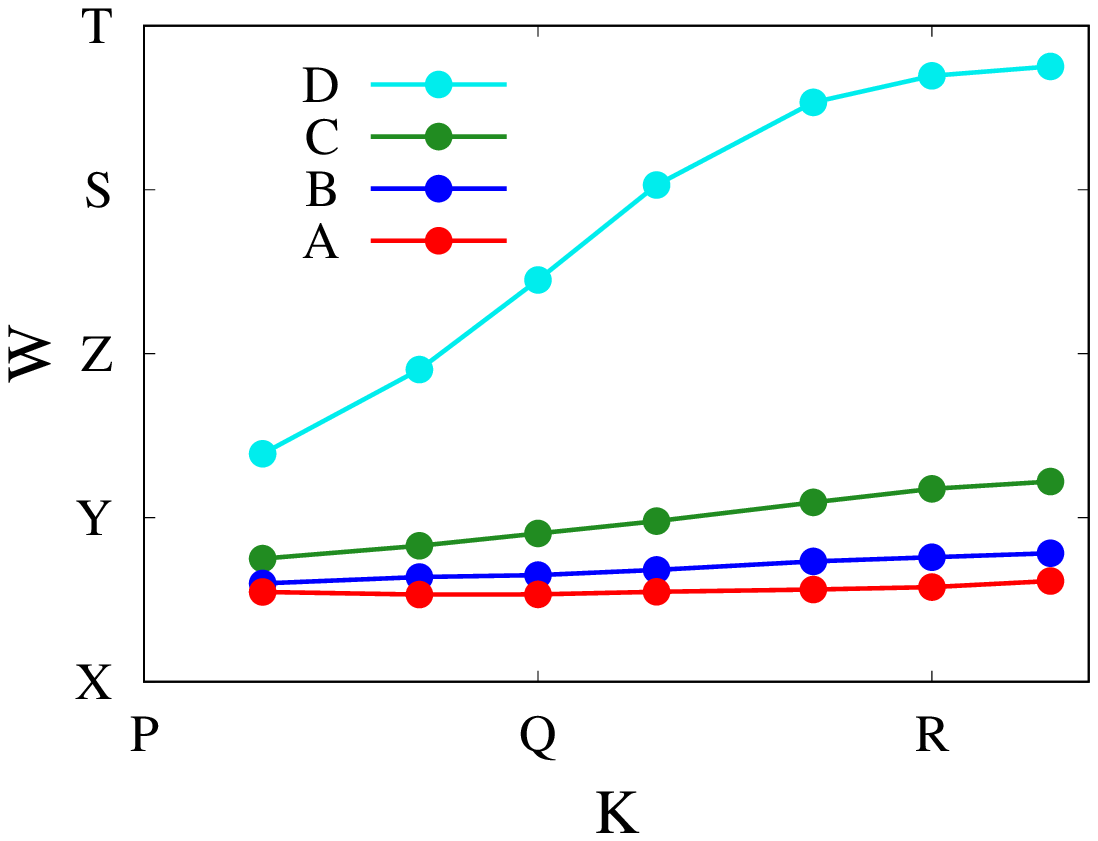}
	} &
	\subfigure{
		\psfrag{K}[c][c][0.9]{Number of clusters: $k$}
		\psfrag{W}[c][c][0.9]{
			\begin{picture}(0,0)
				\put(0,0){\makebox(0,25)[c]{(Clock cycles)~/~Inst}}
			\end{picture}
		}
		\psfrag{P}[c][r][0.82]{$10^2$}
		\psfrag{Q}[c][c][0.82]{$10^3$}
		\psfrag{R}[c][c][0.82]{$10^4$}
		\psfrag{X}[r][r][0.82]{$0$}
		\psfrag{Y}[r][r][0.82]{$0.5$}
		\psfrag{Z}[r][r][0.82]{$1.0$}
		\psfrag{S}[r][r][0.82]{$1.5$}
		\psfrag{T}[r][r][0.82]{$2.0$}
		\psfrag{A}[r][r][0.65]{\em IVF}
		\psfrag{B}[r][r][0.65]{\em IFN}
		\psfrag{C}[r][r][0.65]{\em IFB}
		\psfrag{D}[r][r][0.65]{\em MFN}
		\includegraphics[width=40mm]{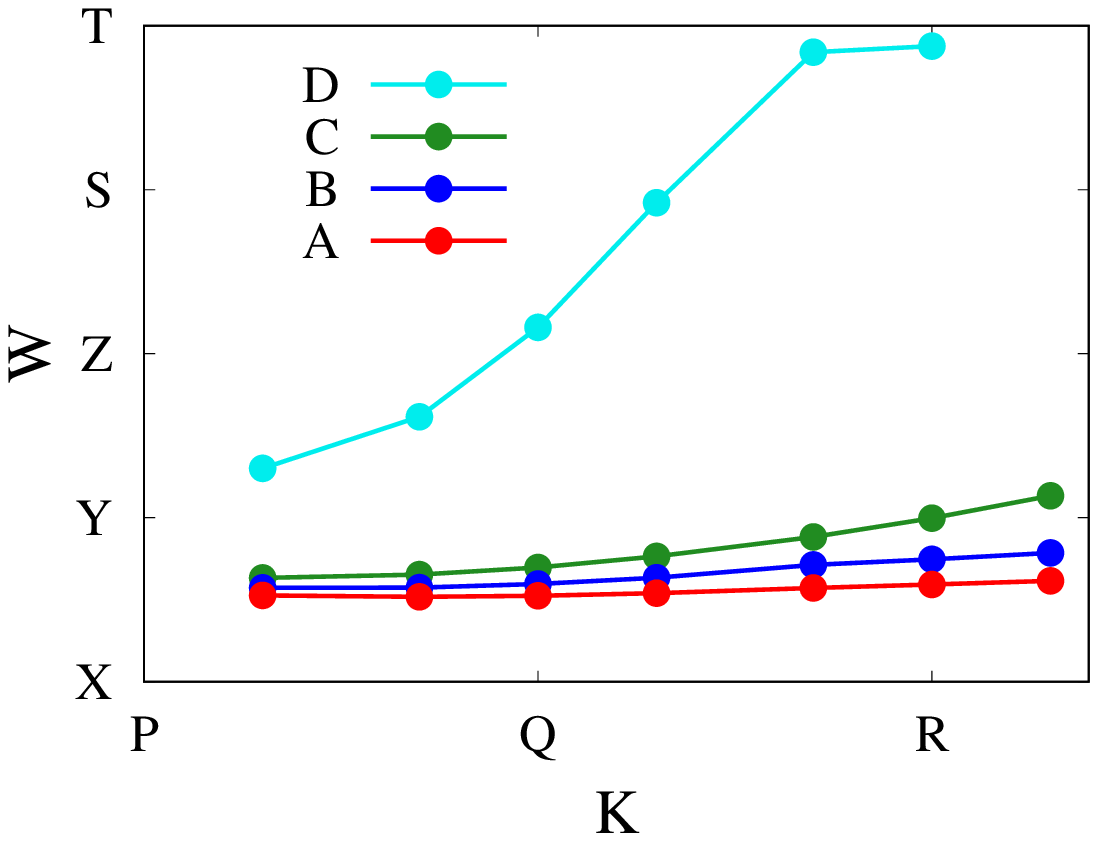}
	}\tabularnewline
	(a)~PubMed & (b)~NYT \tabularnewline
	\end{tabular}
\end{center}
\vspace*{-3mm}
\caption{Actual clock cycles per instruction (CPI)
with various $k$ 
when algorithms were applied to (a) PubMed and (b) NYT
}
\label{fig:clk-inst}
\vspace*{-2mm}
\end{figure}

\begin{figure}[t]
\begin{center}\hspace*{1mm}
	\begin{tabular}{cc}	
	\subfigure{
		\psfrag{K}[c][c][0.9]{Number of clusters: $k$}
		\psfrag{W}[c][c][0.9]{
			\begin{picture}(0,0)
				\put(0,0){\makebox(0,20)[c]{L1CM$'$~/~Inst ($\times 10^{-2}$)}}
			\end{picture}
		}
		\psfrag{P}[c][r][0.82]{$10^2$}
		\psfrag{Q}[c][c][0.82]{$10^3$}
		\psfrag{R}[c][c][0.82]{$10^4$}
		\psfrag{X}[r][r][0.82]{$0$}
		\psfrag{Y}[r][r][0.82]{$1$}
		\psfrag{Z}[r][r][0.82]{$2$}
		\psfrag{S}[r][r][0.82]{$3$}
		\psfrag{T}[r][r][0.82]{$4$}
		\psfrag{U}[r][r][0.82]{$5$}
		\psfrag{A}[r][r][0.65]{\em IVF}
		\psfrag{B}[r][r][0.65]{\em IFN}
		\psfrag{C}[r][r][0.65]{\em IFB}
		\psfrag{D}[r][r][0.65]{\em MFN}
		\includegraphics[width=40mm]{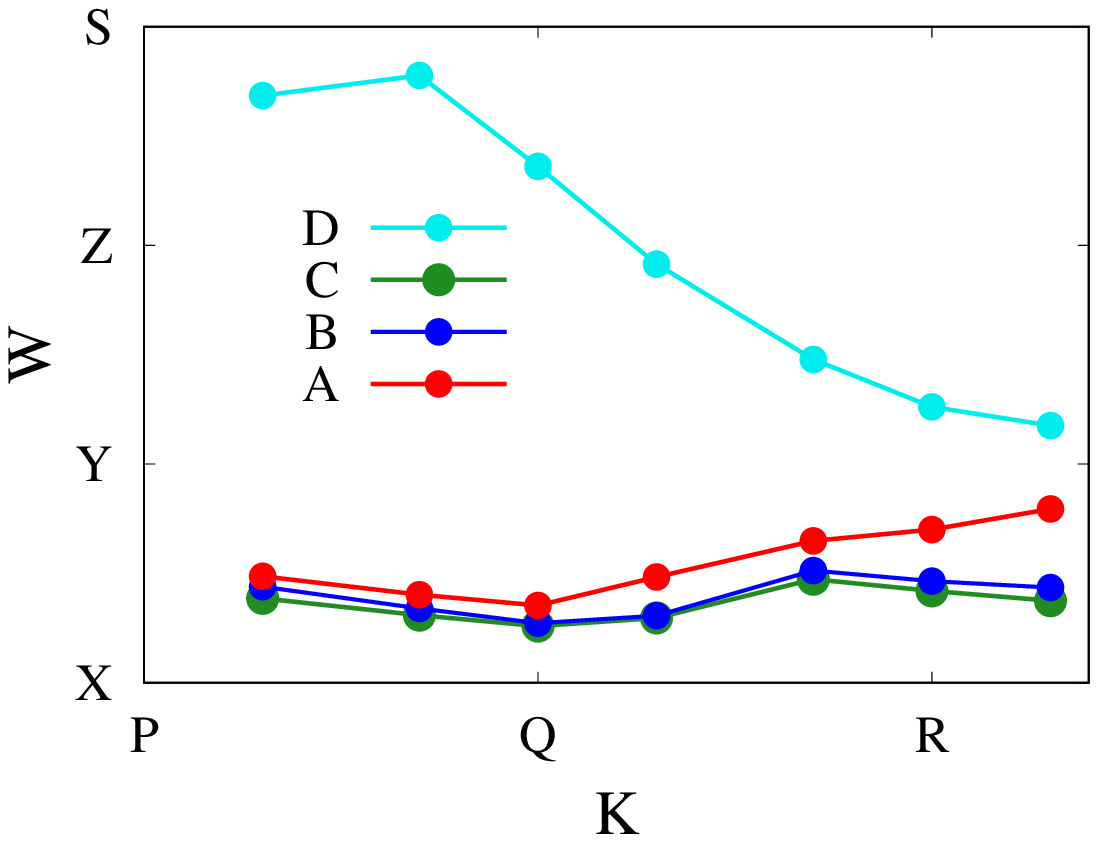}
	} &
	\subfigure{
		\psfrag{K}[c][c][0.9]{Number of clusters: $k$}
		\psfrag{W}[c][c][0.9]{
			\begin{picture}(0,0)
				\put(0,0){\makebox(0,20)[c]{L1CM$'$~/~Inst ($\times 10^{-2}$)}}
			\end{picture}
		}
		\psfrag{P}[c][r][0.82]{$10^2$}
		\psfrag{Q}[c][c][0.82]{$10^3$}
		\psfrag{R}[c][c][0.82]{$10^4$}
		\psfrag{X}[r][r][0.82]{$0$}
		\psfrag{Y}[r][r][0.82]{$1$}
		\psfrag{Z}[r][r][0.82]{$2$}
		\psfrag{S}[r][r][0.82]{$3$}
		\psfrag{T}[r][r][0.82]{$4$}
		\psfrag{U}[r][r][0.82]{$5$}
		\psfrag{A}[r][r][0.65]{\em IVF}
		\psfrag{B}[r][r][0.65]{\em IFN}
		\psfrag{C}[r][r][0.65]{\em IFB}
		\psfrag{D}[r][r][0.65]{\em MFN}
		\includegraphics[width=40mm]{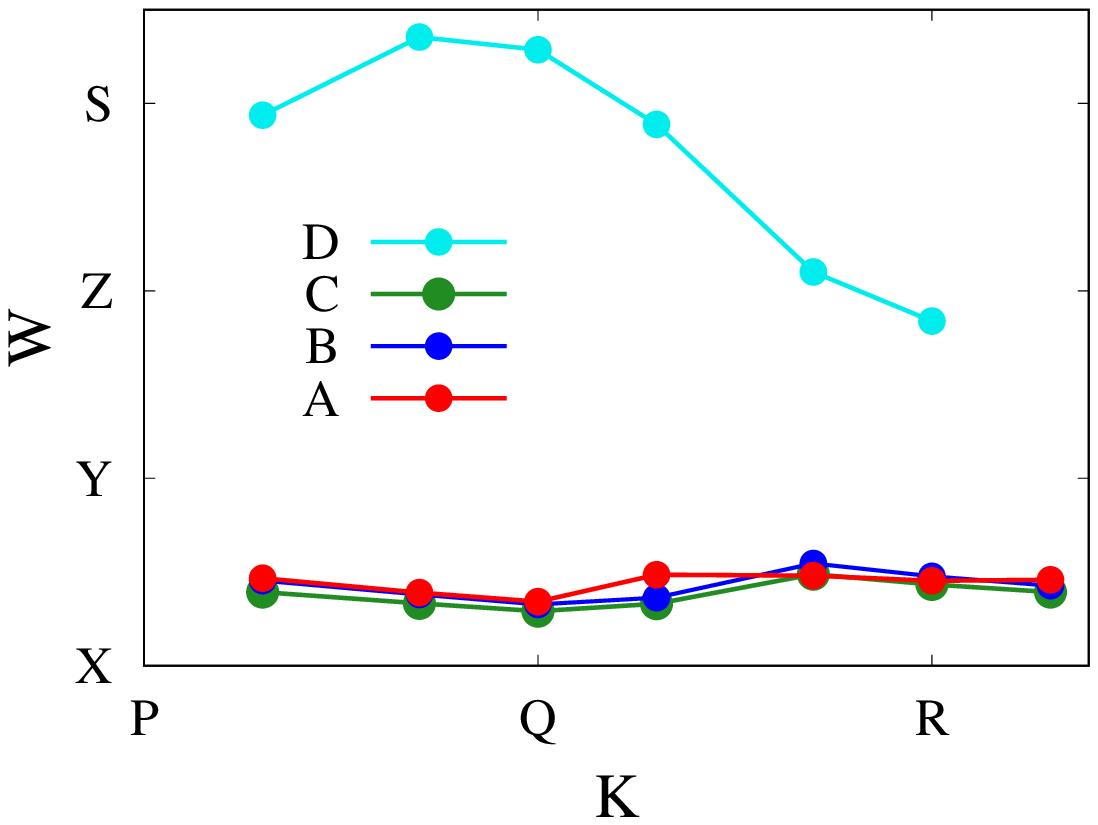}
	}\tabularnewline
	(a)~PubMed & (b)~NYT \tabularnewline
	\end{tabular}
\end{center}
\vspace*{-3mm}
\caption{Difference of numbers of L1-cache and LL-cache misses
per instruction, $\phi_1(k)$, with various $k$ 
when algorithms were applied to (a) PubMed and (b) NYT.
L1CM$'$/Inst denotes (L1CM$-$LLCM)/Inst.
}
\label{fig:l1cm-inst}
\vspace*{-2mm}
\end{figure}

\begin{figure}[t]
\begin{center}\hspace*{1mm}
	\begin{tabular}{cc}	
	\subfigure{
		\psfrag{K}[c][c][0.9]{Number of clusters: $k$}
		\psfrag{W}[c][c][0.9]{
			\begin{picture}(0,0)
				\put(0,0){\makebox(0,20)[c]{LLCM~/~Inst ($\times 10^{-2}$)}}
			\end{picture}
		}
		\psfrag{P}[c][r][0.82]{$10^2$}
		\psfrag{Q}[c][c][0.82]{$10^3$}
		\psfrag{R}[c][c][0.82]{$10^4$}
		\psfrag{X}[r][r][0.82]{$0$}
		\psfrag{Y}[r][r][0.82]{$1$}
		\psfrag{Z}[r][r][0.82]{$2$}
		\psfrag{S}[r][r][0.82]{$3$}
		\psfrag{T}[r][r][0.82]{$4$}
		\psfrag{U}[r][r][0.82]{$5$}
		\psfrag{A}[r][r][0.65]{\em IVF}
		\psfrag{B}[r][r][0.65]{\em IFN}
		\psfrag{C}[r][r][0.65]{\em IFB}
		\psfrag{D}[r][r][0.65]{\em MFN}
		\includegraphics[width=40mm]{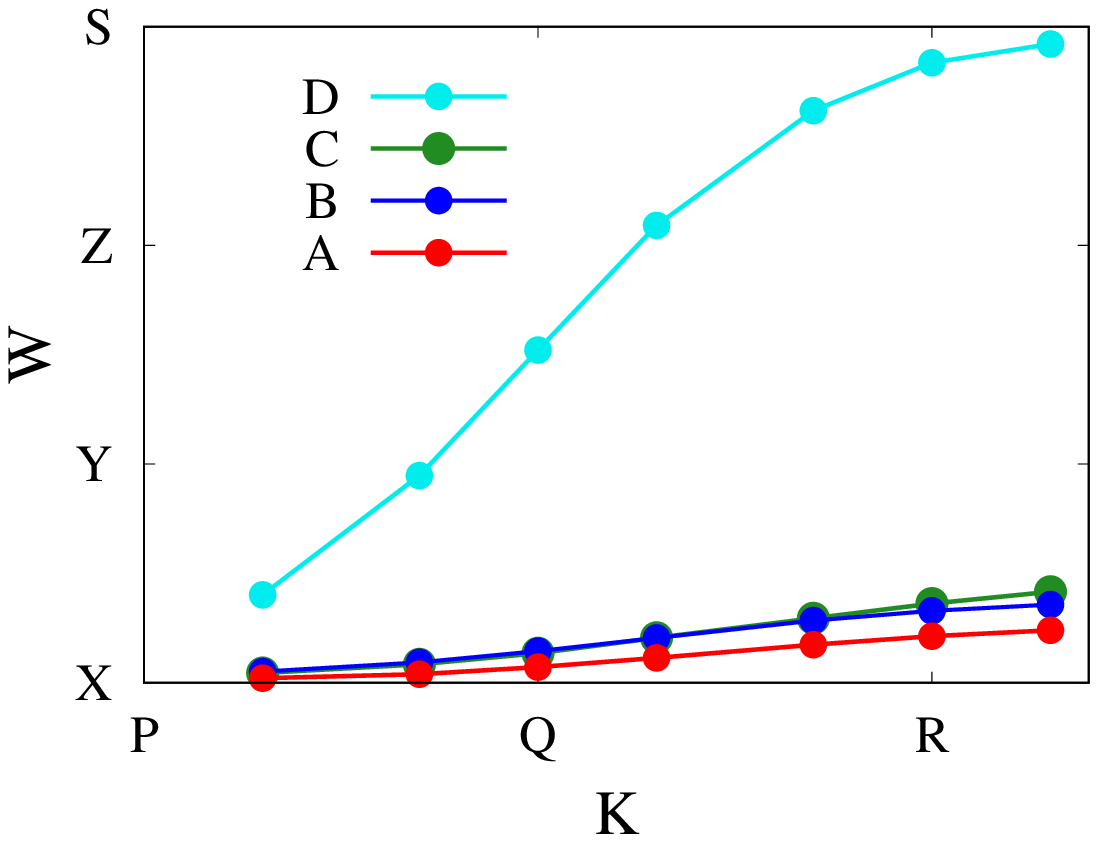}
	} &
	\subfigure{
		\psfrag{K}[c][c][0.9]{Number of clusters: $k$}
		\psfrag{W}[c][c][0.9]{
			\begin{picture}(0,0)
				\put(0,0){\makebox(0,20)[c]{LLCM~/~Inst ($\times 10^{-2}$)}}
			\end{picture}
		}
		\psfrag{P}[c][r][0.82]{$10^2$}
		\psfrag{Q}[c][c][0.82]{$10^3$}
		\psfrag{R}[c][c][0.82]{$10^4$}
		\psfrag{X}[r][r][0.82]{$0$}
		\psfrag{Y}[r][r][0.82]{$1$}
		\psfrag{Z}[r][r][0.82]{$2$}
		\psfrag{S}[r][r][0.82]{$3$}
		\psfrag{T}[r][r][0.82]{$4$}
		\psfrag{U}[r][r][0.82]{$5$}
		\psfrag{A}[r][r][0.65]{\em IVF}
		\psfrag{B}[r][r][0.65]{\em IFN}
		\psfrag{C}[r][r][0.65]{\em IFB}
		\psfrag{D}[r][r][0.65]{\em MFN}
		\includegraphics[width=40mm]{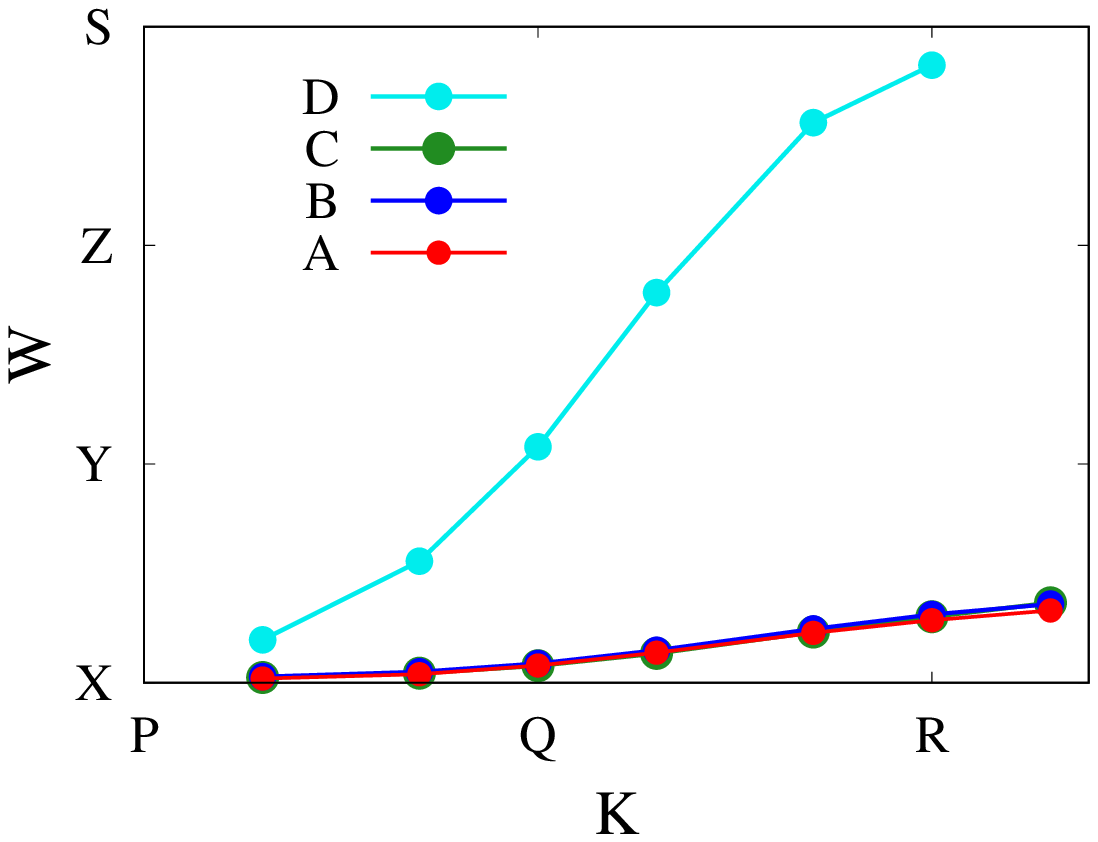}
	}\tabularnewline
	(a)~PubMed & (b)~NYT \tabularnewline
	\end{tabular}
\end{center}
\vspace*{-3mm}
\caption{Number of LL-cache misses per instruction, $\phi_2(k)$,
with various $k$ 
when algorithms were applied to (a) PubMed and (b) NYT
}
\label{fig:llcm-inst}
\vspace*{-2mm}
\end{figure}

\begin{figure}[t]
\begin{center}\hspace*{1mm}
	\begin{tabular}{cc}	
	\subfigure{
		\psfrag{K}[c][c][0.9]{Number of clusters: $k$}
		\psfrag{W}[c][c][0.9]{
			\begin{picture}(0,0)
				\put(0,0){\makebox(0,32)[c]{BM~/~Inst}}
			\end{picture}
		}
		\psfrag{P}[c][r][0.82]{$10^2$}
		\psfrag{Q}[c][c][0.82]{$10^3$}
		\psfrag{R}[c][c][0.82]{$10^4$}
		\psfrag{X}[r][r][0.82]{$10^{-6}$}
		\psfrag{Y}[r][r][0.82]{$10^{-5}$}
		\psfrag{Z}[r][r][0.82]{$10^{-4}$}
		\psfrag{S}[r][r][0.82]{$10^{-3}$}
		\psfrag{T}[r][r][0.82]{$10^{-2}$}
		\psfrag{A}[r][r][0.65]{\em IVF}
		\psfrag{B}[r][r][0.65]{\em IFN}
		\psfrag{C}[r][r][0.65]{\em IFB}
		\psfrag{D}[r][r][0.65]{\em MFN}
		\includegraphics[width=40mm]{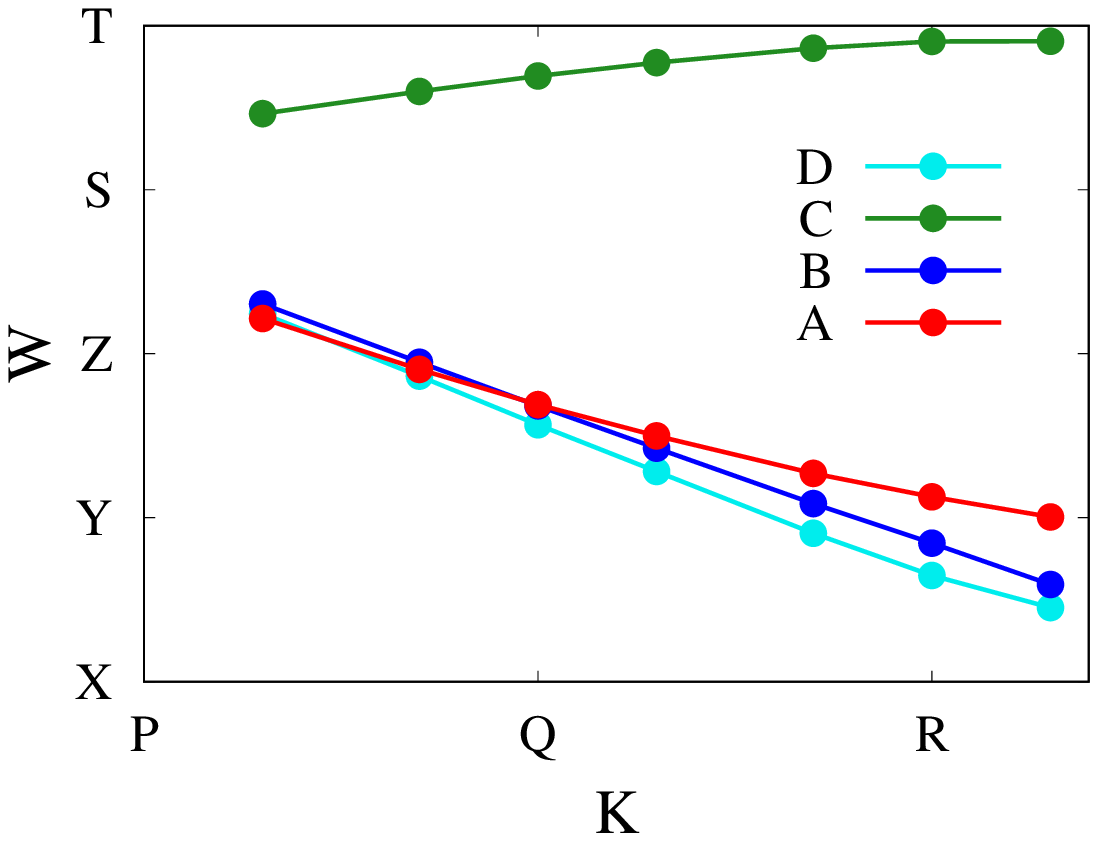}
	} &
	\subfigure{
		\psfrag{K}[c][c][0.9]{Number of clusters: $k$}
		\psfrag{W}[c][c][0.9]{
			\begin{picture}(0,0)
				\put(0,0){\makebox(0,32)[c]{BM~/~Inst}}
			\end{picture}
		}
		\psfrag{P}[c][r][0.82]{$10^2$}
		\psfrag{Q}[c][c][0.82]{$10^3$}
		\psfrag{R}[c][c][0.82]{$10^4$}
		\psfrag{X}[r][r][0.82]{$10^{-6}$}
		\psfrag{Y}[r][r][0.82]{$10^{-5}$}
		\psfrag{Z}[r][r][0.82]{$10^{-4}$}
		\psfrag{S}[r][r][0.82]{$10^{-3}$}
		\psfrag{T}[r][r][0.82]{$10^{-2}$}
		\psfrag{A}[r][r][0.65]{\em IVF}
		\psfrag{B}[r][r][0.65]{\em IFN}
		\psfrag{C}[r][r][0.65]{\em IFB}
		\psfrag{D}[r][r][0.65]{\em MFN}
		\includegraphics[width=40mm]{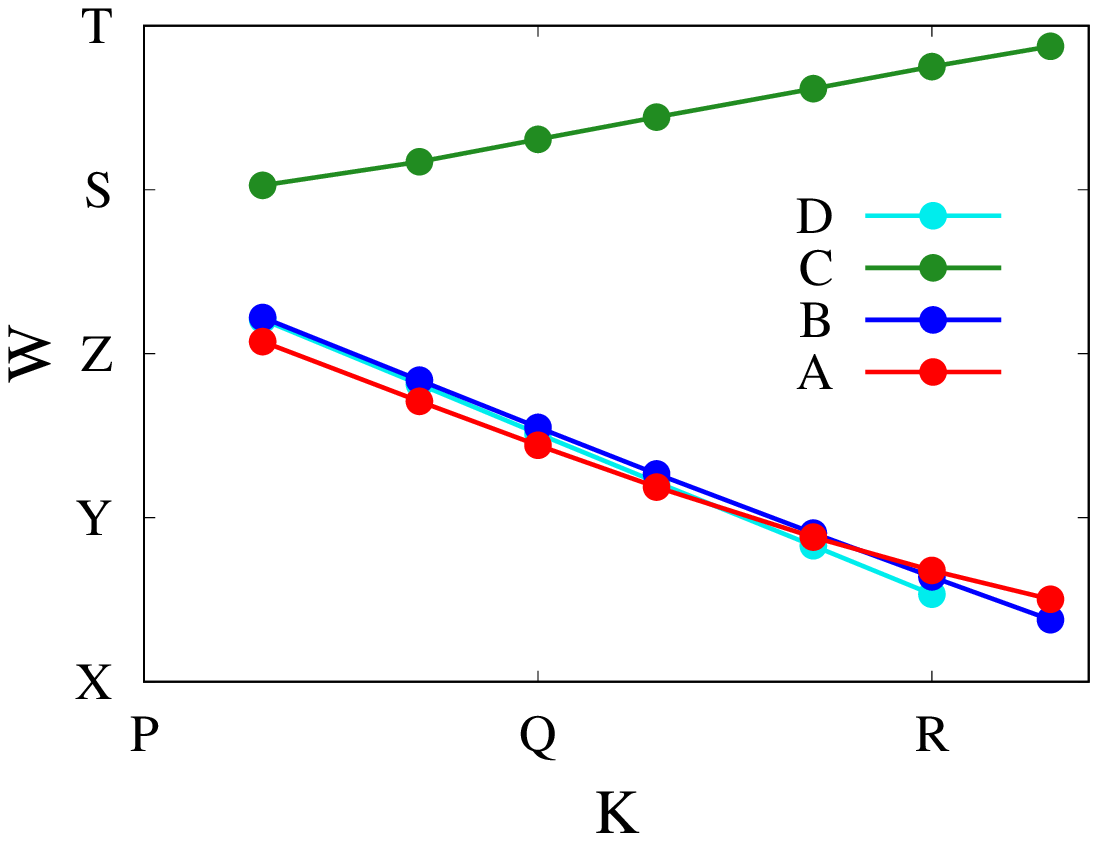}
	}\tabularnewline
	(a)~PubMed & (b)~NYT \tabularnewline
	\end{tabular}
\end{center}
\vspace*{-3mm}
\caption{Number of branch mispredictions per instruction,
$\phi_3(k)$, with various $k$ 
when algorithms were applied to (a) PubMed and (b) NYT
}
\label{fig:bm-inst}
\vspace*{-2mm}
\end{figure}

To analyze performance based on CPI, 
we introduced performance degradation factors per instruction defined by
\begin{equation}
\phi_1 = \frac{(\mbox{L1CM}\!-\!\mbox{LLCM})}{\mbox{Inst}},\:\:
\phi_2 = \frac{\mbox{LLCM}}{\mbox{Inst}},\:\:
\phi_3 = \frac{\mbox{BM}}{\mbox{Inst}}\: ,
\label{eq:model_funct}
\end{equation}
in addition to L1CM, LLCM, and BM.
Figures~\ref{fig:clk-inst}, \ref{fig:l1cm-inst}, \ref{fig:llcm-inst}, and
\ref{fig:bm-inst} show the actual CPI, $\phi_1$, $\phi_2$, and $\phi_3$
with the number of clusters $k$, 
where
L1CM$'$ in Fig.~\ref{fig:l1cm-inst} denotes $(\mbox{L1CM}\!-\!\mbox{LLCM})$
and $k$ is omitted from $\phi_i(k)$ for simplicity.
From all the figures, 
each of the algorithms indicated the same tendencies on 
the characteristics when applied to PubMed and NYT.

Figure~\ref{fig:clk-inst} shows that 
the inverted-file data structure was effective for lowering CPI.
The three algorithms with an inverted-file data structure operated at CPIs
from 0.26 to 0.61 through all $k$ values in both data sets 
while {\em MFN} ranged from 0.65 to 1.94.
{\em MFN} whose CPI exceeded 1.0 in the large $k$ range 
lost the effect of superscalar execution.
The others' CPIs were reasonable because the CPU core had eight units, 
including four ALUs \cite{hammarlund}.
In the large $k$ range, 
we arranged the four algorithms in ascending order of CPI: 
{\em IVF}, {\em IFN}, {\em IFB}, and {\em MFN}.

Figures~\ref{fig:l1cm-inst} and \ref{fig:llcm-inst} show 
L1CM$'$ per instruction ($\phi_1$) and LLCM per instruction ($\phi_2$) 
for the algorithms along $k$.
These figures indicate that the L1CM$'$/Inst and LLCM/Inst of {\em MFN} were 
conspicuously large.
The decrease of L1CM$'$/Inst in the large $k$ range was attributed to 
the high joint probability at which the L1 and LL cache misses occurred.
{\em IFB} and {\em IFN} had identical characteristics 
in terms of L1CM$'$/Inst and LLCM/Inst in the $k$ range.
This fact is used for the assumption of the parameter optimization 
in Section~\ref{subsec:cpimodel}.
Regarding LLCM/Inst, 
{\em IVF} achieved the lowest values as a whole.

Figure~\ref{fig:bm-inst} shows BM per instruction ($\phi_3$) with $k$ in the log-log scale.
{\em IFB} showed different characteristics from the others.
Its conditional branch induced many branch mispredictions 
because the branch predictor in the CPU core often failed to select the next true 
instruction due to the zeros' irregular positions in the inverted file.
This characteristic negatively impacted the speed performance of {\em IFB}, 
as shown in Section~\ref{subsec:cpimodel}.

\subsection{Clock-Cycle per Instruction (CPI) Model}\label{subsec:cpimodel}
We introduce a clock-cycle per instruction (CPI) model,
which is a simple linear function of $k$,
expressed by 
\begin{equation}
\mbox{CPI}(k) = w_0 +\sum_{i=1}^{3} w_i \cdot \phi_i(k) \: ,
\label{eq:model}
\end{equation}
where $w_0$ denotes the expected clock cycles per instruction 
when cache misses and branch mispredictions do not occur,
$w_1$ is the overall penalty per L1CM$'$/Inst 
when a level-1 data cache miss occurs and a last-level cache hit occurs 
at the worst case,
$w_2$ is the expected memory stall cycles per LLCM/Inst, and 
$w_3$ is the expected branch misprediction penalty per BM/Inst including 
the penalty of the number of wasted instructions. 
Note that $w_2$ does not mean the expected memory latency
per instruction due to the out-of-order execution \cite{hennepat}.

\begin{table}[t]
\centering
\caption{Optimized CPI model parameters and errors on actual CPIs}
\vspace*{-3mm}
\begin{spacing}{1.1}
\begin{tabular}{|c|c|c|c|c||c|c|}\hline
\multirow{2}{*}{Algo.} & \multicolumn{4}{c||}{Parameters} &
{\small Avg. err.} & {\small Max. err.} \tabularnewline \cline{2-5}
& $w_0$ & $w_1$ & $w_2$ & $w_3$ & (\%) & (\%)
\tabularnewline \hline\hline
{\em MFN} & 0.255 & 7.52 & 56.1 & 23.8 & 5.96 & 9.32 
\tabularnewline \hline
{\em IFB} & 0.262 & 5.52 & 30.8 & 23.8 & 0.969 & 3.93
\tabularnewline \hline
{\em IFN} & 0.255 & 5.52 & 30.8 & 23.8 & 0.617 & 4.55
\tabularnewline \hline
{\bf\em IVF} & {\bf 0.243} & {\bf 3.13} & {\bf 13.5} & {\bf 23.8} 
& {\bf 0.461} & {\bf 3.19} \tabularnewline \hline
\end{tabular}\label{table:params}
\end{spacing}
\vspace*{-2mm}
\end{table}

\begin{figure}[t]
\begin{center}\hspace*{1mm}
	\begin{tabular}{cc}	
	\subfigure{
		\psfrag{K}[c][c][0.9]{Number of clusters: $k$}
		\psfrag{W}[c][c][0.9]{
			\begin{picture}(0,0)
				\put(0,0){\makebox(0,25)[c]{CPI} }
			\end{picture}
		}
		\psfrag{P}[c][r][0.82]{$10^2$}
		\psfrag{Q}[c][c][0.82]{$10^3$}
		\psfrag{R}[c][c][0.82]{$10^4$}
		\psfrag{X}[r][r][0.82]{$0$}
		\psfrag{Y}[r][r][0.82]{$0.5$}
		\psfrag{Z}[r][r][0.82]{$1.0$}
		\psfrag{S}[r][r][0.82]{$1.5$}
		\psfrag{T}[r][r][0.82]{$2.0$}
		\psfrag{A}[r][r][0.65]{\em IVF}
		\psfrag{B}[r][r][0.65]{\em IFN}
		\psfrag{C}[r][r][0.65]{\em IFB}
		\psfrag{D}[r][r][0.65]{\em MFN}
		\psfrag{M}[r][r][0.57]{Model}
		\includegraphics[width=41mm]{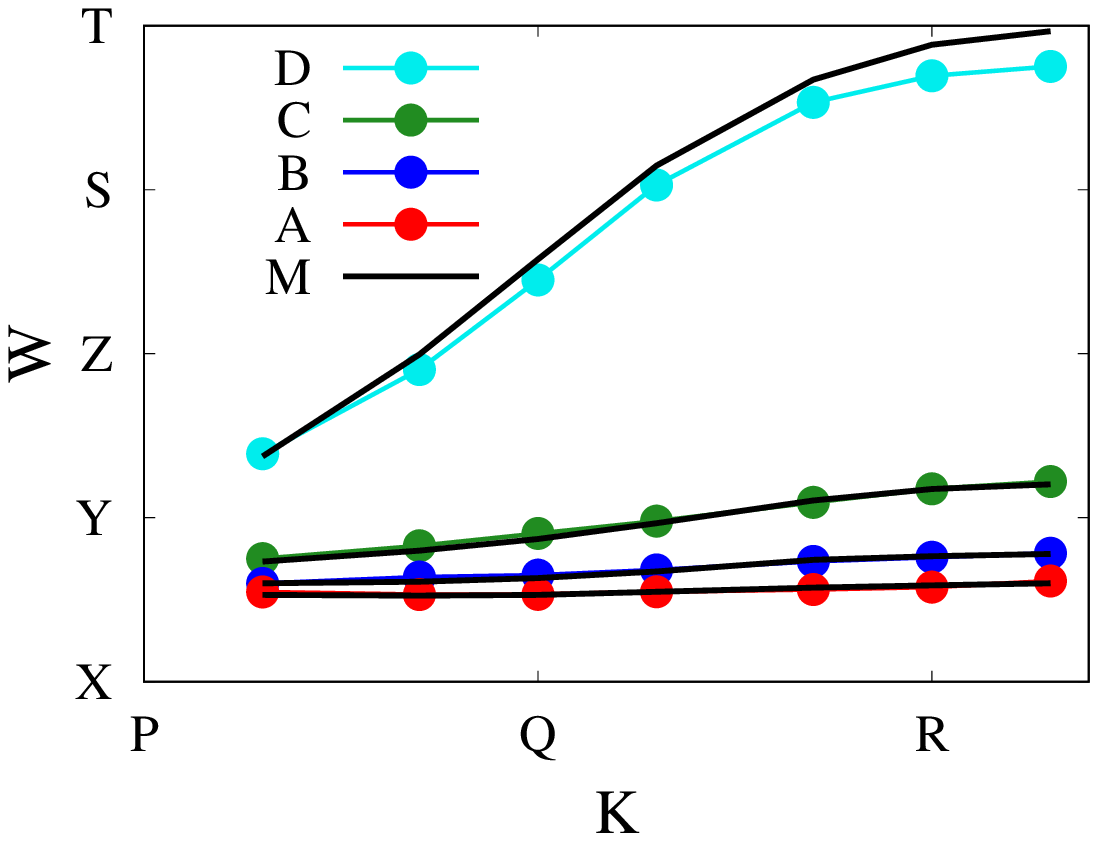}
	} &
	\subfigure{
		\psfrag{K}[c][c][0.9]{Number of clusters: $k$}
		\psfrag{W}[c][c][0.9]{
			\begin{picture}(0,0)
				\put(0,0){\makebox(0,25)[c]{CPI} }
			\end{picture}
		}
		\psfrag{P}[c][r][0.82]{$10^2$}
		\psfrag{Q}[c][c][0.82]{$10^3$}
		\psfrag{R}[c][c][0.82]{$10^4$}
		\psfrag{X}[r][r][0.82]{$0$}
		\psfrag{Y}[r][r][0.82]{$0.5$}
		\psfrag{Z}[r][r][0.82]{$1.0$}
		\psfrag{S}[r][r][0.82]{$1.5$}
		\psfrag{T}[r][r][0.82]{$2.0$}
		\psfrag{A}[r][r][0.65]{\em IVF}
		\psfrag{B}[r][r][0.65]{\em IFN}
		\psfrag{C}[r][r][0.65]{\em IFB}
		\psfrag{D}[r][r][0.65]{\em MFN}
		\psfrag{M}[r][r][0.63]{Model}
		\includegraphics[width=41mm]{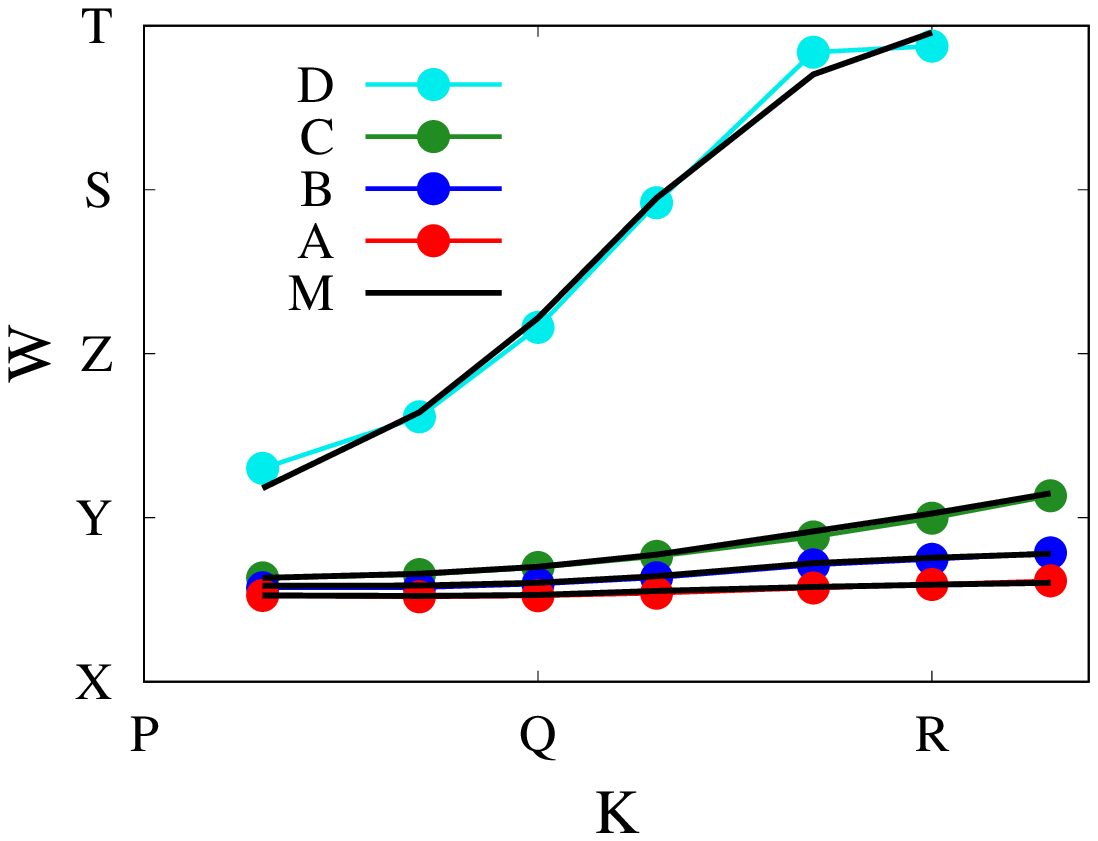}
	}\tabularnewline
	(a)~PubMed & (b)~NYT \tabularnewline
	\end{tabular}
\end{center}
\vspace*{-3mm}
\caption{Actual and model CPI with various $k$ in 
(a) PubMed and (b) NYT
}
\label{fig:cpi}
\vspace*{-2mm}
\end{figure}

For the optimization of parameters $w_i$,
we assumed that they are independent of the data sets
and dependent on the algorithms.
Based on the relationship between the pairs of algorithms,
we also made the following three assumptions.
The first is that {\em MFN} and {\em IFN} share $w_0$
because the algorithms have an identical triple loop\footnote{
Regarding the two algorithms,
the instructions executed in the triple loop were identical 
in the corresponding assembly codes.
}
at their assignment step, except the accessed data arrays 
whose data structure is either standard or inverted-file.
The second is that {\em IFB} and {\em IFN} share $w_1$ and $w_2$
since these algorithms only differ over whether 
the conditional branch in the innermost loop in the triple loop is set.
The last is that all the algorithms share $w_3$.
Under these assumptions, 
we optimized the parameters so that the squared error between 
the actual and model CPI in Eq.~(\ref{eq:model}) was minimized: 
\begin{enumerate}
\item 
{\bf Target}: $w_1$ and $w_2$ of {\em MFN}\\
{\bf Data}: Differences of CPI, $\phi_1$ and $\phi_2$ values\\
\noindent\hspace*{9mm}of {\em MFN} and {\em IFN}\\
{\bf Condition}: $w_3\!=\!0$. 
\item 
{\bf Target}: $w_0$ of {\em MFN}\\
{\bf Data}: {\em MFN}'s CPI data\\
{\bf Condition}: $w_1$ and $w_2$ are fixed at the optimized values\\
\noindent\hspace*{16mm}and $w_3\!=\!0$.
\item 
{\bf Target}: $w_1$ and $w_2$ of {\em IFN}\\
{\bf Data}: {\em IFN}'s CPI data\\
{\bf Condition}: $w_0$ is fixed at the {\em MFN}'s value\\
\noindent\hspace*{16mm} and $w_3\!=\!0$.
\item
{\bf Target}: $w_0$ and $w_3$ of {\em IFB}\\
{\bf Data}: {\em IFB}'s CPI data\\
{\bf Condition}: $w_1$ and $w_2$ are fixed at the {\em IFN}'s values.
\item
{\bf Target}: $w_0$, $w_2$, and $w_3$ of {\em IVF}\\
{\bf Data}: {\em IVF}'s CPI data\\
{\bf Condition}: $w_3$ is fixed at the {\em MFN}'s value.
\end{enumerate}

We obtained the parameters for each algorithm by this procedure 
and evaluated the accuracy of the CPI model by two measures.
One is an average error (Avg. err.): 
\begin{equation}
\mbox{Avg. err.} = \left\{ \frac{1}{|K|} \sum_{k\in K} \left(
	 \mbox{CPI}_a(k) -\mbox{CPI}_m(k) \right)^2 
        \right\}^{\frac{1}{2}}\: ,
\label{eq:avgerr}
\end{equation}
where $K$ is the set of $k$s in the experiments, i.e., 
$K\!=\!\{200,500,\cdots,20000\}$, 
and $\mbox{CPI}_a(k)$ and $\mbox{CPI}_m(k)$ denote
the actual and model CPIs when the number of clusters is $k$.
The other is a maximum error (Max. err.): 
\begin{equation}
\mbox{Max. err.} = \max_{k\in K}
	\left|\frac{\mbox{CPI}_m(k)}{\mbox{CPI}_a(k)} -1\right|\: .
\label{eq:maxerr}
\end{equation}

Table~\ref{table:params} shows the optimized parameters 
and the evaluation results.
The parameters were reasonable values 
based on the computer architecture \cite{hammarlund} in our experiments.
The errors were also below 10\% in the range of all the $k$ values.
{\em IVF}, in particular, reduced the wasted clock cycles that
were caused by the cache misses.
Figures~\ref{fig:cpi}(a) and (b) show the actual and model CPIs
of the four algorithms.
We confirmed the model CPIs agree well with the actual CPIs 
of all the algorithms.

\begin{table}[t]
\centering
\caption{Comparison of the triple loops in {\em IFN} and {\em IVF}}
\vspace*{-4mm}
\begin{spacing}{1.2}
\begin{tabular}{|c|l|l|}\hline
\# & ~\hspace{14mm}{\em IFN} & ~\hspace{16mm}{\em IVF} \tabularnewline\hline
\multirow{3}{*}{1}
 & {\small {\bf for~all} $\hat{\bm{x}}_i\!\in\!\hat{\cal X}$ {\bf do} }
 & {\small {\bf for~all} $\hat{\bm{x}}_i\!\in\!\hat{\cal X}$ {\bf do} }
\tabularnewline
 & {\small $/\!/~|\hat{\cal X}|\!=\!N$ repeats.}
 & \multirow{2}{*}{\small $\Leftrightarrow$~Identical to {\em IFN} }
\tabularnewline
 & {\small $/\!/~\hat{\bm{x}_i}\!=\!(t_{(i,h)},v_{(i,h)})_{h=1}^{(nt)_i}$ } 
 & \tabularnewline\hline
\multirow{2}{*}{2}
 & {\small {\bf for~all} $s\!\leftarrow\!t_{(i,h)}\!\in\!S_i$ {\bf do} }
 & {\small {\bf for~all} $s\!\leftarrow\!t_{(i,h)}\!\in\!S_i$ {\bf do} }
\tabularnewline
 & {\small $/\!/~|S_i|\!=\!(nt)_i$ repeats.}
 & {\small $\Leftrightarrow$~Identical to {\em IFN} } \tabularnewline\hline\hline
\multirow{5}{*}{3}
 & {\small {\bf for~all} ~$u_{(s,j)}\!\in \bar{\bm{\xi}}_s$~{\bf do} }
 & {\small {\bf for~all} ~$(c_{(s,q)},u_{(s,q)})\!\in \breve{\bm{\xi}}_s$~{\bf do} }
\tabularnewline
 & {\small $/\!/~\bar{\bm{\xi}_s}   \in \bar{\cal M}$  }
 & {\small $/\!/~\breve{\bm{\xi}_s} \in \breve{\cal M}$  }
\tabularnewline
 & {\small $/\!/~\bar{\bm{\xi}_s}\!=\!(u_{(s,j)})_{j=1}^{k}$  }
 & {\small $/\!/~\breve{\bm{\xi}_s}\!=\!(c_{(s,q)},u_{(s,q)})_{q=1}^{(nc)_s}$  }
\tabularnewline
 & {\small $\rho_j\!\leftarrow\!\rho_j\!+\!v_{(i,h)}\!\times\!u_{(s,j)}$  }
 & {\small $\rho_{c_{(s,q)}}\!\leftarrow\!\rho_{c_{(s,q)}}\!+\!v_{(i,h)}\!\times\!u_{(s,q)}$  }
\tabularnewline
 & {\small $/\!/~\bm{k}$ repeats.}
 & {\small $/\!/~\bm{(nc)_s}$ repeats.} \tabularnewline\hline
\end{tabular}\label{table:ifn}
\end{spacing}
\vspace*{-2mm}
\end{table}

\begin{figure}[t]
\begin{center}
	\begin{tabular}{cc}
	\subfigure{
		\includegraphics[height=33mm]{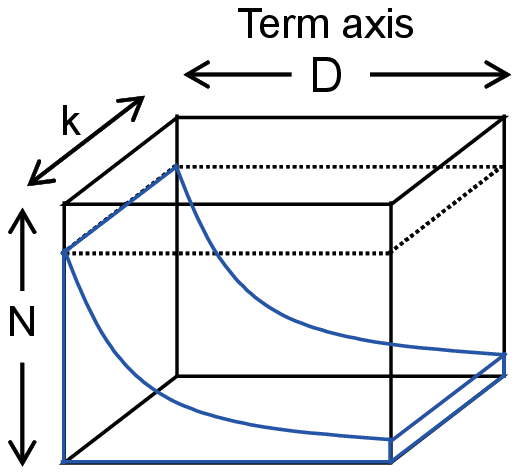}
	} &
	\subfigure{
		\hspace*{-3mm}
		\includegraphics[height=33mm]{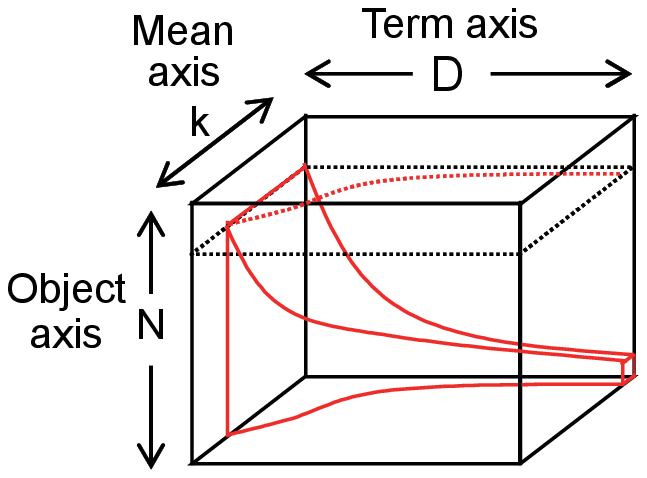}
	}\tabularnewline
	(a)~IFN  & (b)~IVF\tabularnewline
	\end{tabular}
\end{center}
\vspace*{-2mm}
\caption{
Conceptual diagram of number of multiplications
executed in triple loop of (a) {\em IFN} and (b) {\em IVF}.
Numbers correspond to volumes surrounded by curves 
illustrated in rectangles.
}
\label{fig:diag}
\end{figure}
 
\section{Discussion}\label{sec:disc}
We compare {\em IVF} with two similar algorithms, 
{\em IFN} in Section~\ref{sec:algos} and {\em IVFD} 
that is related to wand-k-means \cite{broder} in Section~\ref{subsec:ivfstr},
and discuss their performances.

\vspace*{-3mm}
\subsection{IFN and IVF}\label{subsec:ifn}
{\em IFN} operated in less CPU time than {\em IVF} 
in the small $k$ range in Figs.~\ref{fig:time}(a) and (b).
From the viewpoints of the performance degradation factors,
{\em IVF} was inferior in this range to {\em IFN} based on 
the number of instructions in Figs.~\ref{fig:inst}(a) and (b).
We focus on the number of instructions needed by each algorithm, 
especially in the triple loop at the assignment step 
because most of the CPU time was spent in the triple loop based on 
our preliminary analyses.

Table~\ref{table:ifn} shows an overview of the triple loops 
in {\em IFN} and {\em IVF}.
The two algorithms only differ in the innermost loop labeled as 3.
{\em IFN} loads feature value $u_{(s,j)}$ in the $j$-th entry 
in array $\bar{\bm \xi}_s$ 
from an external memory or a cache, 
multiplies $u_{(s,j)}$ with $v_{(i,h)}$, and 
adds a multiplication value to partial similarity $\rho_j$.
This procedure is repeated by 
\begin{equation}
k\cdot \sum_{i=1}^N (nt)_i = \sum_{i=1}^N \sum_{h=1}^{(nt)_i} k\: ,
\label{eq:ifn_mult}
\end{equation}
where the number of repetitions corresponds to the number of multiplications.
By contrast, {\em IVF} loads the tuple of 
mean ID $c_{(s,q)}$ and feature value $u_{(s,q)}$ in the $q$-th entry
in array $\breve{\bm \xi}_s$. 
The number of repetitions of the foregoing procedure is expressed by
\begin{equation}
\sum_{i=1}^N \sum_{h=1}^{(nt)_i} (nc)_s,\:~~~\: s = t_{(i,h)} \: .
\label{eq:ivf_mult}
\end{equation}

\begin{figure}[t]
\begin{center}\hspace*{1mm}
	\begin{tabular}{cc}	
	\subfigure{
		\psfrag{K}[c][c][0.9]{
			\begin{picture}(0,0)
				\put(0,0){\makebox(0,-6)[c]{Number of clusters: $k$} }
			\end{picture}
		}
		\psfrag{W}[c][c][0.9]{
			\begin{picture}(0,0)
				\put(0,0){\makebox(0,30)[c]{\# multiplications} }
			\end{picture}
		}
		\psfrag{P}[l][r][0.82]{$10^2$}
		\psfrag{Q}[c][c][0.82]{$10^3$}
		\psfrag{R}[c][c][0.82]{$10^4$}
		\psfrag{X}[r][r][0.82]{$10^{10}$}
		\psfrag{Y}[r][r][0.82]{$10^{11}$}
		\psfrag{Z}[r][r][0.82]{$10^{12}$}
		\psfrag{A}[r][r][0.65]{{\em IVF}}
		\psfrag{B}[r][r][0.65]{{\em IFN}}
		\includegraphics[width=38mm]{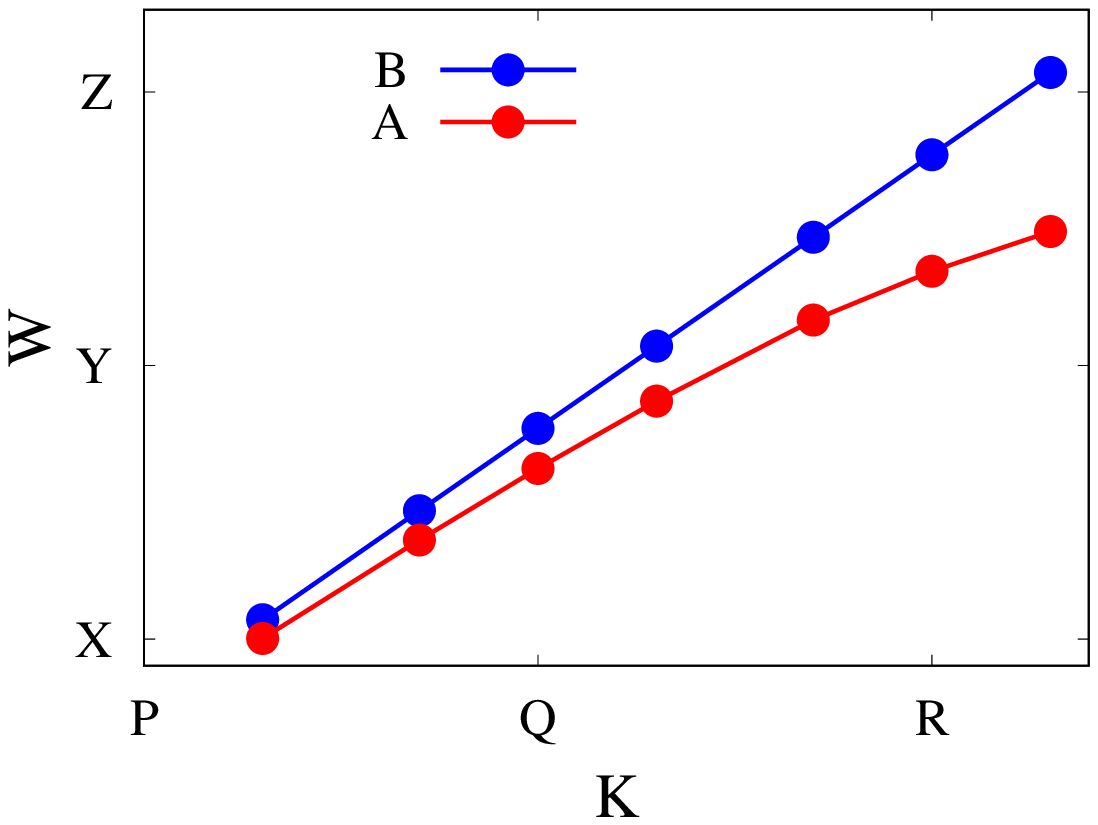}
	} &
	\subfigure{
		\psfrag{K}[c][c][0.9]{Number of clusters: $k$}
		\psfrag{W}[c][c][0.9]{
			\begin{picture}(0,0)
				\put(0,0){\makebox(0,30)[c]{\# multiplications}}
			\end{picture}
		}
		\psfrag{P}[l][r][0.82]{$10^2$}
		\psfrag{Q}[c][c][0.82]{$10^3$}
		\psfrag{R}[c][c][0.82]{$10^4$}
		\psfrag{X}[r][r][0.82]{$10^{10}$}
		\psfrag{Y}[r][r][0.82]{$10^{11}$}
		\psfrag{Z}[r][r][0.82]{$10^{12}$}
		\psfrag{S}[r][r][0.82]{$10^{13}$}
		\psfrag{A}[r][r][0.65]{{\em IVF}}
		\psfrag{B}[r][r][0.65]{{\em IFN}}
		\includegraphics[width=38mm]{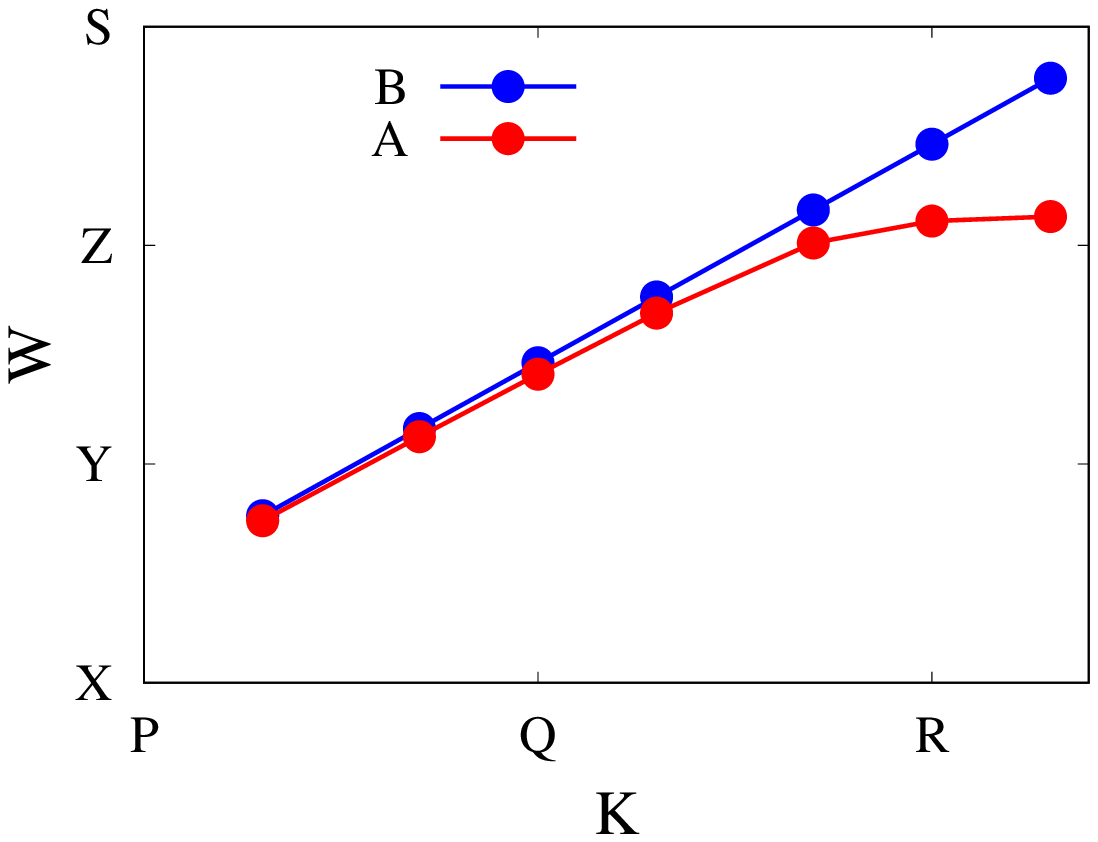}
	}\tabularnewline
	(a)~PubMed  & (b)~NYT
	\end{tabular}
\end{center}
\vspace*{-3mm}
\caption{Number of multiplications in triple loop in log-log scale 
for (a) PubMed and (b) NYT
}
\label{fig:triple_mult}
\vspace*{-2mm}
\end{figure}

Figures~\ref{fig:diag}(a) and (b) intuitively clarify 
the number of multiplications.
This shows a conceptual diagram\footnote{
Actually the term order sorted on the number of centroids 
does not always meet that sorted on the number of objects.
For this reason, both the numbers of centroids and objects do not 
decrease monotonically, as shown in Fig.~\ref{fig:diag}(b).
}
of the number of multiplications executed in the triple loops 
by {\em IFN} and {\em IVF}.
The number of multiplications is represented as the volume 
surrounded by the curves in the rectangle.
The curve in the (Term axis)-(Object axis) plane, 
which is shared by the two algorithms, 
depicts a distribution of objects each of whose feature vectors 
contains a value of the corresponding term.
The area surrounded by the curve in Fig.~\ref{fig:diag}(a) is 
$\sum_{i=1}^{N} (nt)_i$, and the volume is expressed by 
Eq.~(\ref{eq:ifn_mult}) for {\em IFN}.
The curve in the (Term axis)-(Mean axis) plane in Fig.~\ref{fig:diag}(b) 
illustrates the distribution of means, each of whose feature vectors 
contains a value of the corresponding term.
The volume of {\em IVF} is expressed by Eq.~(\ref{eq:ivf_mult}).
Figures~\ref{fig:triple_mult}(a) and (b) show the numbers of 
multiplications executed by {\em IFN} and {\em IVF} 
in their triple loops. 
The number of multiplications by {\em IVF} is smaller 
than that by {\em IFN} in every $k$ range, and 
such differences gradually increase with $k$, i.e., 
where the increase of {\em IVF}'s curve is suppressed.
This is because 
the average sparsity of the mean feature vectors decreases with $k$ 
(Fig.~\ref{fig:mterms}(b)).

\begin{figure}[t]
\vspace*{-2mm}
\begin{center}\hspace*{1mm}
	\begin{tabular}{cc}	
	\subfigure{
		\psfrag{K}[c][c][0.9]{
			\begin{picture}(0,0)
				\put(0,0){\makebox(0,-6)[c]{Number of clusters: $k$} }
			\end{picture}
		}
		\psfrag{W}[c][c][0.9]{
			\begin{picture}(0,0)
				\put(0,0){\makebox(0,30)[c]{\# instructions} }
			\end{picture}
		}
		\psfrag{P}[l][r][0.82]{$10^2$}
		\psfrag{Q}[c][c][0.82]{$10^3$}
		\psfrag{R}[c][c][0.82]{$10^4$}
		\psfrag{X}[r][r][0.82]{$10^{11}$}
		\psfrag{Y}[r][r][0.82]{$10^{12}$}
		\psfrag{Z}[r][r][0.82]{$10^{13}$}
		\psfrag{S}[r][r][0.82]{$10^{14}$}
		\psfrag{A}[r][r][0.65]{{\em IVF}}
		\psfrag{B}[r][r][0.65]{{\em IFN}}
		\includegraphics[width=40mm]{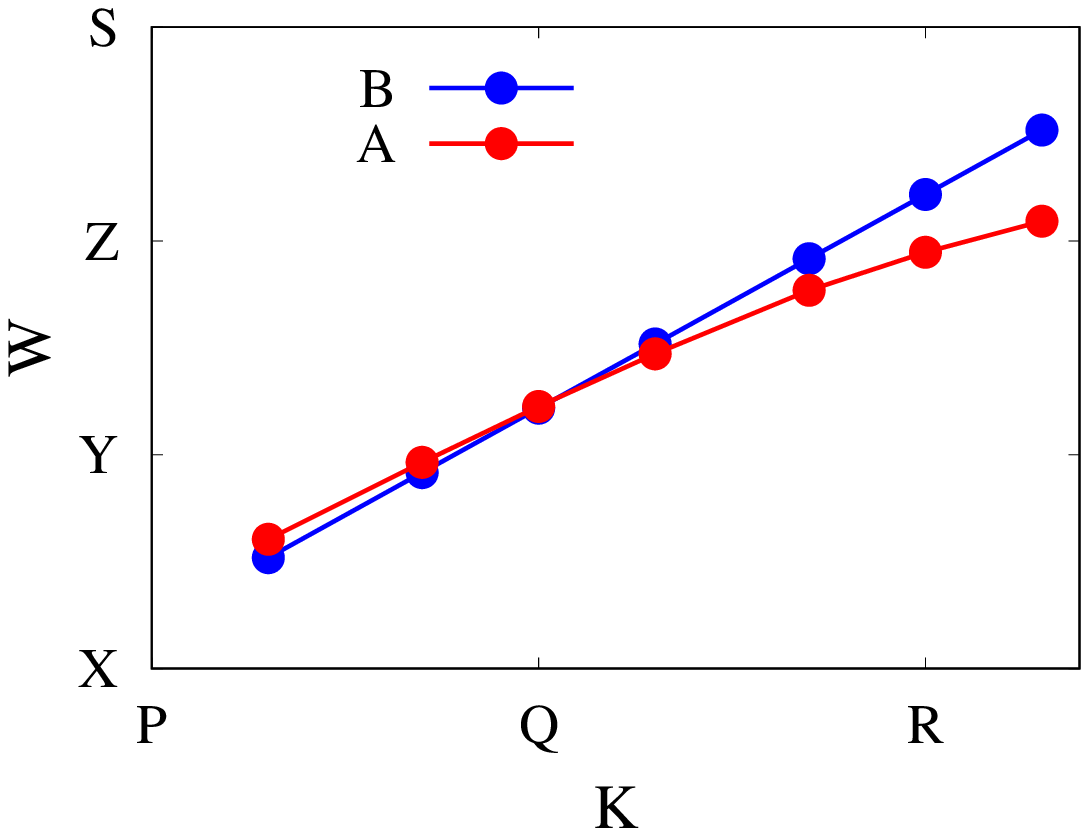}
	} &
	\subfigure{
		\psfrag{K}[c][c][0.9]{Number of clusters: $k$}
		\psfrag{W}[c][c][0.9]{
			\begin{picture}(0,0)
				\put(0,0){\makebox(0,30)[c]{\# instructions}}
			\end{picture}
		}
		\psfrag{P}[l][r][0.82]{$10^2$}
		\psfrag{Q}[c][c][0.82]{$10^3$}
		\psfrag{R}[c][c][0.82]{$10^4$}
		\psfrag{Y}[r][r][0.82]{$10^{12}$}
		\psfrag{Z}[r][r][0.82]{$10^{13}$}
		\psfrag{S}[r][r][0.82]{$10^{14}$}
		\psfrag{T}[r][r][0.82]{$10^{15}$}
		\psfrag{A}[r][r][0.65]{{\em IVF}}
		\psfrag{B}[r][r][0.65]{{\em IFN}}
		\includegraphics[width=40mm]{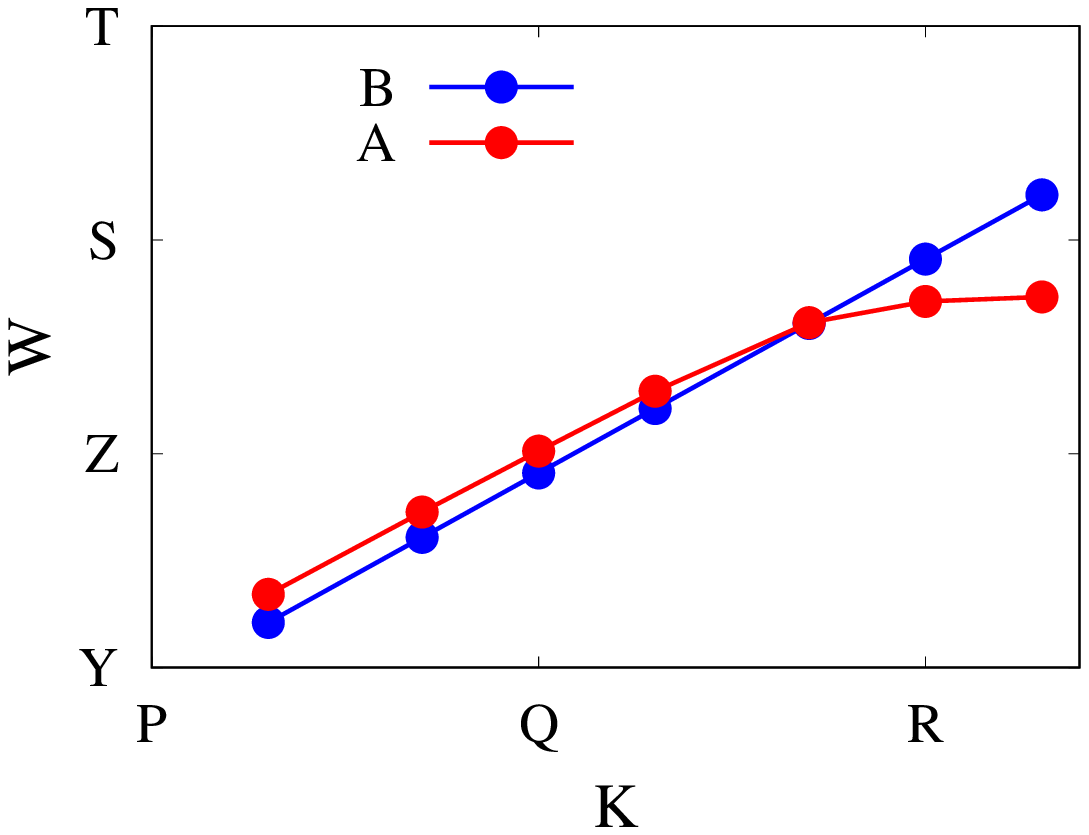}
	}\tabularnewline
	(a)~PubMed  & (b)~NYT
	\end{tabular}
\end{center}
\vspace*{-3mm}
\caption{Number of instructions in the triple loop in log-log scale 
for (a) PubMed and (b) NYT.
}
\label{fig:triple_inst}
\vspace*{-2mm}
\end{figure}

Assume that 
when a procedure for an entry in an array 
($\bar{\bm \xi}_s$ or $\breve{\bm \xi}_s$) 
in the innermost loop is performed once,
the numbers of instructions executed by {\em IFN} and {\em IVF} 
are $\alpha$ and $\beta$.
Note that $\beta$ is larger than $\alpha$ 
by the number of instructions 
by which {\em IVF} loads the mean IDs, $c_{(s,q)}$, 
$q\!\in\!\{ 1,2,\cdots,(nc)_s \}$.
We ignore the instructions for loading $(nc)_s$ itself due to their smaller numbers.
Then the numbers of instructions are expressed by
\begin{equation}
\left\{ \begin{array}{lll}
\alpha\cdot \sum_{i=1}^N \sum_{h=1}^{(nt)_i} k      & \mbox{for} 
 & \mbox{\em IFN} \\
\beta\cdot  \sum_{i=1}^N \sum_{h=1}^{(nt)_i} (nc)_s & \mbox{for}
 & \mbox{\em IVF}
\end{array}\right. \: .
\label{eq:vol}
\end{equation}
Both $\alpha$ and $\beta$ depend on the computer architecture on which 
the algorithms operate.
The number of multiplications depends on the sparsity of the object feature vectors,
and in {\em IVF} it furthermore depends on the sparsity of 
the mean feature vectors.

We obtained $\alpha\!=\!28$ and $\beta\!=\!40$ in our preliminary analysis of 
the assembly codes generated from the source codes of the algorithms 
and applied them to Eq.~(\ref{eq:vol}).
Figures~\ref{fig:triple_inst}(a) and (b) show the results, 
which are compared to the average numbers of instructions per iteration 
in Figs.~\ref{fig:inst}(a) and (b).
The cross points of the two curves of {\em IFN} and {\em IVF} appeared 
at almost the same $k$ values in 
Figs.~\ref{fig:triple_inst} and \ref{fig:inst}.
We believe that 
the difference in the speed performance of {\em IFN} and {\em IVF} 
is mainly caused by the difference of the number of instructions in the triple loop.

We provide the condition that {\em IVF} achieves better performance than {\em IFN} 
as follows:
\begin{equation}
\frac{\alpha}{\beta} >
\frac{\sum_{i=1}^N\sum_{h=1}^{(nt)_i}(nc)_s}{k\cdot\sum_{i=1}^N (nt)_i}
=\frac{1}{k}\cdot
\frac{\sum_{p=1}^D\{ (nc)_p\!\times\!(no)_p\}}{\sum_{p=1}^D (no)_p}\:,
\label{eq:ifn-ivf_cond}
\end{equation}
where $(nc)_p$ and $(no)_p$ denote the numbers of centroids (means) and objects 
that contain a term with global term ID $p$, i.e., 
$(nc)_p$ and $(no)_p$ are the centroid and document frequencies of 
the $p$-th term.
Thus {\em IVF} operates faster than {\em IFN}
when value ($\alpha/\beta$) that is determined by a computer architecture is larger 
than a right-hand side value in Eq.~(\ref{eq:ifn-ivf_cond}) that is 
determined by given data objects and generated $k$ means.

\subsection{IVFD and IVF}\label{subsec:ivfd}
Assume that both the object and mean feature vectors are represented 
with a sparse expression.
This presents a problem: 
which feature vectors should be inverted to achieve high performance? 

\subsubsection{Inverted-File for Data Object Feature Vectors}
To address the foregoing problem, 
we designed a Lloyd-type algorithm {\em IVFD} 
that applies the inverted-file data structure to the data object feature vectors
described in Section~\ref{subsec:ivfstr}.
This approach is the same as that of wand-k-means \cite{broder}, 
although it employs 
a heuristic search instead of a linear-scan search 
for determining the objects' assignments to clusters.
To focus on only the basic data structure, 
{\em IVFD} adopts a linear-scan search 
to find the most similar centroid (mean) 
when each mean feature vector is given as a query.
To reduce the computational cost for updating the mean feature vectors,
{\em IVFD} utilizes not only the inverted-file data structure but also 
the standard data structure for the object feature vectors
at the expense of consuming double memory capacitance\footnote{
A mean-update step using 
object feature vectors with inverted-file data structure 
required much more CPU time than that with the standard data structure 
in our preliminary experiments.
}.
%
%-- IVFD \ref{algo:ivfd} --%
\begin{algorithm}[t]
  \caption{~~{\em IVFD} at the $r$-th iteration} 
  \label{algo:ivfd}
    \begin{algorithmic}[1]
		\STATE{\textbf{Input:} $\breve{\cal X}$ (sparse \& inverted-file),
			~~$\hat{\cal X}$ (sparse),
			~$\hat{\cal M}^{[r-1]}$},~($k$)
		\STATE{\textbf{Output:} ${\cal C}^{[r]}\!=\!\{ C_1^{[r]},C_2^{[r]},\cdots,C_k^{[r]}\}$,~
				$\hat{\cal M}^{[r]}$ (sparse)}
		\STATE{$C_{j}^{[r]}\leftarrow\emptyset$~,~~$j=1,2,\cdots,k$~}
		%
		%-- Assignment step --%
		\\\COMMENT{~$/\!/$--~{\bf Assignment step}~--$/\!/$~}
		\STATE{$\bm{\rho}\!=\!(\rho_1,\rho_2,\cdots,\rho_i,\cdots,\rho_N)\!\leftarrow\!\bm{0}$}
		\STATE{$\bm{\rho}_{max}\!=\!(\rho_{max(1)},\cdots,\rho_{max(i)},
					\cdots,\rho_{max(N)})\!\leftarrow\!\bm{0}$}
		\FORALL{$\hat{\bm{\mu}}_j^{[r-1]}\!=\!
			[(\tau_{(j,h)},v'_{(j,h)})_{h=1}^{(n\tau)_j}]^{[r-1]}\in \hat{\cal M}^{[r-1]}$~~}
			\label{algo:ivfd-start}
		\STATE{$S'_{j}\!=\!
			\{\tau_{(j,1)},~\tau_{(j,2)},~\cdots,~\tau_{(j,h)},~\cdots,~\tau_{(j,(n\tau)_j)}\}$}
		\FORALL{~$s\!\leftarrow\!\tau_{(j,h)}\in S'_{j}$~}
		\FORALL{
			$(o_{(s,q)},u_{(s,q)}) \in \breve{\bm{\zeta}}_s$
		}
		\STATE{
			$\breve{\bm{\zeta}}_s\!=\!(o_{(s,q)},u_{(s,q)})_{q=1}^{(no)_s} 
				\in\!\breve{\cal X}$
		}
		\STATE{
			$\rho_{o_{(s,q)}} \leftarrow \rho_{o_{(s,q)}}+v'_{(j,h)}\!\times\! u_{(s,q)}$
		}
		\ENDFOR\ENDFOR
		\FOR{~~$i\!=\! 1$ to~~$N$~}
			\STATE{{\bf if}~~~$\rho_i\!>\! \rho_{max(i)}$~~{\bf then}~
					$\rho_{max(i)}\!\leftarrow\!\rho_i$~and~$a(\hat{\bm{x}}_i)\!\leftarrow\! j$}
		\ENDFOR
		\ENDFOR
		\FOR{~~$i\!=\! 1$ to~~$N$~}
			\STATE{$C_{a(\hat{\bm{x}}_i})^{[r]}\leftarrow 
				C_{a(\hat{\bm{x}}_i)}^{[r]}\cup\{ \hat{\bm{x}}_i\}$}
		\ENDFOR
		%
		%
		%-- Update step --%
		\\\COMMENT{~$/\!/$--~{\bf Update step}~--$/\!/$~}
		\STATE{$h_p\!\leftarrow\!0$,~~~$p=1,2,\cdots,D$}
		\FORALL{$C_j^{[r]} \in {\cal C}^{[r]}$}
		\STATE{$\bm{w}\!=\!(w_1,w_2,\cdots,w_D)\!\leftarrow\! \bm{0}$}
		\FORALL{$\hat{\bm x}_i\!=\!(t_{(i,h)},v_{(i,h)})_{h=1}^{(nt)_i} \in C_j^{[r]}$}
		\FORALL{$s\!\leftarrow\!t_{(i,h)}\in \{t_{(i,1)},\cdots,t_{(i,(nt)_i)} \}$ }
			\STATE{$w_{s}\!\leftarrow\!w_{s}\! +\!v_{(i,h)}$}
		\ENDFOR\ENDFOR
		\STATE{{\bf for}~~$p\!=\!1$~~to~~$D$~~{\bf do}~
			$w_p\!\leftarrow\! w_p/|C_j^{[r]}|$~~{\bf end~for}}
		\FOR{~$p\!=\!1$~~to~~$D$~}
		\IF{~$w_p\!\neq\!0$~~}
		\STATE{$\tau_{(j,h_p)}\!\leftarrow\! p$,~~
			$v'_{(j,h_p)}\!\leftarrow\! w_p/\|\bm{w}\|_2$,~~
			$h_p\!\leftarrow\! h_p\!+\!1$}
		\ENDIF\ENDFOR
		\ENDFOR
       	\STATE{\textbf{return}~~${\cal C}^{[r]}\!=\!\{ C_1^{[r]},C_2^{[r]},\cdots,C_k^{[r]}\}$,~
				$\hat{\cal M}^{[r]}$}
\end{algorithmic}
\end{algorithm}

{\bf Algorithm~\ref{algo:ivfd}} shows the {\em IVFD} pseudocode 
at the $r$-th iteration.
{\em IVFD} receives a set of the mean feature vectors 
represented by a 
standard data structure with sparse expression $\hat{\cal M}^{[r-1]}$ and uses 
two invariant object sets of the feature vectors
with inverted-file sparse expression $\breve{\cal X}$ and 
standard sparse expression $\hat{\cal X}$ 
and returns cluster set ${\cal C}^{[r]}$ consisting of $k$ clusters and 
$\hat{\cal M}^{[r]}$.
At the assignment step, similarities $\rho_i$, $i\!=\!1,\cdots,N$, 
between mean feature vector
$\hat{\bm{\mu}}_j\!=\!(\tau_{(j,h)},v'_{(j,h)})$ 
and every object feature vectors are calculated and stored, 
using inverted-file $\breve{\cal X}$ for the object features.
Note that $\tau_{(j,h)}$ and $v'_{(j,h)}$ denote 
the global term ID accessed by the tuple of mean ID $j$ and local counter $h$ 
and the corresponding feature value.
The inverted-file $\breve{\cal X}$ consists of 
$D$ arrays $\breve{\bm{\zeta}}_s$ with $(no)_s$ entries, where $s\!=\!\tau_{(j,h)}$.
The $q$-th entry in $\breve{\bm{\zeta}}_s$ is tuple $(o_{(s,q)},u_{(s,q)})$, 
where $q\!=\!1,2,\cdots,(no)_s$ and 
$o_{(s,q)}$ and $u_{(s,q)}$ denote the object ID $i$ ($o_{(s,q)}\!=\!i$) 
and the corresponding feature value.
At the update step, mean feature vector with standard sparse expression 
$\hat{\bm{\mu}}_j^{[r]}$ is calculated using object feature vectors 
with standard sparse expression
$\hat{\bm{x}}_i\!\in\!C_j^{[r]}$.
This algorithm differs from that in
{\bf Algorithm~\ref{algo:ours}}. 
However, the main difference between them 
is only the order of the triple loop and the data structures 
for the object feature vectors and the mean feature vectors\footnote{
Exactly, {\em IVFD} differs from {\em IVF} in the positions in source codes
at which the final assignment of each object to a cluster is executed.
{\em IVFD} executes the assignment outside the triple loop; 
{\em IVF} does so inside.
}.

\subsubsection{Performance Comparison}
\begin{figure}[t]
\begin{center}\hspace*{1mm}
	\begin{tabular}{cc}	
	\subfigure{
		\psfrag{K}[c][c][0.88]{
			\begin{picture}(0,0)
				\put(0,0){\makebox(0,-6)[c]{Number of clusters: $k (\times 10^3)$} }
			\end{picture}
		}
		\psfrag{W}[c][c][0.88]{
			\begin{picture}(0,0)
				\put(0,0){\makebox(0,27)[c]{Max. memory size (GB)} }
			\end{picture}
		}
		\psfrag{O}[l][r][0.82]{0}
		\psfrag{P}[c][c][0.82]{5}
		\psfrag{Q}[c][c][0.82]{10}
		\psfrag{R}[c][c][0.82]{15}
		\psfrag{S}[c][c][0.82]{20}
		\psfrag{T}[r][r][0.82]{0}
		\psfrag{X}[r][r][0.82]{0.5}
		\psfrag{Y}[r][r][0.82]{1.0}
		\psfrag{Z}[r][r][0.82]{1.5}
		\psfrag{U}[r][r][0.82]{2.0}
		\psfrag{A}[r][r][0.65]{{\em IVF}}
		\psfrag{B}[r][r][0.65]{{\em IVFD}}
		\psfrag{C}[c][c][0.65]{0.707}
		\psfrag{D}[c][c][0.65]{1.414}
		\includegraphics[width=40mm]{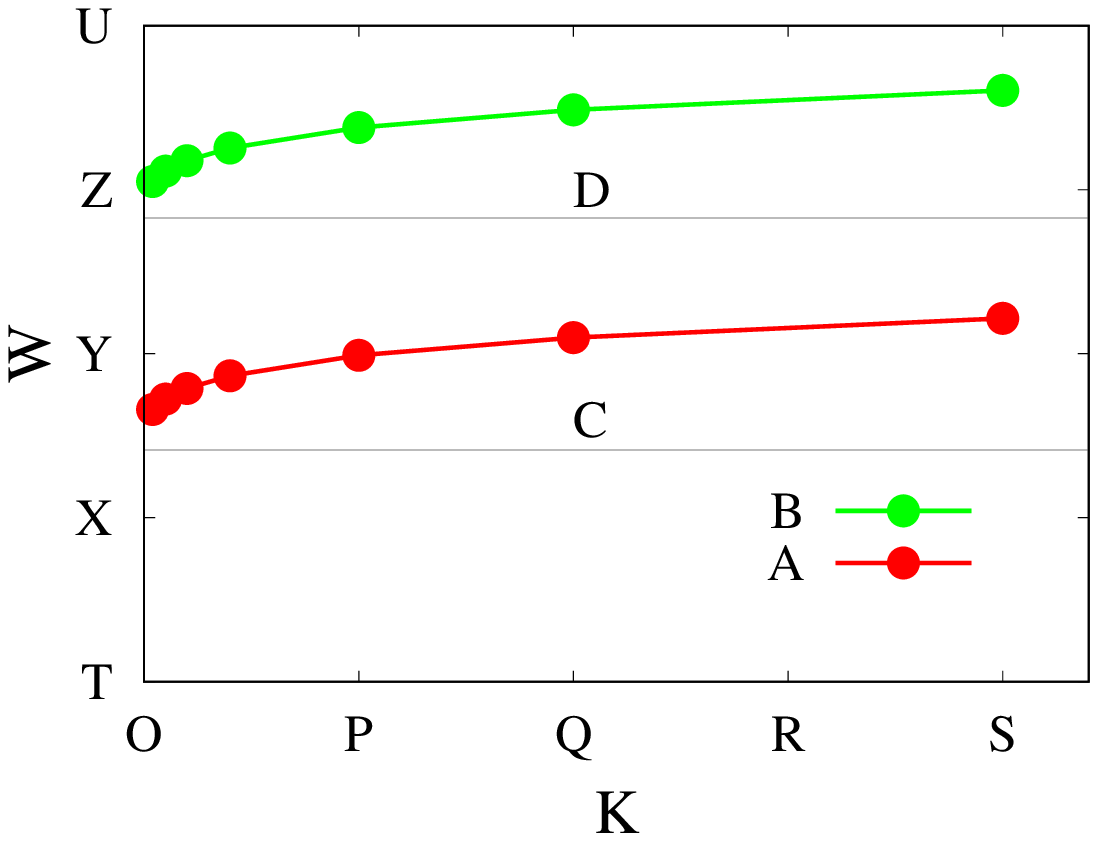}
	} &
	\subfigure{
		\psfrag{K}[c][c][0.9]{
			\begin{picture}(0,0)
				\put(0,0){\makebox(0,-6)[c]{Number of clusters: $k$} }
			\end{picture}
		}
		\psfrag{W}[c][c][0.9]{
			\begin{picture}(0,0)
				\put(0,0){\makebox(0,27)[c]{Avg. CPU time (sec)} }
			\end{picture}
		}
		\psfrag{P}[c][r][0.82]{$10^2$}
		\psfrag{Q}[c][c][0.82]{$10^3$}
		\psfrag{R}[c][c][0.82]{$10^4$}
		\psfrag{X}[r][r][0.82]{$10^{1}$}
		\psfrag{Y}[r][r][0.82]{$10^{2}$}
		\psfrag{Z}[r][r][0.82]{$10^{3}$}
		\psfrag{U}[r][r][0.82]{$10^{4}$}
		\psfrag{A}[r][r][0.65]{{\em IVF}}
		\psfrag{B}[r][r][0.65]{{\em IVFD}}
		\includegraphics[width=40mm]{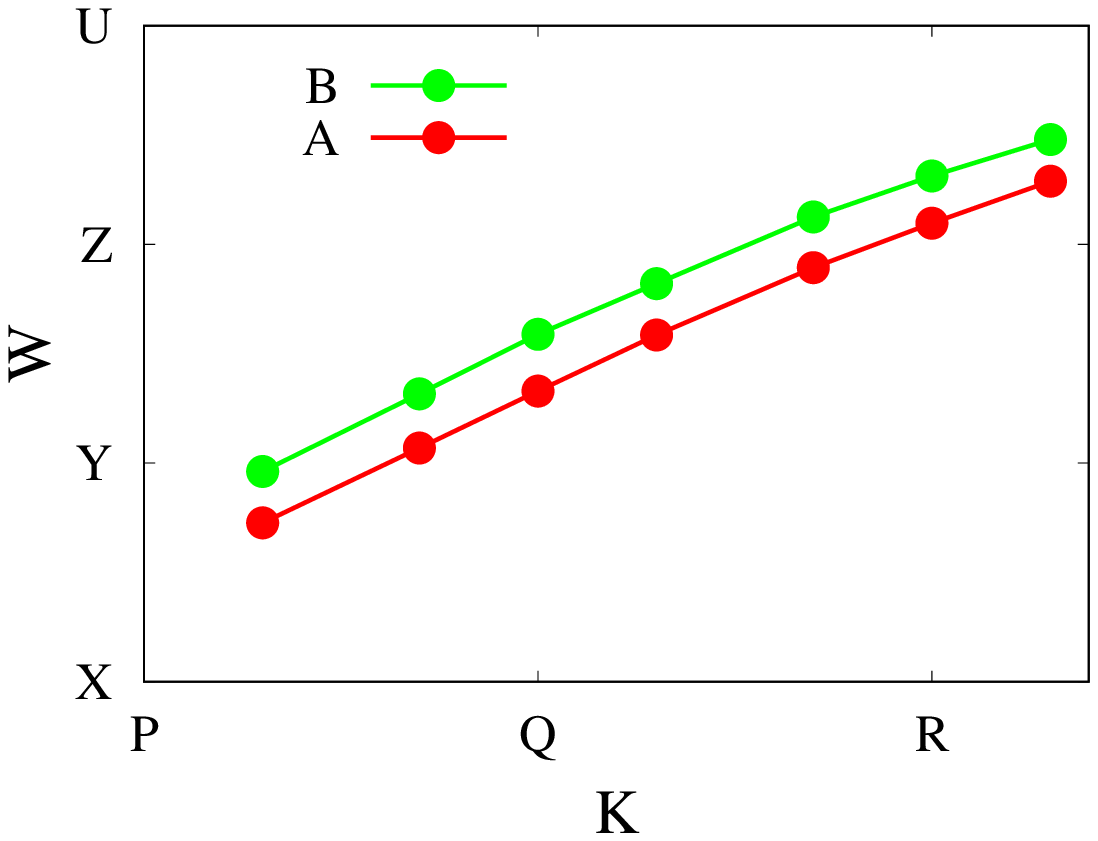}
	}\tabularnewline
	(a)~Required memory size  & (b)~CPU time
	\end{tabular}
\end{center}
\vspace*{-3mm}
\caption{Performance comparison of {\em IVFD} and {\em IVF} in PubMed: 
(a) Maximum memory size required through iterations is illustrated with $k$
and horizontal line of 0.707 indicates memory size occupied 
by object feature vectors.
(b) Average CPU time per iteration is depicted with $k$ in log-log scale.
}
\label{fig:ivfd_perform}
\vspace*{-1mm}
\end{figure}

{\em IVFD} and {\em IVF} were applied to PubMed for evaluating their performance.
Figures~\ref{fig:ivfd_perform}(a) and (b) show the performance-comparison 
results in terms of 
the maximum memory capacitance required by the algorithms through iterations
until the convergence and the average CPU time per iteration. 
The horizontal lines labeled 0.707 and 1.414 
in Fig.~\ref{fig:ivfd_perform}(a) denote the memory capacitances occupied by
the object feature vectors and the double capacitance.
{\em IVFD} used double capacitance for the object feature vectors 
as designed.
Regarding speed performance, 
{\em IVFD} needed more CPU time than {\em IVF} in all the $k$ ranges.
The maximum and minimum rates of the {\em IVFD}'s CPU time to the {\em IVF}'s 
were 1.82 at $k\!=\!1,000$ and 1.55 at $k\!=\!20,000$.
Although {\em IVF} employed the inverted-file data structure 
for the {\em variable} mean feature vectors at the update step, 
it operated faster than {\em IVFD}.
This is because constructing the inverted-file mean feature vectors is not
costly.
Importantly,
most CPU time is spent at not the update step but the assignment step.
In particular, both the algorithms spent 
at least 92\% of their CPU time 
for the triple loop in their assignment steps in all the $k$ ranges.

\begin{figure}[t]
\begin{center}\hspace*{1mm}
	\begin{tabular}{cc}	
	\subfigure{
		\psfrag{K}[c][c][1.0]{Number of clusters: $k$}
		\psfrag{W}[c][c][0.9]{
			\begin{picture}(0,0)
				\put(0,0){\makebox(0,31)[c]{Avg. \#instructions} }
			\end{picture}
		}
		\psfrag{P}[c][r][0.85]{$10^2$}
		\psfrag{Q}[c][c][0.85]{$10^3$}
		\psfrag{R}[c][c][0.85]{$10^4$}
		\psfrag{X}[r][r][0.85]{$10^{11}$}
		\psfrag{Y}[r][r][0.85]{$10^{12}$}
		\psfrag{Z}[r][r][0.85]{$10^{13}$}
		\psfrag{S}[r][r][0.85]{$10^{14}$}
		\psfrag{A}[r][r][0.68]{{\em IVF}}
		\psfrag{B}[r][r][0.68]{{\em IVFD}}
		\includegraphics[width=41mm]{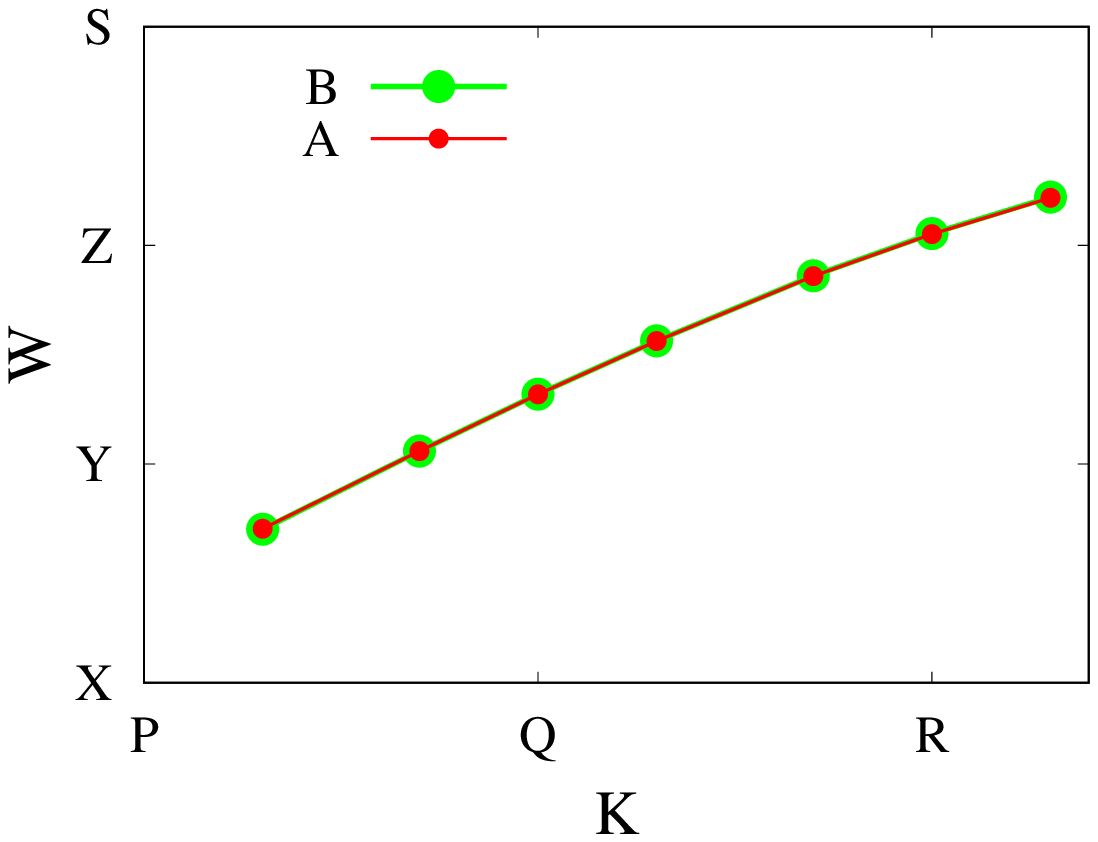}	
	} &
	\subfigure{
		\psfrag{K}[c][c][1.0]{Number of clusters: $k$}
		\psfrag{W}[c][c][0.9]{
			\begin{picture}(0,0)
				\put(0,0){\makebox(0,31)[c]{Avg. \#multiplications} }
			\end{picture}
		}
		\psfrag{P}[c][r][0.85]{$10^2$}
		\psfrag{Q}[c][c][0.85]{$10^3$}
		\psfrag{R}[c][c][0.85]{$10^4$}
		\psfrag{X}[r][r][0.85]{$10^{10}$}
		\psfrag{Y}[r][r][0.85]{$10^{11}$}
		\psfrag{Z}[r][r][0.85]{$10^{12}$}
		\psfrag{A}[r][r][0.68]{{\em IVF}}
		\psfrag{B}[r][r][0.68]{{\em IVFD}}
		\includegraphics[width=41mm]{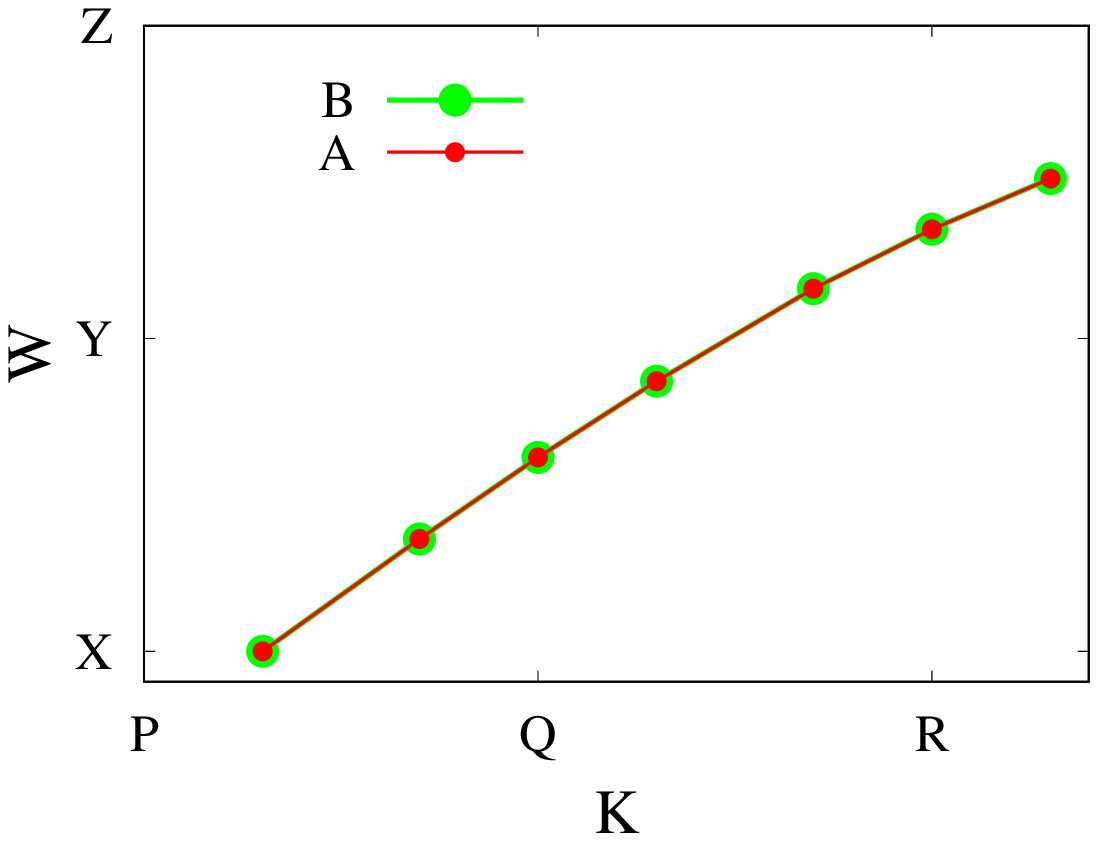}	
	}\tabularnewline
	(a)~Avg. \#instructions & (b)~Avg. \#multiplications
	\end{tabular}
\end{center}
\vspace*{-3mm}
\caption{Comparison of {\em IVFD} and {\em IVF} in PubMed: 
(a) Average number of instructions per iteration.
Values of two algorithms were almost equal.
Absolute difference of values at each $k$ value was at most 0.7\%.
(b) Average number of multiplications in triple loop per iteration.
Not only average but also exact numbers of multiplications
for each iteration at each $k$ value were equal.
}
\label{fig:ivfd_inst}
\vspace*{-1mm}
\end{figure}

Figure~\ref{fig:ivfd_inst}(a) shows the average number of instructions 
executed in the triple loop at the assignment step per iteration, and 
Fig.~\ref{fig:ivfd_inst}(b) shows  
the average number of multiplications operated in the triple loop 
per iteration.
The numbers of instructions executed by
{\em IVFD} and {\em IVF} were almost equal.
The absolute difference of the numbers of instructions of the algorithms 
at each $k$ value was at most 0.7\%.
Both used identical number of multiplications
at each iteration.
This can be confirmed by comparing {\bf Algorithm~\ref{algo:ivfd}} with 
{\bf Algorithm~\ref{algo:ours}}.
Both performed the multiplications 
illustrated as the volume in Fig.~\ref{fig:diag}(b).

\begin{figure}[t]
\begin{center}\hspace*{1mm}
	\begin{tabular}{cc}	
	\subfigure{
		\psfrag{K}[c][c][0.9]{
			\begin{picture}(0,0)
				\put(0,0){\makebox(0,-6)[c]{Number of clusters: $k$} }
			\end{picture}
		}
		\psfrag{W}[c][c][1.0]{
			\begin{picture}(0,0)
				\put(0,0){\makebox(0,24)[c]{CPI} }
			\end{picture}
		}
		\psfrag{P}[l][r][0.82]{$10^2$}
		\psfrag{Q}[c][c][0.82]{$10^3$}
		\psfrag{R}[c][c][0.82]{$10^4$}
		\psfrag{X}[r][r][0.82]{0}
		\psfrag{Y}[r][r][0.82]{0.1}
		\psfrag{Z}[r][r][0.82]{0.2}
		\psfrag{S}[r][r][0.82]{0.3}
		\psfrag{T}[r][r][0.82]{0.4}
		\psfrag{U}[r][r][0.82]{0.5}
		\psfrag{A}[r][r][0.65]{\em IVF}
		\psfrag{B}[r][r][0.65]{\em IVFD}
		\includegraphics[width=41mm]{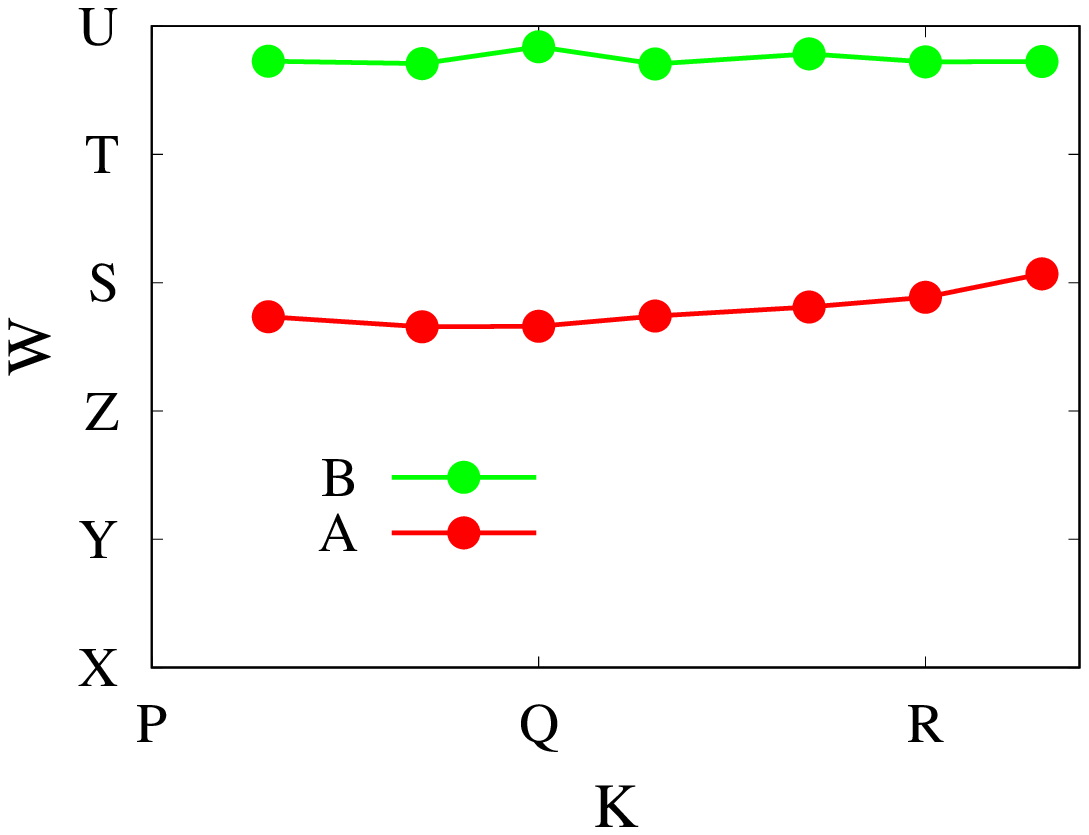}
	} &
	\subfigure{
		\psfrag{K}[c][c][0.9]{
			\begin{picture}(0,0)
				\put(0,0){\makebox(0,-6)[c]{Number of clusters: $k$} }
			\end{picture}
		}
		\psfrag{W}[c][c][0.9]{
			\begin{picture}(0,0)
				\put(0,0){\makebox(0,22)[c]{L1CM$'$/~Inst $(\times\!10^{-3})$} }
			\end{picture}
		}
		\psfrag{P}[l][r][0.82]{$10^2$}
		\psfrag{Q}[c][c][0.82]{$10^3$}
		\psfrag{R}[c][c][0.82]{$10^4$}
		\psfrag{X}[r][r][0.82]{0}
		\psfrag{Y}[r][r][0.82]{5}
		\psfrag{Z}[r][r][0.82]{10}
		\psfrag{S}[r][r][0.82]{15}
		\psfrag{T}[r][r][0.82]{20}
		\psfrag{A}[r][r][0.65]{{\em IVF}}
		\psfrag{B}[r][r][0.65]{{\em IVFD}}
		\includegraphics[width=41mm]{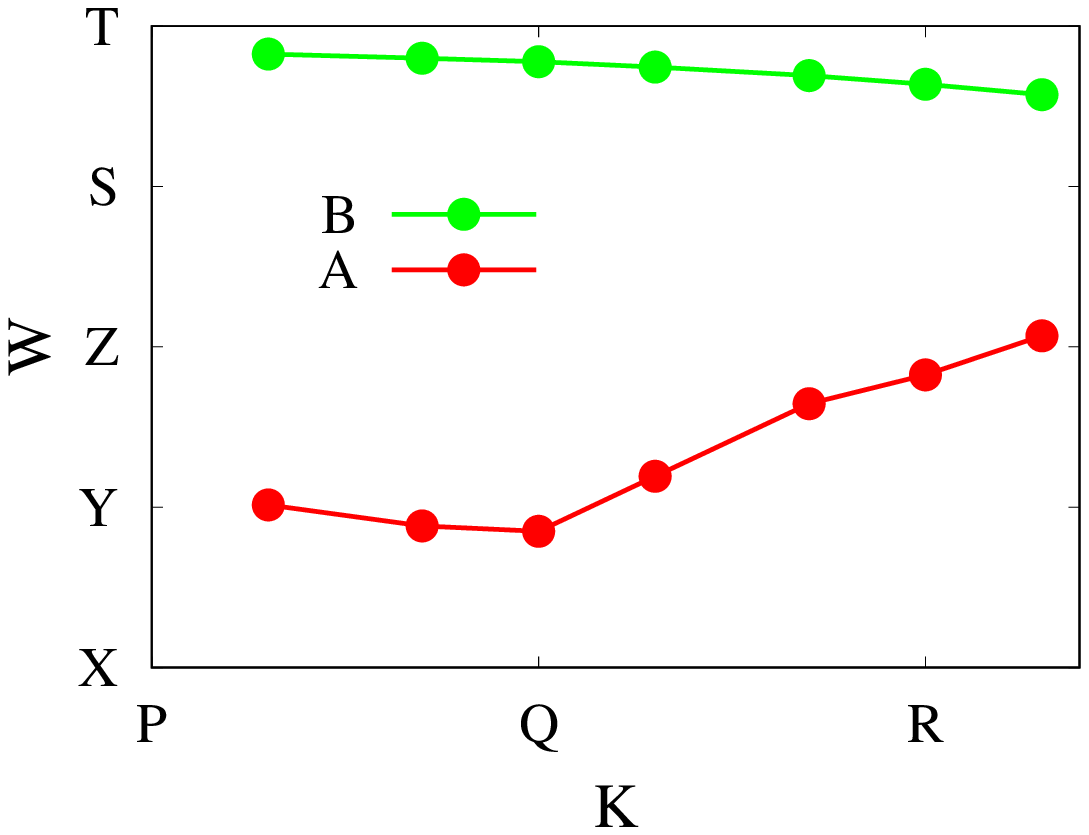}
	}\tabularnewline
	(a)~CPI & (b)~\#L1-cache misses
	\tabularnewline
	\subfigure{
		\psfrag{K}[c][c][0.9]{
			\begin{picture}(0,0)
				\put(0,0){\makebox(0,-6)[c]{Number of clusters: $k$} }
			\end{picture}
		}
		\psfrag{W}[c][c][0.9]{
			\begin{picture}(0,0)
				\put(0,0){\makebox(0,18)[c]{LLCM~/~Inst $(\times\! 10^{-3})$} }
			\end{picture}
		}
		\psfrag{P}[l][r][0.82]{$10^2$}
		\psfrag{Q}[c][c][0.82]{$10^3$}
		\psfrag{R}[c][c][0.82]{$10^4$}
		\psfrag{X}[r][r][0.82]{0}
		\psfrag{Y}[r][r][0.82]{1}
		\psfrag{Z}[r][r][0.82]{2}
		\psfrag{S}[r][r][0.82]{3}
		\psfrag{T}[r][r][0.82]{4}
		\psfrag{A}[r][r][0.65]{{\em IVF}}
		\psfrag{B}[r][r][0.65]{{\em IVFD}}
		\includegraphics[width=41mm]{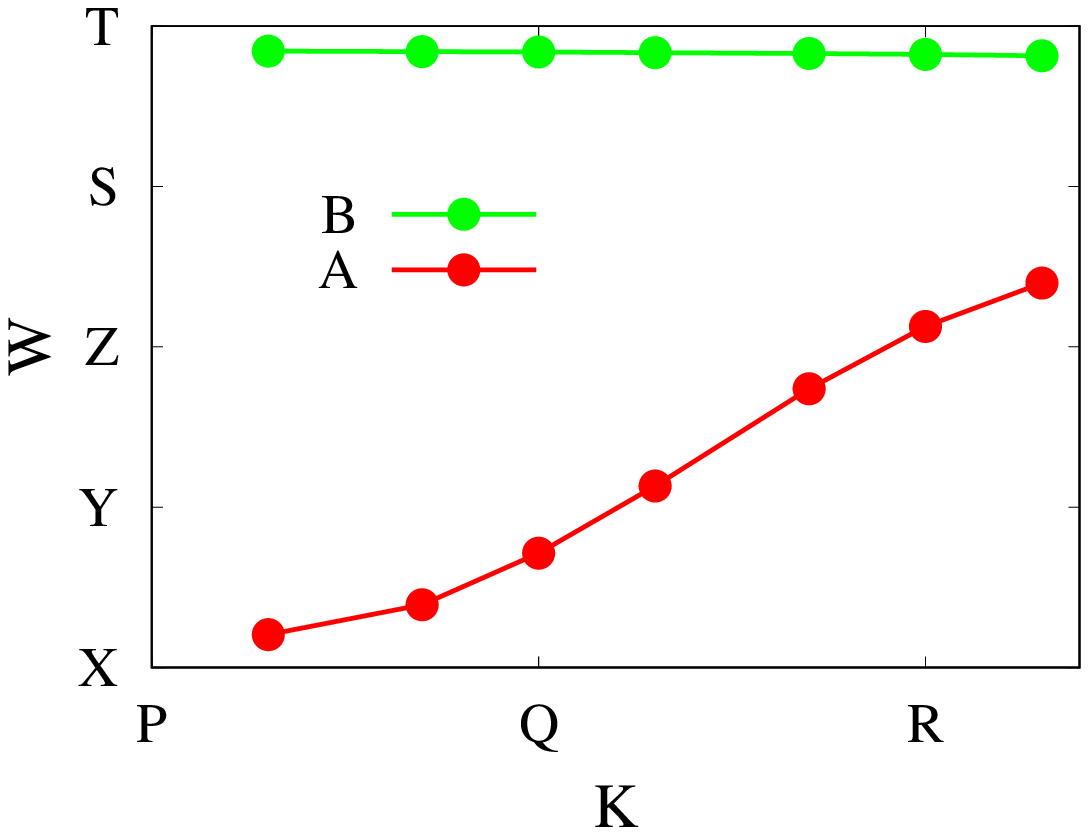}
	} &
	\subfigure{
		\psfrag{K}[c][c][0.9]{
			\begin{picture}(0,0)
				\put(0,0){\makebox(0,-6)[c]{Number of clusters: $k$} }
			\end{picture}
		}
		\psfrag{W}[c][c][0.9]{
			\begin{picture}(0,0)
				\put(0,0){\makebox(0,31)[c]{BM/~Inst} }
			\end{picture}
		}
		\psfrag{P}[l][r][0.82]{$10^2$}
		\psfrag{Q}[c][c][0.82]{$10^3$}
		\psfrag{R}[c][c][0.82]{$10^4$}
		\psfrag{X}[r][r][0.82]{$10^{-6}$}
		\psfrag{Y}[r][r][0.82]{$10^{-5}$}
		\psfrag{Z}[r][r][0.82]{$10^{-4}$}
		\psfrag{S}[r][r][0.82]{$10^{-3}$}
		\psfrag{A}[r][r][0.65]{{\em IVF}}
		\psfrag{B}[r][r][0.65]{{\em IVFD}}
		\includegraphics[width=41mm]{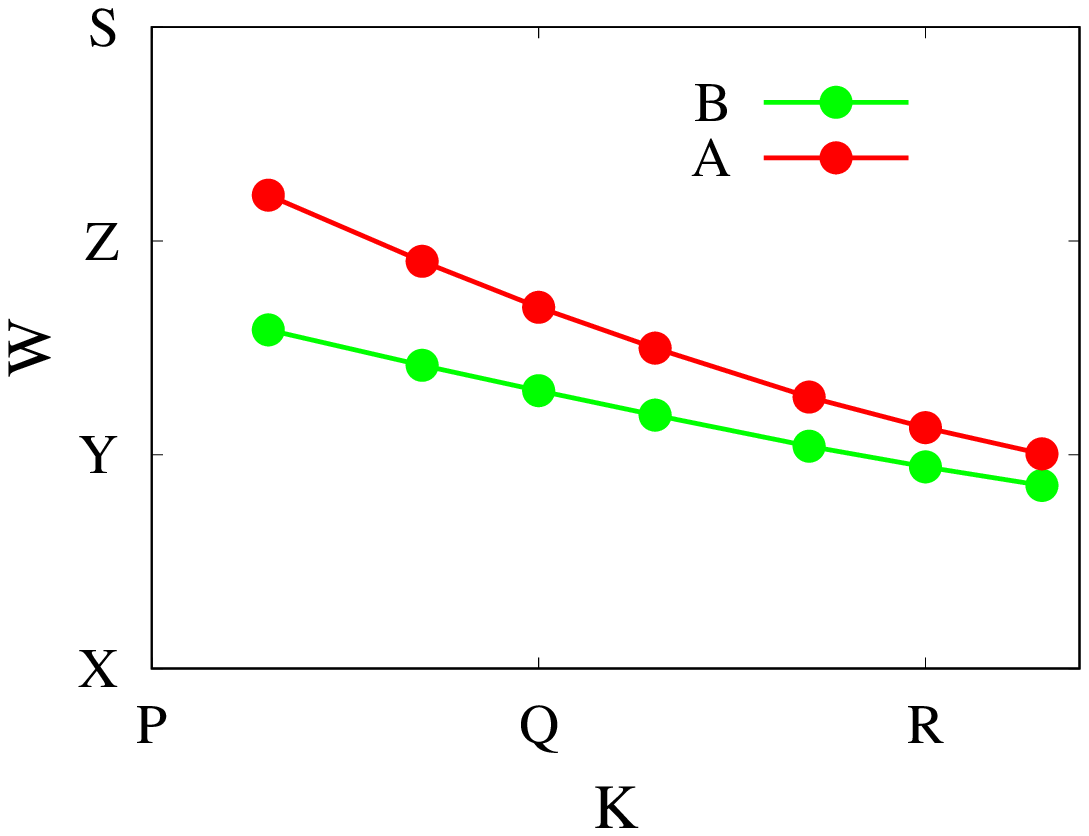}
	}\tabularnewline
	(c)~\#LL-cache misses & (d)~\#branch mispredictions
	\end{tabular}
\end{center}
\vspace*{-3mm}
\caption{Comparison of performance degradation factor characteristics 
obtained by {\em IVFD} and {\em IVF} in PubMed:
(a) Clock cycles per instruction (CPI),
(b) Difference of numbers of L1-cache and LL-cache misses
per instruction,
(c) Number of LL-cache misses per instruction, and 
(d) Number of branch mispredictions per instruction in log-log scale.
}
\label{fig:ivfd_dfs}
\vspace*{-2mm}
\end{figure}

\subsubsection{Analysis Based on CPI Model}
To identify why 
{\em IVFD} needed more CPU time 
despite executing almost the same number of instructions as {\em IVF},
we analyzed {\em IVFD} from the viewpoint of 
performance degradation factors (DFs) and compared it with {\em IVF}.
Figures~\ref{fig:ivfd_dfs}(a), (b), (c), and (d) show 
CPI, the number of L1-data cache misses excluding the LL-cache misses (LLCM$'$)
per instruction, the number of LL-cache misses (LLCM) per instruction,
and the number of branch mispredictions (BM) per instruction, respectively.

\begin{figure}[t]
\begin{center}
	\begin{tabular}{cc}	
	\subfigure{
		\psfrag{K}[c][c][0.9]{
			\begin{picture}(0,0)
				\put(0,0){\makebox(0,-5)[c]{Number of clusters: $k (\times 10^3)$} }
			\end{picture}
		}
		\psfrag{W}[c][c][0.9]{
			\begin{picture}(0,0)
				\put(0,0){\makebox(0,22)[c]{Contribution rate} }
			\end{picture}
		}
		\psfrag{A}[c][c][0.86]{0.2}
		\psfrag{B}[c][c][0.86]{0.5}
		\psfrag{C}[c][c][0.86]{1}
		\psfrag{D}[c][c][0.86]{2}
		\psfrag{E}[c][c][0.86]{5}
		\psfrag{F}[c][c][0.86]{10}
		\psfrag{G}[c][c][0.86]{20}
		\psfrag{X}[r][r][0.86]{0}
		\psfrag{Y}[r][r][0.86]{0.2}
		\psfrag{Z}[r][r][0.86]{0.4}
		\psfrag{S}[r][r][0.86]{0.6}
		\psfrag{T}[r][r][0.86]{0.8}
		\psfrag{U}[r][r][0.86]{1.0}
		\psfrag{P}[r][r][0.86]{L1CM$'$}
		\psfrag{Q}[r][r][0.86]{LLCM}
		\includegraphics[width=41mm]{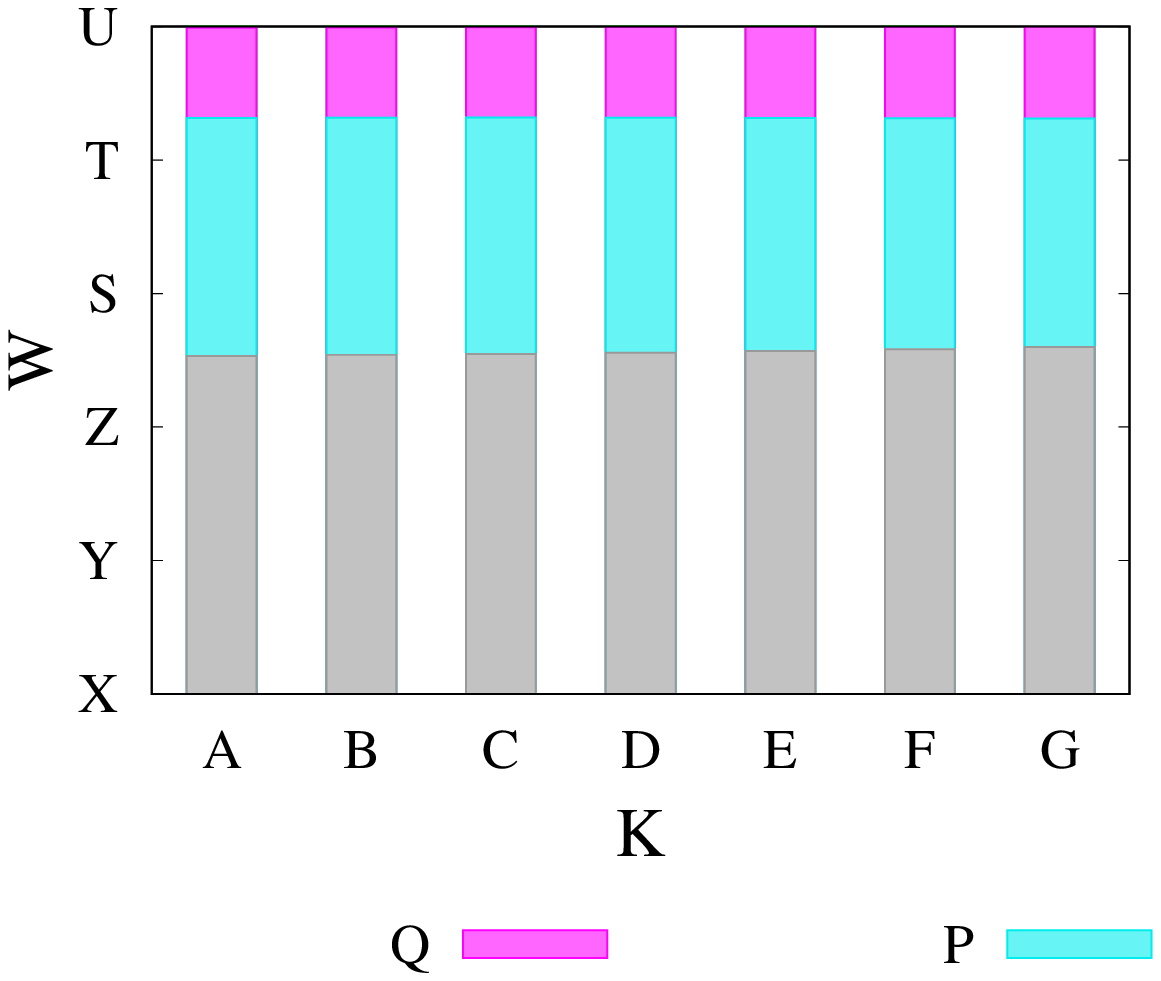}
	} &
	\subfigure{
		\psfrag{K}[c][c][0.9]{
			\begin{picture}(0,0)
				\put(0,0){\makebox(0,-5)[c]{Number of clusters: $k (\times 10^3)$} }
			\end{picture}
		}
		\psfrag{W}[c][c][0.9]{
			\begin{picture}(0,0)
				\put(0,0){\makebox(0,22)[c]{Contribution rate} }
			\end{picture}
		}
		\psfrag{A}[c][c][0.86]{0.2}
		\psfrag{B}[c][c][0.86]{0.5}
		\psfrag{C}[c][c][0.86]{1}
		\psfrag{D}[c][c][0.86]{2}
		\psfrag{E}[c][c][0.86]{5}
		\psfrag{F}[c][c][0.86]{10}
		\psfrag{G}[c][c][0.86]{20}
		\psfrag{X}[r][r][0.86]{0}
		\psfrag{Y}[r][r][0.86]{0.2}
		\psfrag{Z}[r][r][0.86]{0.4}
		\psfrag{S}[r][r][0.86]{0.6}
		\psfrag{T}[r][r][0.86]{0.8}
		\psfrag{U}[r][r][0.86]{1.0}
		\psfrag{O}[r][r][0.86]{Inst}
		\psfrag{R}[r][r][0.86]{BM}
		\includegraphics[width=41mm]{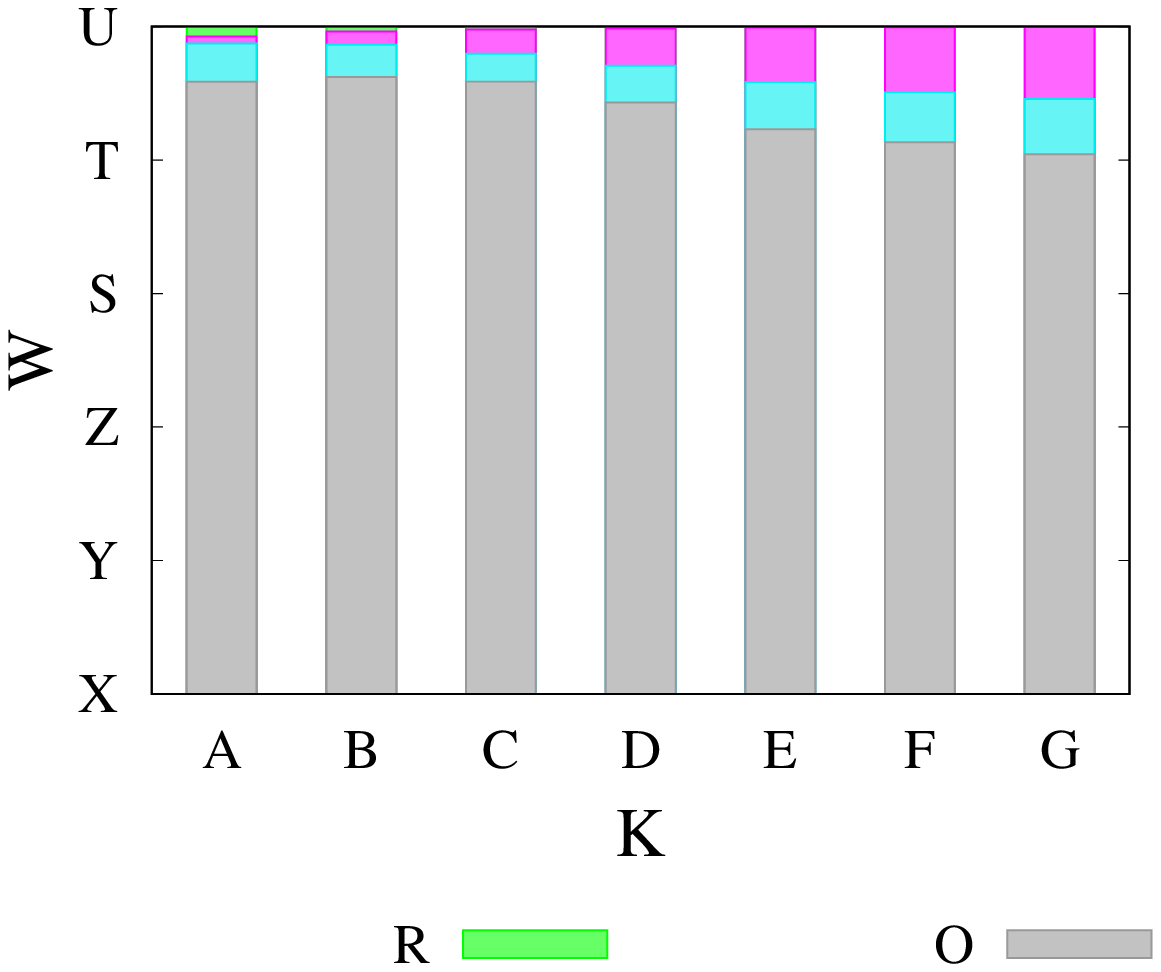}
	}\tabularnewline
	(a)~{\em IVFD} & (b)~{\em IVF}
	\end{tabular}
\end{center}
\vspace*{-3mm}
\caption{Contribution rate of each DF to CPU time
when (a) {\em IVFD} and (b) {\em IVF} were applied to PubMed. 
DFs are Inst, L1CM$'$, LLCM, and BM.
}
\label{fig:contrib}
\vspace*{-2mm}
\end{figure}

The difference in the CPIs in Fig.~\ref{fig:ivfd_dfs}(a) corresponds to
the CPU time in Fig.~\ref{fig:ivfd_perform}(b) 
since the numbers of instructions executed by both algorithms 
were almost identical.
Actually, the {\em IVF}'s CPI ranged from 0.27 to 0.31, and 
{\em IVFD}'s ranged from 0.47 to 0.48.
The rates of the {\em IVFD}'s CPIs to the {\em IVF}'s at 
$k\!=\!1,000,~20,000$ 
were 1.82 and 1.54, nearly equal to the CPU time rates.
In Fig.~\ref{fig:ivfd_dfs}(d), 
{\em IVF} had more branch mispredictions than {\em IVFD}.
However, 
the number was too small, compared with those of the other DFs;
its contribution to the CPU time can be ignored, 
as shown in Fig.~\ref{fig:contrib}(b).
The difference in the CPU times (the clock cycles) came from 
the number of cache misses in Figs.~\ref{fig:ivfd_dfs}(b) and (c).
{\em IVFD}'s L1CM$'$ and LLCM per instruction were constant high values.
By contrast, {\em IVF}'s L1CM$'$ and LLCM per instruction 
increased with $k$.
These characteristics of the LLCMs are explained based on 
our cache-miss models in Section~\ref{subsubsec:LLCM_model}.

\begin{table}[t]
\centering
\caption{Optimized CPI model parameters and errors on actual CPIs}
\vspace*{-2mm}
\begin{spacing}{1.1}
\begin{tabular}{|c|c|c|c|c||c|c|}\hline
\multirow{2}{*}{Algo.} & \multicolumn{4}{c||}{Parameters} &
{\small Avg. err.} & {\small Max. err.} \tabularnewline \cline{2-5}
& $w_0$ & $w_1$ & $w_2$ & $w_3$ & (\%) & (\%)
\tabularnewline \hline\hline
{\em IVFD} & 0.243 & {\bf 8.94} & {\bf 16.8} & 23.8 & 0.445 & 1.52
\tabularnewline \hline
{\em IVF} & 0.243 & {\bf 3.13} & {\bf 13.5} & 23.8
& 0.461 & 3.19 \tabularnewline \hline
\end{tabular}\label{table:ivfd_params}
\end{spacing}
\vspace*{-2mm}
\end{table}

\begin{figure}[t]
\begin{center}
	\psfrag{K}[c][c][1.0]{
		\begin{picture}(0,0)
			\put(0,0){\makebox(0,-6)[c]{Number of clusters: $k$} }
		\end{picture}
	}
	\psfrag{W}[c][c][1.1]{
		\begin{picture}(0,0)
			\put(0,0){\makebox(0,20)[c]{CPI} }
		\end{picture}
	}
	\psfrag{P}[c][r][0.9]{$10^2$}
	\psfrag{Q}[c][c][0.9]{$10^3$}
	\psfrag{R}[c][c][0.9]{$10^4$}
	\psfrag{X}[r][r][0.9]{0}
	\psfrag{Y}[r][r][0.9]{0.1}
	\psfrag{Z}[r][r][0.9]{0.2}
	\psfrag{S}[r][r][0.9]{0.3}
	\psfrag{T}[r][r][0.9]{0.4}
	\psfrag{U}[r][r][0.9]{0.5}
	\psfrag{A}[r][r][0.75]{\em IVF}
	\psfrag{B}[r][r][0.75]{\em IVFD}
	\psfrag{M}[r][r][0.75]{Model}
	\includegraphics[width=55mm]{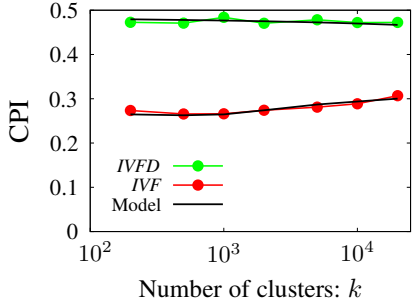}
\end{center}
\vspace*{-2mm}
\caption{Actual and model CPI of {\em IVFD} and {\em IVF} along $k$ 
in PubMed}
\label{fig:ivfd_fit}
\end{figure}

We optimized the {\em IVFD} parameters by referring to the procedure 
in Section~\ref{subsec:cpimodel}
to determine the contribution rates of the DFs to the CPU time.
We assumed for the optimization that
parameters $w_0$ and $w_3$ of {\em IVFD} in Eq.~(\ref{eq:model})
were fixed at the same values as those of {\em IVF}. 
The optimized parameters and results are shown in 
Table~\ref{table:ivfd_params} and Fig.~\ref{fig:ivfd_fit}.
The optimized CPI model agrees well with the actual CPIs since 
the average error and the maximum error in {\em IVFD} were
0.445\% and 1.52\%.
Parameters $w_1$ and $w_2$ were larger than those of {\em IVF}; 
the stall clock cycles per cache miss were longer.
Thus {\em IVFD} had more cache misses, 
each of which induced longer stall clock cycles.

Figures~\ref{fig:contrib}(a) and (b) show the contribution rates 
of each DF to the CPU times in {\em IVFD} and {\em IVF}.
The rates of L1CM$'$ and LLCM were high in {\em IVFD}, and 
the rate of Inst occupied much of the whole of contribution rate
in {\em IVF}. 
In terms of branch misprediction (BM),
its contribution rates in {\em IVFD} and {\em IVF} were 
very small, and we can ignore its values.
Since the number of instructions and 
parameter $w_0$ were equal in {\em IVFD} and {\em IVF},
the {\em IVFD}'s performance degradation was caused by cache misses, 
more of which were caused by the long arrays $\bm{\zeta}_s$ 
of object inverted-file $\breve{\cal X}$ 
in the innermost loop in the triple loop in
{\bf Algorithm~\ref{algo:ivfd}}.

\subsubsection{LL-Cache-Miss Models for IVFD and IVF}\label{subsubsec:LLCM_model}
Figure~\ref{fig:ivfd_dfs}(c) shows that 
the number of last-level cache misses (LLCM) of {\em IVF} 
increased with $k$ 
while that of {\em IVFD} was 
almost constant in the $k$ range.
We analyzed these characteristics.

The last-level (LL) cache used in our experiments contained 
36,700,160 (35 M) bytes in 64-byte blocks 
with 20-way set associative placement 
and least-recently used (LRU) replacement.
Instead of the actual set associative replacement,
we assumed fully associative one in our analysis.
Both {\em IVFD} and {\em IVF} used an inverted-file data structure
that consisted of two arrays for 4-byte IDs of objects or centroids
and 8-byte feature values.

{\em IVF} calculates similarities (inner products) between 
an object and all centroids (means) in the middle and innermost loop 
in the triple loop at its assignment step.
A probability that a term with global term ID $p$ is used for 
a similarity calculation is $(no)_p/N$, 
where $(no)_p$ denotes the number of objects 
that contain the $p$-th term, i.e., the document frequency of the term.
When the array related to the $p$-th term, 
$\breve{\bm{\xi}}_p\!=\!(c_{(p,q)},u_{(p,q)})_{q=1}^{(nc)_p}\!\in\!\breve{\cal M}$, 
is accessed, 
the number of blocks ({\em NB}$_{[\mbox{\tiny\em IVF}]}$) that are placed 
into the LL cache from the main memory is given by
\begin{equation}
\mbox{\em NB}_{[\mbox{\tiny\em IVF}]} =
\left\lceil (nc)_p \times 
\frac{\mbox{sizeof(int + double)}}{\mbox{block size}}
\right\rceil \: ,
\label{eq:nb_ivf}
\end{equation}
where $(nc)_p$ denotes the number of centroids that contain the $p$-th term
and depends on $k$.
Then the expected number of blocks that are placed into the LL cache 
is expressed by
\begin{equation}
\mathbb{E}[\mbox{\em NB}_{[\mbox{\tiny\em IVF}]}]=
\sum_{p=1} ^D \frac{(no)_p}{N}\cdot
\left\lceil (nc)_p\cdot \gamma \right\rceil\: ,
\label{eq:E_nb_ivf}
\end{equation}
where $\gamma$ denotes (sizeof(int + double))/(block size).
By contrast, 
when {\em IVFD} calculates similarities between a centroid and all objects,
the expected number of blocks ({\em NB}$_{[\mbox{\tiny\em IVFD}]}$)
is expressed by  
\begin{equation}
\mathbb{E}[\mbox{\em NB}_{[\mbox{\tiny\em IVFD}]}]=
\sum_{p=1} ^D \frac{(nc)_p}{k}\cdot
\left\lceil (no)_p\cdot \gamma \right\rceil\: .
\label{eq:E_nb_ivfd}
\end{equation}
Assume that $(nc)_p\cdot \gamma$ and $(no)_p\cdot \gamma$ are integers.
Then Eqs.~(\ref{eq:E_nb_ivf}) and (\ref{eq:E_nb_ivfd}) are simplified:
\begin{eqnarray}
\mathbb{E}[\mbox{\em NB}_{[\mbox{\tiny\em IVF}]}] &=&
\frac{1}{N}\left(\gamma \textstyle{\sum_{p=1}^D (no)_p (nc)_p}\right) \\
\mathbb{E}[\mbox{\em NB}_{[\mbox{\tiny\em IVFD}]}] &=&
\frac{1}{k}\left(\gamma \textstyle{\sum_{p=1}^D (no)_p (nc)_p}\right) \: ,
\end{eqnarray}
where $\sum_{p=1}^D (no)_p (nc)_p$ is the number of multiplications 
that is illustrated as the volume in Fig.~\ref{fig:diag}(b).
It is clear that 
\begin{equation}
\mathbb{E}[\mbox{\em NB}_{[\mbox{\tiny\em IVF}]}] \ll
\mathbb{E}[\mbox{\em NB}_{[\mbox{\tiny\em IVFD}]}]~~~~(N\gg k)\: . 
\label{eq:comp_nb0}
\end{equation}

We compare 
$\mathbb{E}[\mbox{\em NB}_{[\mbox{\tiny\em IVF}]}]$ and 
$\mathbb{E}[\mbox{\em NB}_{[\mbox{\tiny\em IVFD}]}]$
with the number of blocks in the actual LL cache  
when {\em IVF} and {\em IVFD} are applied to PubMed 
($N\!=\!1\!\times\! 10^6$), given $k\!=\!1\!\times\!10^4$. 
In this comparison, 
we assume that the number of available blocks 
($\mbox{\em NB}_{[\mbox{\tiny\em LLC}]}$)
is $5\!\times\! 10^5$ that corresponds to 32 MB.
The number of multiplications executed by {\em IVF} was $2.21\!\times\! 10^{11}$
shown in Fig.~\ref{fig:triple_mult}(a) and $\gamma\!=\!(4+8)/64\!=\!3/16$.
Then 
$\mathbb{E}[\mbox{\em NB}_{[\mbox{\tiny\em IVF}]}]\!\sim\!4\!\times\! 10^4$
and
$\mathbb{E}[\mbox{\em NB}_{[\mbox{\tiny\em IVFD}]}]\!\sim\!4\!\times\! 10^6$.
The inequality in Eq.~(\ref{eq:comp_nb0}) is rewritten as 
\begin{equation}
\mathbb{E}[\mbox{\em NB}_{[\mbox{\tiny\em IVF}]}] \ll
\mbox{\em NB}_{[\mbox{\tiny\em LLC}]} \ll
\mathbb{E}[\mbox{\em NB}_{[\mbox{\tiny\em IVFD}]}]\:.
\label{eq:comp_nb}
\end{equation}
This inequality held in the $k$ range 
from 200 to 20,000 when the algorithms were applied to PubMed.

\begin{figure}[t]
\begin{center}
	\psfrag{K}[c][c][1.0]{
		\begin{picture}(0,0)
			\put(0,0){\makebox(0,-6)[c]{Number of clusters: $k$} }
		\end{picture}
	}
	\psfrag{W}[c][c][0.9]{
		\begin{picture}(0,0)
			\put(0,0){\makebox(0,18)[c]{LLCM~/~Inst $(\times\! 10^{-3})$} }
		\end{picture}
	}
	\psfrag{P}[l][r][0.9]{$10^2$}
	\psfrag{Q}[c][c][0.9]{$10^3$}
	\psfrag{R}[c][c][0.9]{$10^4$}
	\psfrag{X}[r][r][0.9]{0}
	\psfrag{Y}[r][r][0.9]{1}
	\psfrag{Z}[r][r][0.9]{2}
	\psfrag{S}[r][r][0.9]{3}
	\psfrag{T}[r][r][0.9]{4}
	\psfrag{A}[r][r][0.75]{{\em IVF}}
	\psfrag{B}[r][r][0.75]{{\em IVFD}}
	\psfrag{C}[r][r][0.75]{{Model}}
	\includegraphics[width=52mm]{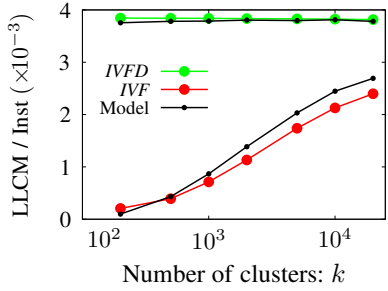}
\end{center}
\vspace*{-2mm}
\caption{Actual and model (LLCM/Inst) of {\em IVFD} and {\em IVF} along $k$ 
in PubMed}
\label{fig:append_llcm}
\end{figure}

The fact of $\mathbb{E}[\mbox{\em NB}_{[\mbox{\tiny\em IVFD}]}] \!\gg\!
\mbox{\em NB}_{[\mbox{\tiny\em LLC}]}$ means 
that {\em IVFD} almost always fails to use feature values 
in the LL cache like {\em cold-start misses}.
Based on this, 
we assume that the blocks required by {\em IVFD} must be always brought 
into the LL cache from the main memory.
Then the number of LL-cache misses (LLCM$_{[\mbox{\tiny\em IVFD}]}$) is 
given by
\begin{eqnarray}
\mbox{LLCM}_{[\mbox{\tiny\em IVFD}]}
&=& \textstyle{\sum_{p=1}^D (nc)_p\lceil (no)_p\cdot\gamma \rceil} 
\label{eq:exact_llcm_ivfd} \\
&\sim& \gamma\textstyle{\sum_{p=1}^D (no)_p (nc)_p} \: .
\label{eq:approx_llcm_ivfd}
\end{eqnarray}
We show 
the rate of LLCM$_{[\mbox{\tiny\em IVFD}]}$ in Eq.~(\ref{eq:exact_llcm_ivfd})
to the number of instructions (Inst$_{[\mbox{\tiny\em IVFD}]}$) that 
was obtained in our experiments  
as Model in Fig.~\ref{fig:append_llcm}.
The model curve coincided with the actual rate depicted as {\em IVFD} 
in Fig.~\ref{fig:ivfd_dfs}(c).
Furthermore, we approximate the rate as 
\begin{equation}
\frac{\mbox{LLCM}_{[\mbox{\tiny\em IVFD}]}}{\mbox{Inst}_{[\mbox{\tiny\em IVFD}]}}\sim
\frac{\gamma\sum_{p=1}^D (no)_p (nc)_p}{\beta\times \mbox{(\#multiplications)}}
= \frac{\gamma}{\beta}\: ,
\label{eq:approx_rate_ivfd}
\end{equation}
where $\beta$ is the same constant value\footnote{
Analysis of {\em IVFD} and {\em IVF} assembly codes showed that
both algorithms used the identical number of instructions 
for each multiplication and addition operation.
} 
as that for {\em IVF} in Eq.~(\ref{eq:vol})
and \#multiplications denotes $\sum_{p=1}^D (no)_p (nc)_p$.
Since $\beta\!=\!40$ and $\gamma\!=\!3/16$ in our experiments,
$\mbox{LLCM}_{[\mbox{\tiny\em IVFD}]}/\mbox{Inst}_{[\mbox{\tiny\em IVFD}]}\!=\!4.7\!\times\!10^{-3}$.
This approximate value is not so far from the {\em IVFD} values 
in Fig.~\ref{fig:ivfd_dfs}(c) 
and higher than the corresponding values in Fig.~\ref{fig:append_llcm} 
because $\beta\!\times\!\mbox{(\#multiplications)}$ is slightly smaller than 
actual Inst$_{[\mbox{\tiny\em IVFD}]}$.
Thus 
$\mbox{LLCM}_{[\mbox{\tiny\em IVFD}]}/\mbox{Inst}_{[\mbox{\tiny\em IVFD}]}$ 
becomes the constant value depending on the computer architecture.

%-- IVF --%
Next, we model the {\em IVF} LLCM 
(LLCM$_{[\mbox{\tiny\em IVF}]}$). 
When {\em IVF} calculates similarities 
among successive $z$ objects and all centroids,
the expected number of blocks that are placed into the LL cache
($\mathbb{E}[\mbox{\em NB}_{[\mbox{\tiny\em IVF}]}^{(z)}]$) 
is given by
\begin{equation}
\mathbb{E}[\mbox{\em NB}_{[\mbox{\tiny\em IVF}]}^{(z)}]
=\sum_{p=1}^D\left\{ 1 -\left(1-\frac{(no)_p}{N}\right)^z \right\}
\lceil (nc)_p \gamma \rceil \:.
\label{eq:ivf_nb}
\end{equation}
Note that 
$\mathbb{E}[\mbox{\em NB}_{[\mbox{\tiny\em IVF}]}^{(1)}]\!=\!
\mathbb{E}[\mbox{\em NB}_{[\mbox{\tiny\em IVF}]}]$ in Eq.~(\ref{eq:E_nb_ivf}).
Let $z^{*}$ denote the maximum integer $z$ under the condition that
$\mathbb{E}[\mbox{\em NB}_{[\mbox{\tiny\em IVF}]}^{(z)}]$ satisfies 
\begin{equation}
\mathbb{E}[\mbox{\em NB}_{[\mbox{\tiny\em IVF}]}^{(z)}] \leq 
\mbox{\em NB}_{[\mbox{\tiny\em LLC}]}\: .
\label{eq:z_cond}
\end{equation}
When $z\!=\!z^*$,
intuitively, the LL cache is fully occupied by 
arrays $\breve{\bm \xi}_{t_{(i,h)}}\!\in\!\breve{\cal M}$
related to terms that successive $z^*$ objects $\hat{\bm x}_i$ contain.
Consider that when the LL cache is at this state,
{\em IVF} requires array $\breve{\bm \xi}_p$
related to the $p$-th term, which is not placed in the LL cache.
Then 
the expected number of blocks that are placed into the LL cache
($\mathbb{E}[\mbox{\em NB}_{[\mbox{\tiny\em IVF}]}^{(z^*,\mbox{\scriptsize miss})}]$) 
is given by
\begin{equation}
\mathbb{E}[\mbox{\em NB}_{[\mbox{\tiny\em IVF}]}^{(z^*,\mbox{\scriptsize miss})}]
=\sum_{p=1}^D \frac{(no)_p}{N} \left(1-\!\frac{(no)_p}{N}\right)^{z^*}
\left\lceil (nc)_p \gamma \right\rceil \:.
\label{eq:miss_nb}
\end{equation}
Using this value, LLCM$_{[\mbox{\tiny\em IVF}]}$ is given
and approximated as
\begin{eqnarray}
\mbox{LLCM}_{[\mbox{\tiny\em IVF}]}
&=& N\cdot 
\mathbb{E}[\mbox{\em NB}_{[\mbox{\tiny\em IVF}]}^{(z^*,\mbox{\scriptsize miss})}]
\label{eq:exact_llcm_ivf} \\
&\sim& \gamma \sum_{p=1}^D (no)_p (nc)_p
\left(1-\!\frac{(no)_p}{N}\right)^{z^*} \: .
\label{eq:llcm_ivf}
\end{eqnarray}
The rate of LLCM$_{[\mbox{\tiny\em IVF}]}$ in Eq.~(\ref{eq:exact_llcm_ivf}) 
to Inst$_{[\mbox{\tiny\em IVF}]}$ in Fig.~\ref{fig:ivfd_dfs}(c) 
is shown as Model in Fig.~\ref{fig:append_llcm}.
The model curve gave close agreement with the values obtained by the experiments
and increased with $k$.
Furthermore, this rate is approximated as 
\begin{equation}
\frac{\mbox{LLCM}_{[\mbox{\tiny\em IVF}]}}{\mbox{Inst}_{[\mbox{\tiny\em IVF}]}}
\sim \left( \frac{\gamma}{\beta} \right)
\frac{\sum_{p=1}^D (no)_p (nc)_p \left(1-\!\frac{(no)_p}{N}\right)^{z^*}}
{\sum_{p=1}^D (no)_p (nc)_p} \: .
\label{eq:approx_rate_ivf}
\end{equation}
When $k$ approaches asymptotically to $N$, 
$z^*\!\rightarrow\! 0$ and $(nc)_p\!\rightarrow\!(no)_p$ 
in Eq.~(\ref{eq:approx_rate_ivf}).
Then 
\begin{equation}
\lim_{k\rightarrow N}
\frac{\mbox{LLCM}_{[\mbox{\tiny\em IVF}]}}{\mbox{Inst}_{[\mbox{\tiny\em IVF}]}}
\sim \frac{\gamma}{\beta}
\sim \frac{\mbox{LLCM}_{[\mbox{\tiny\em IVFD}]}}{\mbox{Inst}_{[\mbox{\tiny\em IVFD}]}} \: .
\label{eq:lim}
\end{equation}
$\mbox{LLCM}_{[\mbox{\tiny\em IVF}]}/\mbox{Inst}_{[\mbox{\tiny\em IVF}]}$ 
increases with $k$ and approached to $(\gamma/\beta)$ 
that is the approximate rate of {\em IVFD}.

Due to the reasons mentioned above, 
applying an inverted-file data structure to the mean feature vectors 
leads to better performance.
We should use {\em IVF} rather than {\em IVFD} to achieve high performance 
for large-scale sparse data sets.

\section{Conclusion}\label{sec:conc}
We proposed an inverted-file $k$-means clustering algorithm ({\em IVF})
that operated at high speed and with low memory consumption 
in large-scale high-dimensional sparse document data sets 
when large $k$ values were given.
{\em IVF} represents both the given object feature vectors and 
the mean feature vectors with sparse expression 
to conserve occupied memory capacitance and 
exploits the inverted-file data structure for 
the mean feature vectors to achieve high-speed performance.
We analyzed {\em IVF} using a newly introduced 
clock-cycle per instruction (CPI) model 
to identify factors for high-speed operation in a modern computer system.
Consequently, 
{\em IVF} suppressed the three performance degradation factors of
the numbers of cache misses, branch mispredictions, and completed instructions.

As future work, 
we will evaluate {\em IVF} in such practical environments as 
with parallel and distributed modern computer systems.

\section*{Acknowledgments}
This work was partly supported by JSPS KAKENHI Grant
Number JP17K00159.

\end{document}